\documentclass{article}

\usepackage{dynin_preprint,times}
\usepackage[utf8]{inputenc}
\usepackage[T1]{fontenc}
\usepackage{hyperref}
\usepackage{url}
\usepackage{booktabs}
\usepackage{amsfonts}
\usepackage{nicefrac}
\usepackage{microtype}
\usepackage{xcolor}
\usepackage{float}

\usepackage{url}
\usepackage{amssymb}
\usepackage[utf8]{inputenc}
\usepackage{microtype}
\usepackage{booktabs}
\usepackage{pifont} 
\usepackage{multirow}
\usepackage{makecell}
\usepackage{paralist}
\usepackage{xspace}
\usepackage{color}
\usepackage{xcolor}
\usepackage{colortbl}
\usepackage{adjustbox}
\usepackage{wrapfig}
\usepackage{environ}
\usepackage{multicol}
\usepackage{enumitem}
\usepackage{subcaption}
\usepackage{booktabs}
\usepackage{tabularx}
\usepackage{xcolor}
\usepackage{lipsum} 
\usepackage{amsmath,amssymb}
\usepackage{cleveref} 
\usepackage{caption}
\usepackage{amsfonts}
\usepackage{caption}
\usepackage{placeins}
\usepackage{amssymb}
\usepackage{pifont}
\usepackage{float}
\usepackage{afterpage}

\usepackage[toc,page,header]{appendix}
\usepackage{etoc}

\usepackage{fontawesome5} 
\usepackage{hyperref} 

\definecolor{lightblue}{RGB}{220,235,250}

\newcommand{\method}[1]{Dynin-Omni}
\newcommand{\ie}{\textit{i.e.}\xspace}
\newcommand{\eg}{\textit{e.g.}\xspace}

\newcommand{\nipsver}[1]{{\color{black}#1}}
\definecolor{omnirow}{RGB}{232,240,255}
\definecolor{oursrow}{RGB}{190,210,245}
\definecolor{undomni}{RGB}{230,240,255}   %
\definecolor{genomni}{RGB}{230,240,255}
\definecolor{darkgreen}{RGB}{0,128,0}

\title{\method{}: Omnimodal Unified Large Diffusion Language Model}

\newcommand{\authorid}[1]{\href{mailto:#1@snu.ac.kr}{#1}}

\author{%
  \textbf{Jaeik Kim}\textsuperscript{\ensuremath{\dagger},\P}\enspace
  \textbf{Woojin Kim}\textsuperscript{\P}\enspace
  \textbf{Jihwan Hong}\textsuperscript{\P}\enspace
  \textbf{Yejoon Lee}\textsuperscript{\P}\enspace
  \textbf{Sieun Hyeon}\enspace
  \textbf{Mintaek Lim}\\[1pt]
  \textbf{Yunseok Han}\enspace
  \textbf{Dogeun Kim}\enspace
  \textbf{Hoeun Lee}\enspace
  \textbf{Hyeonggeun Kim}\enspace
  \textbf{Jinhyeok Kim}\enspace
  \textbf{Jaeyoung Do}\textsuperscript{\S}\\[3pt]
  AIDAS Laboratory, Seoul National University\\[2pt]
  {\small \textsuperscript{\ensuremath{\dagger}} Project lead\quad
    \textsuperscript{\P} Core contributors\quad
    \textsuperscript{\S} Corresponding author}\\[3pt]
  {\footnotesize\ttfamily
    \{\authorid{jake630}, \authorid{wjk9904}, \authorid{csjihwanh}, \authorid{leeyejoon},
      \authorid{zxc2692}, \authorid{victorlim}\}@snu.ac.kr}\\[-1pt]
  {\footnotesize\ttfamily
    \{\authorid{qicher}, \authorid{kdg6245}, \authorid{hoeunlee}, \authorid{khenrydragon},
      \authorid{jhkim00}, \authorid{jaeyoung.do}\}@snu.ac.kr}\\[7pt]
  {\small
    \href{https://dynin.ai/omni/}{\faHome~Project Page}\quad
    \href{https://github.com/AIDASLab/Dynin-Omni}{\faGithub~Code}\quad
    \href{https://huggingface.co/snu-aidas/Dynin-Omni}{\faDatabase~Model}\quad
    \href{https://huggingface.co/spaces/AIDAS-Lab/Dynin-Omni}{\faTv~Demo}%
  }
}

\hypersetup{
  hidelinks,
  pdftitle={Dynin-Omni: Omnimodal Unified Large Diffusion Language Model},
  pdfauthor={Jaeik Kim, Woojin Kim, Jihwan Hong, Yejoon Lee, Sieun Hyeon, Mintaek Lim,
             Yunseok Han, Dogeun Kim, Hoeun Lee, Hyeonggeun Kim, Jinhyeok Kim, Jaeyoung Do}
}

\begin{document}

\maketitle
\suppressfloats[t]

\begin{abstract}
Omnimodal unified models aim to understand text, images, video, and speech and generate coordinated text, image, and speech outputs. The challenge is to {\color{black} extend a single backbone to} %
new modalities efficiently while preserving {\color{black}established capabilities and enabling coordinated cross-modal generation.}
We present \method{}, an open-source 8B-scale native omnimodal masked-diffusion model that unifies these capabilities within a single token-level backbone under a shared masked-token prediction objective.
We propose a three-stage training recipe for effective omnimodal adaptation that first aligns new modalities without replaying the full task mixture, then applies modality-disentangled merging to recover inherited capabilities. Scheduled padding supervision delays termination learning until semantic grounding is established, enabling flexible-length generation.
Moreover, \method{} is the first masked-diffusion model to natively generate interleaved text, image, and speech within a single reverse diffusion process, enabling mutual conditioning and joint refinement without modality-wise orchestration.
Across 20 benchmark settings covering text, image, video, speech, and joint omnimodal generation, \method{} outperforms prior unified systems on a broad range of directly comparable benchmarks while remaining competitive with modality-specific experts.
Stage-wise and controlled analyses show {\color{black}capability recovery} %
and {\color{black}improved }%
length control, while native joint decoding {\color{black}approaches the quality of the best evaluated two-stage generation configuration with a lower sampling budget.}
\end{abstract}

\section{Introduction}

\begin{figure}[t]
  \centering
  \includegraphics[width=\linewidth,pagebox=cropbox]{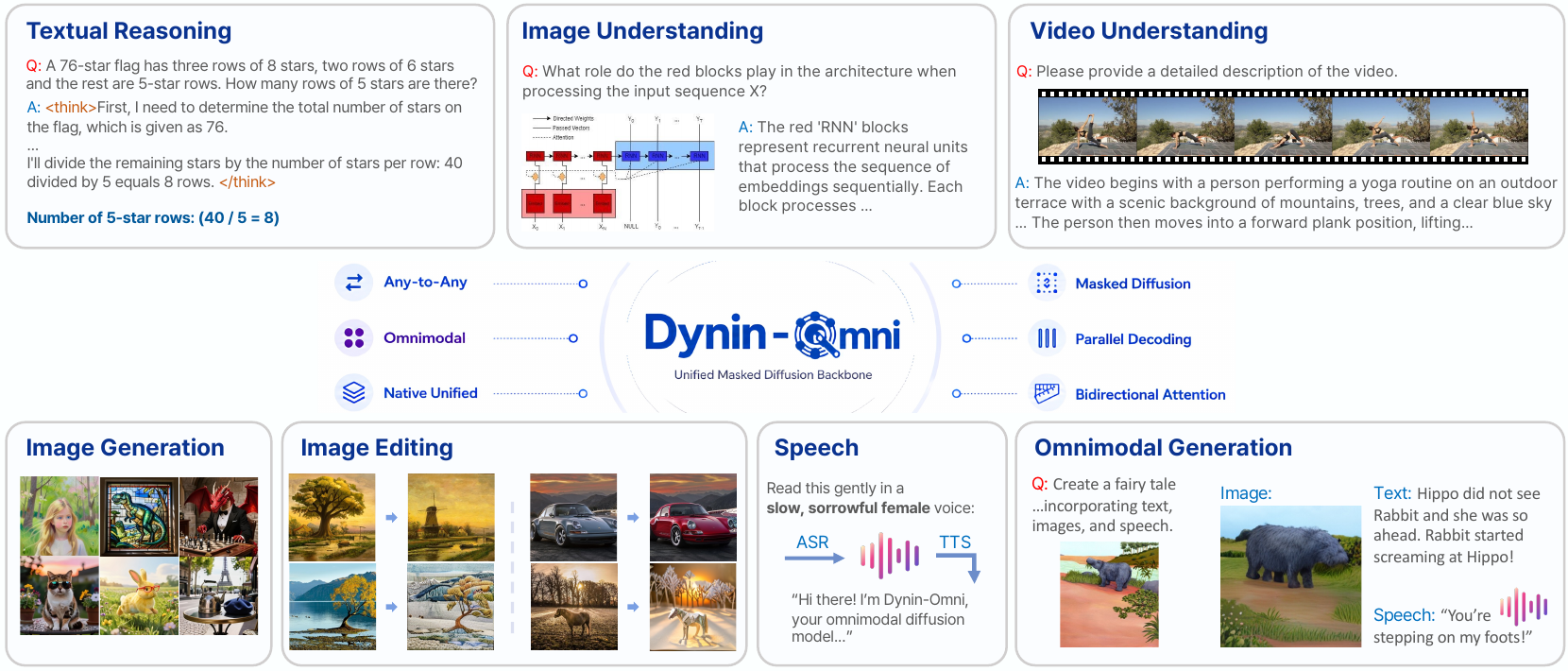}
  \caption{Overview of \method{}'s supported tasks across text, image, video, and speech.}
  \label{fig:overview}
\end{figure}

The pursuit of a \textit{universal interface} has driven the evolution of large-scale foundation models from text-only reasoning toward multimodal understanding and generation.
A natural next step is \emph{omnimodal} modeling, where text, vision, and speech serve as both inputs and generative outputs within a single system.
This is crucial for building agents that can understand and respond across modalities through a unified interface~\citep{next-gpt,anygpt,CoDi-2}.
In practice, the value of an omnimodal model lies in {\color{black}both strong performance on individual tasks and the ability to produce coordinated multimodal responses to mixed inputs.}

Fully autoregressive (AR) models excel at sequence modeling~\citep{yang2025qwen3,zeng2025glm}, but {\color{black}typically serialize} %
heterogeneous outputs {\color{black}and decode them} left-to-right~\citep{lu2024unifiedio2,wang2024emu3,anygpt}, fixing earlier outputs before later ones can inform them and limiting bidirectional context, parallel decoding, and reciprocal cross-modal refinement~\citep{zhao2025unified,shi2025muddit}.
Hybrid models combine AR reasoning with diffusion-based generation or modality-specific experts~\citep{zhou2024transfusion,xie2024showo1,xie2025showo2,cui2025emu3_5,bagle}, but typically use distinct objectives or modality-specific update rules that can restrict joint revision; compositional systems further require external generator handoffs~\citep{next-gpt,team2026hyperclova}.
Yet sharing a generative process is insufficient if learning new modalities erodes established capabilities.
This raises a central question: \textit{Can a single generative process {\color{black}extend to new modalities while preserving existing capabilities and enabling jointly generated}
outputs to refine one another?}

Masked diffusion~\citep{sahoo2024simple,nie2025large}
offers bidirectional context and parallel refinement without a fixed causal order{\color{black}, providing a natural foundation for interleaved outputs whose modalities mutually condition one another.}
Recent multimodal diffusion models have extended this paradigm beyond text~\citep{MMaDA,xin2025luminadimooomnidiffusionlarge,bethune2026trimodal,li2026omnidiffusion}.
Yet efficiently expanding a shared backbone to new modalities while preserving its existing capabilities remains challenging.
Addressing this training problem is central to achieving strong individual-task performance and native interleaved text--image--speech generation in the same model.

We introduce \method{}, an open-source masked-diffusion model to unify text and image understanding and generation, video understanding, and speech understanding and generation within a single 8B-scale token-level backbone.
As shown in Fig.~\ref{fig:overview}, \method{} represents text, image, video, and speech as discrete tokens in a shared vocabulary and models them under a unified masked-token prediction objective.
{\color{black}At inference time, this shared formulation accommodates modality-aware decoding schedules, including block-wise decoding for text and speech and fully parallel decoding for images.
The backbone and objective remain unified, while decoding schedules adapt to each modality.}
Moreover, \method{} is the first masked-diffusion model to natively generate interleaved text, image, and speech outputs within a single diffusion process, allowing them to mutually condition one another and be jointly refined without external generators or orchestration.

 We  adopt a  three-stage  progressive training recipe.  Stage~1 aligns newly introduced video and speech modalities  through text-centric supervision without replaying the full task mixture, using language as a semantic anchor. Stage~2 integrates new and  inherited capabilities through omnimodal supervised fine-tuning. Stage~3 scales task difficulty, resolution, and sequence length while introducing joint omnimodal generation.
Selective alignment can weaken inherited capabilities, while introducing termination supervision before new-modality content is grounded can cause premature stopping. We address these challenges with two complementary designs within the recipe.
{ \textit{Modality-disentangled merging} restores inherited  representations at the Stage~1-to-Stage~2 transition while retaining  newly aligned  tokens.  Native generative  expansion changes both  input embeddings and the output  head, so we interpolate the shared Transformer, preserve existing-token rows from the original model, and import new-token rows from the adapted model.  \textit{ Scheduled padding supervision}  defers EOS targets during initial  alignment and restores them during multimodal integration,  separating content grounding from termination learning.  Unlike Omni-Diffusion's attenuation-based masking~\citep{li2026omnidiffusion}, which reduces padding-token masking through a heuristically selected coefficient $\gamma$, our schedule controls when termination targets enter training during  modality adaptation, without searching over a continuous attenuation coefficient.}

Extensive evaluations show that \method{} achieves strong and balanced performance across textual reasoning, image understanding, generation and editing, video understanding, speech recognition and synthesis, and joint omnimodal generation.
Compared with prior unified models~\citep{xie2025showo2,team2026hyperclova,next-omni}, it achieves stronger results on a broad range of directly comparable  benchmarks while remaining competitive with modality-specific experts~\citep{yang2025qwen3,wu2025qwenimagetechnicalreport} and perception-centric models~\citep{Qwen2.5-Omni} on representative benchmarks.
Stage-wise evaluations show recovery of inherited text--image abilities, while controlled comparisons isolate the effects of {\color{black}modality-disentangled} merging and {\color{black}scheduled padding} supervision.
On Omni-StoryBench~\citep{omnistorybench}, \method{} surpasses orchestration-based systems without external image or speech generators.

Our contributions are summarized as follows:
\vspace{-0.5em}
\begin{itemize}[leftmargin=1.0em]

\item \textbf{Native omnimodal masked diffusion.}
{We introduce \method{}, an open-source 8B-scale omnimodal masked-diffusion model  unifying text and image understanding and generation, video understanding, and speech understanding and generation. It is the first masked-diffusion model to natively generate interleaved text, image, and speech within a single backbone and reverse diffusion process.}

\item \textbf{Training recipe for omnimodal  expansion and capability preservation. }
We propose  an effective three-stage recipe with modality-disentangled merging and scheduled  EOS supervision  to integrate new modalities,  recover inherited capabilities, and stabilize flexible-length generation.  Stage-wise  evaluations  track capability recovery, while controlled comparisons  support both components.

\item \textbf{Broad and balanced performance.}
Across 20 benchmark settings, \method{} outperforms prior unified systems on a broad range of directly comparable benchmarks and orchestration-based systems in joint generation, while remaining competitive with modality-specific experts  on representative benchmarks.

\end{itemize}

\section{\method{}}

\label{subsec:architecture}

\subsection{Native Omnimodal Token Space}
\label{subsec:omnimodal_token_space}

\begin{figure}[t]
    \centering
    \vspace{-0.3em}
\includegraphics[width=0.85\linewidth]{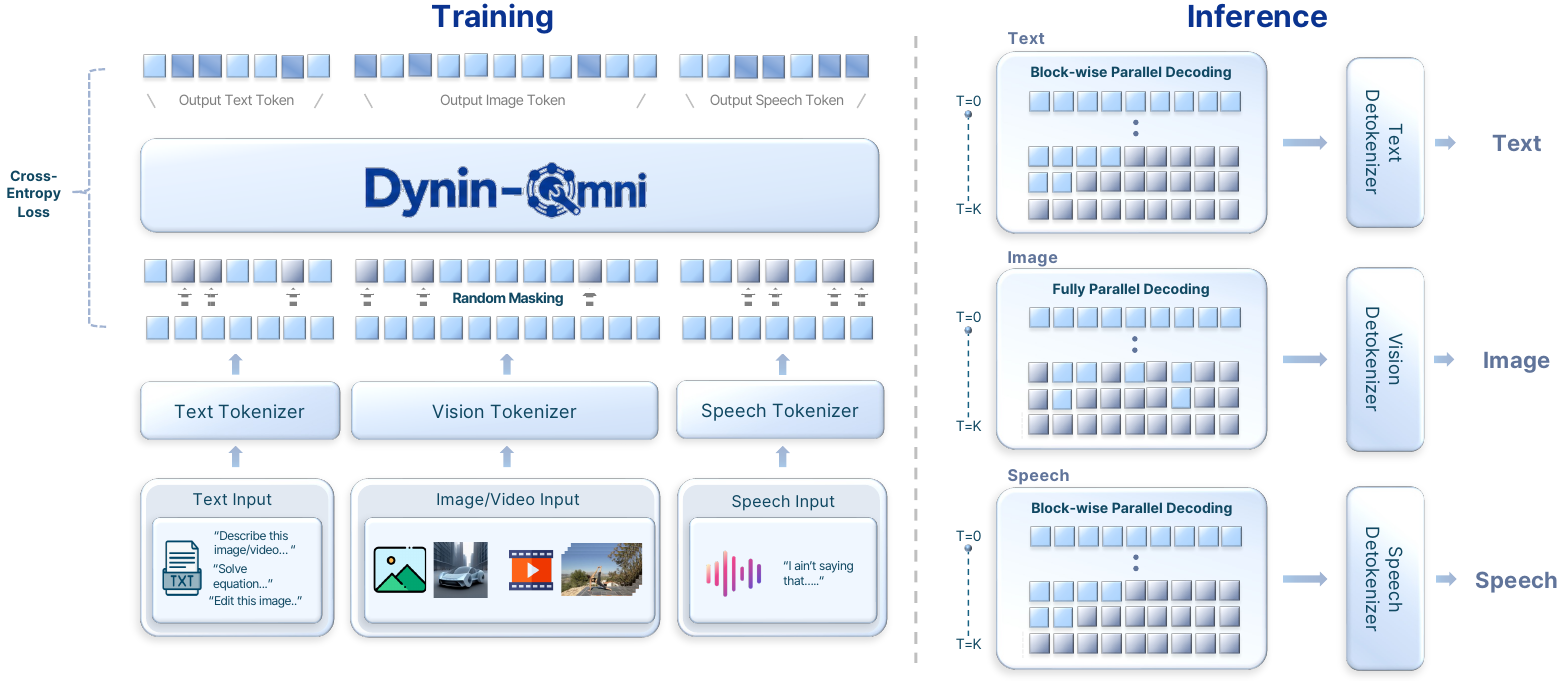}
    \vspace{-1.0em}
\caption{
Training and inference pipeline of \method{}.
All modalities are discretized into a shared token space and trained with a unified masked-diffusion objective.
At inference, \method{} applies modality-aware parallel decoding followed by deterministic detokenization.
}
    \label{fig:main_architecture}
\end{figure}

Building on LLaDA~\citep{nie2025large} and MMaDA~\citep{MMaDA}, \method{} models text, image, video input, and speech input/output as discrete tokens using a single masked-diffusion backbone with shared embeddings and a unified prediction head (Fig.~\ref{fig:main_architecture}).
The vocabulary $\mathcal{V}=\mathcal{V}_{\text{text}}\cup\mathcal{V}_{\text{vision}}\cup\mathcal{V}_{\text{speech}}$ extends MMaDA's text--image vocabulary with EMOVA speech tokens~\citep{chen2024emova}, while sampled video frames reuse the existing MAGViT-v2 vision tokens~\citep{MMaDA,xie2024showo1}.
All tokenizers and detokenizers are frozen, with detokenizers used only at inference; training therefore requires no reconstruction loss or modality-specific generative objective.
Tokenization details are provided in App.~\ref{sec:architecture}.

\subsection{Unified Masked Diffusion Objective}
\label{subsec:masked_diffusion_objective}
\method{} formulates omnimodal modeling as masked diffusion over a single discrete token sequence $\mathbf{x}_0 = (x_1, \ldots, x_L)$ drawn from the joint vocabulary $\mathcal{V}$.
The sequence may interleave tokens from multiple modalities and is processed by the backbone Transformer with bidirectional attention.

\paragraph{Training objective.}
Following masked diffusion (App.~\ref{app:discrete_diffusion})~\citep{he_diffusionbert_2023,sahoo2024simple,nie2025large}, we sample a masking ratio $t \sim \mathcal{U}(0,1)$ and independently replace each token in $\mathbf{x}_0$ with a special \texttt{[MASK]} token with probability $t$, yielding a corrupted sequence $\mathbf{x}_t$. The model is trained to reconstruct masked tokens conditioned on the visible context by minimizing the negative log-likelihood over masked positions:
\begin{equation}
\mathcal{L}(\theta)
=
\mathbb{E}_{\mathbf{x}_0,\, t,\, \mathbf{x}_t}
\left[
-\frac{1}{t}
\sum_{i=1}^{L}
\mathbb{I}[x_{t,i} = \texttt{[MASK]}]
\log p_\theta(x_{0,i} \mid \mathbf{x}_t)
\right].
\label{eq:mdm}
\end{equation}
This objective is shared across all modalities within the unified token space, without modality-specific heads or reconstruction objectives, enabling fully unified omnimodal training under a single objective (Eq.~\ref{eq:mdm}).
For flexible-length text and speech generation, the model also learns to predict a special \texttt{[EOS]} token, which serves as the padding and termination token~\citep{nie2025large}.

\paragraph{Tasks as masking patterns.}
A central consequence of this formulation is that task identity is specified entirely by the observed--masked partition of the sequence. Decomposing $\mathbf{x}_0 = [\mathbf{x}^{\mathrm{cond}}, \mathbf{x}^{\mathrm{tar}}]$ into observed conditioning tokens and masked target tokens, every task reduces to
\begin{equation}
p_\theta(\mathbf{x}^{\mathrm{tar}}\mid \mathbf{x}^{\mathrm{cond}})
=
\prod_{i:m_i=1}
p_\theta(x_{0,i}\mid \mathbf{x}_{\mathbf{m}}),
\end{equation}
where $\mathbf{m}\in\{0,1\}^{L}$ is a binary task mask ($m_i=1$ for target, $m_i=0$ for observed). For example, Visual-Question-Answering (VQA) observes image and question tokens while masking the answer ($\mathbf{x}_0=[\mathbf{x}^{I},\mathbf{x}^{Q},\mathbf{x}^{A}]$); text-to-image generation observes text and masks image tokens ($[\mathbf{x}^{T},\mathbf{x}^{I}]$). Modality transfer, understanding, and generation thus all collapse to a single conditional masked prediction problem under the same backbone and denoising objective (Tab.~\ref{tab:global_templates}). 
{For joint generation, interleaved text, image, and speech output tokens belong to the target partition, allowing predictions to condition on the intermediate tokens of other generated modalities along the shared denoising trajectory.}

\subsection{Omnimodal Inference}
\label{subsec:omnimodal_inference}

At inference time, \method{} performs understanding and generation by initializing target tokens as \texttt{[MASK]} and iteratively predicting masked positions conditioned on the observed context.
Although the backbone and reverse-diffusion process are shared, decoding is modality-aware: text and speech use block-wise refinement to accommodate their sequential structure, while image tokens are refined fully in parallel.
Confidence-based remasking revisits uncertain tokens within each schedule~\citep{Chang_2022_CVPR,nie2025large,dream2025}.
For faster inference, we combine KV-cache approximation and confidence-based parallel decoding in a compiled implementation~\citep{wu2026fastdllm,ma2025dinferefficientinferenceframework}, reaching 183.7 tokens per second on GSM8K with a single A100 GPU.
Configurations and further efficiency results are provided in App.~\ref{app:additional_results}.

\section{Efficient and Stable Omnimodal Training}
\label{sec:training_recipe}

\subsection{Overall Training Paradigm}
\label{subsec:training_overview}

To expand the model without replaying the full omnimodal training mixture from the outset, we first focus on aligning newly introduced modalities. This selective training can weaken inherited capabilities, while learning when to terminate before new-modality content is grounded can lead to prematurely short outputs.
We address these challenges with a three-stage recipe (Fig.~\ref{fig:training_stages_main}).
Stage~1 aligns video and speech through text-centric supervision~\citep{chen2024emova}, while deferring termination supervision.
Stage~2 merges the adapted model with the original backbone and performs omnimodal supervised fine-tuning, restoring termination supervision after new-modality alignment.
Stage~3 scales task difficulty, resolution, and sequence length while introducing joint omnimodal generation.
Modality-disentangled merging helps retain inherited capabilities, while scheduled padding supervision separates new-modality alignment from flexible-length learning.
Data and training details are in Apps.~\ref{sec:data_composition} and~\ref{sup:training_recipes}, with stage-wise results in \S\ref{subsec:training_recipe_analysis}. {Reduced-data controls for training order, EOS schedules, and post-merging learning are in App.~\ref{app:adaptation-controls}.}

\subsection{Modality-Disentangled Model Merging}
\label{subsec:model_merging}

\paragraph{Challenge.}
Stage~1 aligns new modalities without replaying the full omnimodal mixture, risking catastrophic forgetting of inherited capabilities~\citep{kirkpatrick2017overcoming}.
{Prior multimodal model-merging work largely targets  understanding capabilities: NaiveMC reuses modality-specific encoders and averages language-model weights~\citep{chen2024naivemc}, while MMER and OptMerge  study multimodal  expert merging~\citep{li2025mmer,wei2026optmerge}.  Native generative  expansion introduces a distinct interface issue: extending the  vocabulary changes both input embeddings and the output head.  Merging the original backbone $\theta^{(0)}$ and Stage~1 model $\theta^{(1)}$ therefore requires an explicit rule for original and newly added token rows  alongside  shared Transformer parameters.}
\begin{figure}
    \centering
    \includegraphics[width=.9\linewidth]{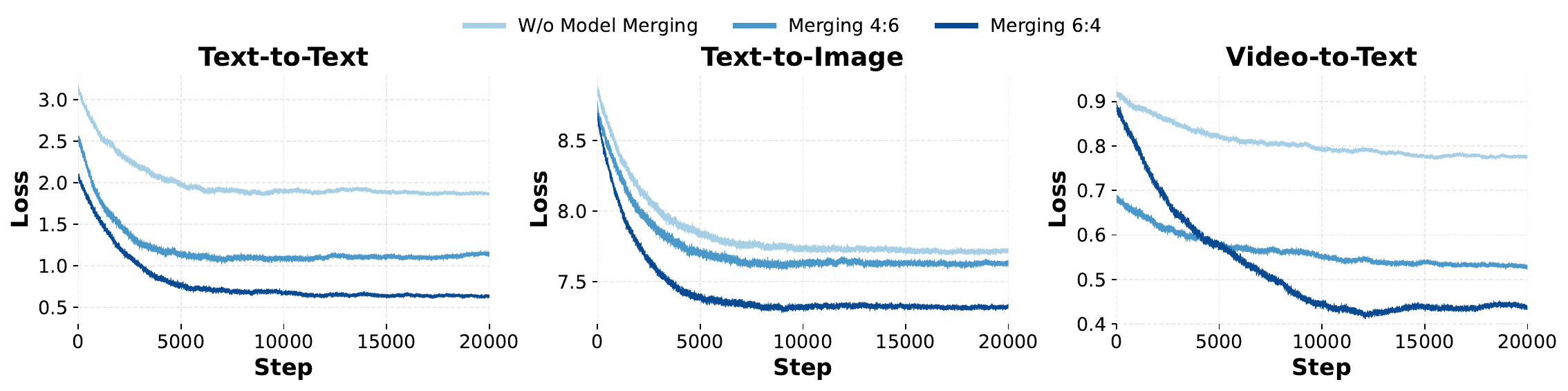}
\caption{
Stage-2 training loss across original (text-to-text and text-to-image) and new (video-to-text) tasks, where modality-disentangled merging accelerates convergence and reduces final loss.
}
    \label{fig:merging_loss}
\end{figure}

\paragraph{Method.}
At the transition to Stage~2, we apply \emph{modality-disentangled model merging} by separating vocabulary-independent and token-indexed parameters.
Let $\theta=\{\theta_{\mathrm{shared}},\theta_{\mathrm{vocab}}\}$ denote the Transformer parameters and the token embeddings and output head, respectively.
We interpolate the shared parameters:
\begin{equation}
\theta_{\mathrm{shared}}^{\mathrm{merge}}
=
(1-\alpha)\theta_{\mathrm{shared}}^{(0)}
+
\alpha\theta_{\mathrm{shared}}^{(1)}.
\label{eq:merging}
\end{equation}
For vocabulary-dependent parameters, existing token rows $i\in\mathcal{V}^{(0)}$ come from $\theta_{\mathrm{vocab}}^{(0)}$, while new-token rows $i\in\mathcal{V}^{\mathrm{new}}$ come from $\theta_{\mathrm{vocab}}^{(1)}$.
We use $\alpha=0.4$.
The merged initialization lowers Stage~2 loss (Fig.~\ref{fig:merging_loss}).
After matched Stage~2 training, it also improves inherited and new-modality benchmarks (Tab.~\ref{tab:training_recipe_effects}(b)).
Alternative merging strategies and interpolation ratios are analyzed in \S\ref{subsec:training_recipe_analysis}.

\subsection{Scheduled Padding Supervision}
\label{subsec:scheduled_padding}

\begin{figure}[t]
    \centering
    \begin{subfigure}[t]{0.52\linewidth}
        \centering
        \includegraphics[width=.87\linewidth]{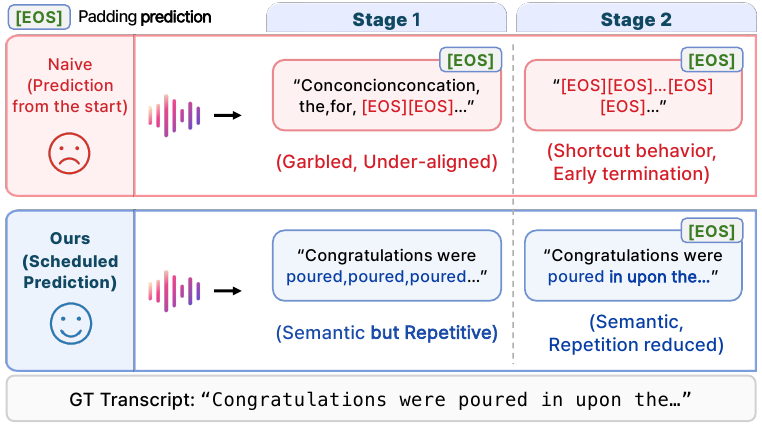}
        \caption{Qualitative effect of scheduled \texttt{[EOS]} supervision.}
        \label{fig:padding_effect_qual}
    \end{subfigure}
    \hfill
    \begin{subfigure}[t]{0.445\linewidth}
        \centering
        \includegraphics[width=.87\linewidth]{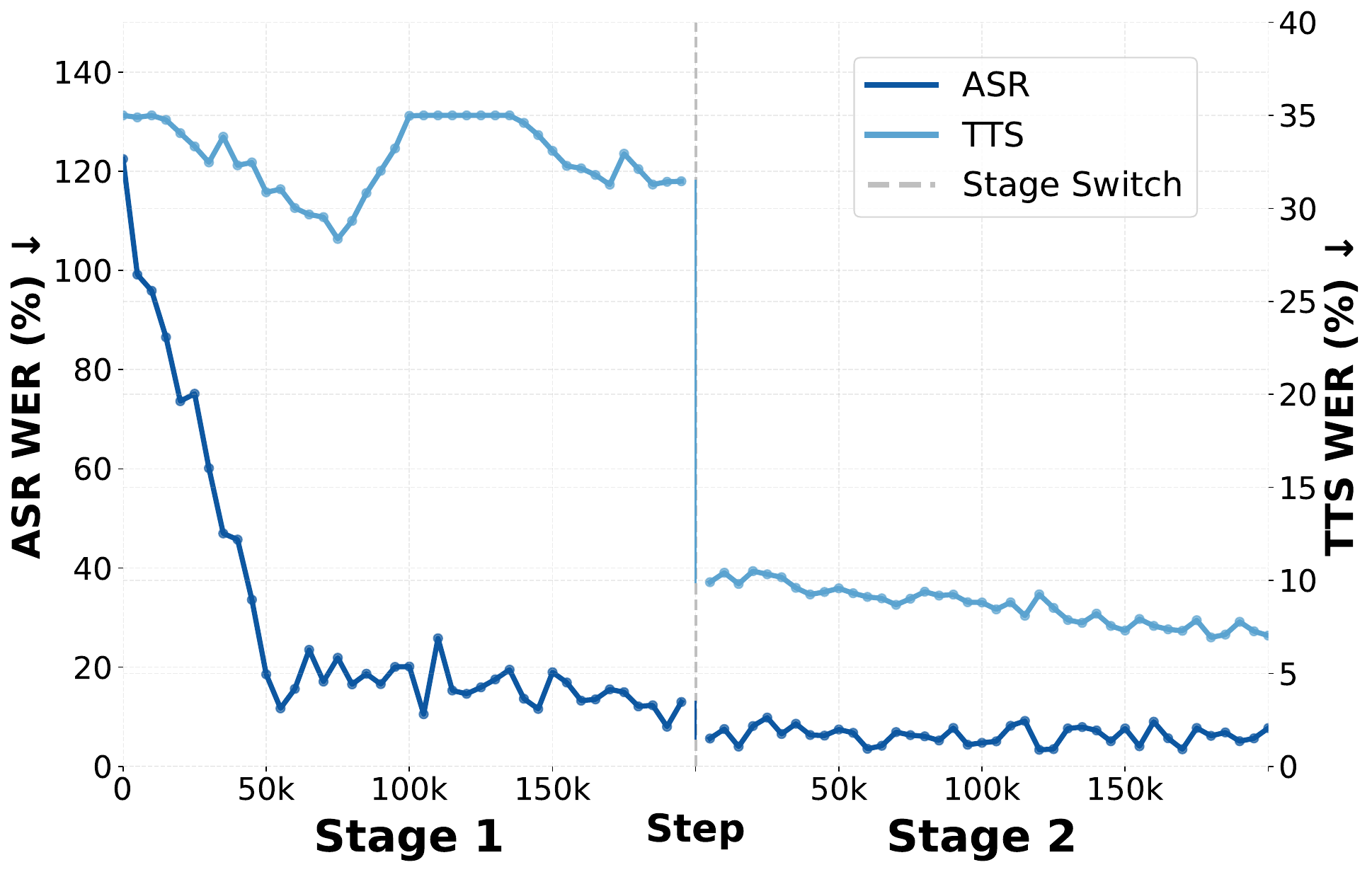}
        \caption{ASR and TTS WER across training stages. }
        \label{fig:padding_effect_wer}
    \end{subfigure}
\caption{
Effect of scheduled padding prediction.
(a) Scheduled padding (\texttt{[EOS]}) supervision avoids early shortcut behavior, and Stage~1 repetition is rapidly stabilized once \texttt{[EOS]} prediction is enabled in Stage~2.
(b) Adaptation stabilizes after Stage~2, confirming improved flexible-length generation.
}

    \label{fig:stage_padding}
\end{figure}

\paragraph{Challenge.}
Flexible-length generation in masked diffusion uses explicit length control~\citep{nie2025large,bie2025llada2,you2025lladav}.
In our formulation, the padding token is tied to \texttt{[EOS]} and therefore determines output termination. { Omni-Diffusion~\citep{li2026omnidiffusion}  reduces  padding-token masking  through an attenuation coefficient $\gamma$,  an additional hyperparameter for heuristic selection or search. Our stage-based schedule instead controls when termination  targets enter training,  without requiring a search over a continuous attenuation coefficient. We study this choice using matched runs that hold initialization, training data, and update counts fixed while varying  \texttt{[EOS]} supervision.}
These controls reveal a stage-dependent trade-off: enabling termination supervision before new-modality grounding causes premature stopping, whereas withholding it after alignment causes repetition and decoding-cap hits (Fig.~\ref{fig:padding_effect_qual}; Tab.~\ref{tab:training_recipe_effects}(c)).
Reduced-data whole-path controls also favor Off$\rightarrow$On over AlwaysOn and attenuation$\rightarrow$On (Tab.~\ref{tab:main-eos-comparison}; App.~\ref{app:adaptation-controls}).
Effective modality expansion therefore benefits from learning semantic alignment before output termination.

\paragraph{Method.}
We implement scheduled padding supervision by applying a stage-dependent weight to each token's contribution to the unified masked-diffusion objective in Eq.~\ref{eq:mdm}:
\begin{equation}
w_i^{(s)}=
\begin{cases}
0, & s=1 \ \text{and}\ x_{0,i}=\texttt{[EOS]},\\
1, & \text{otherwise}.
\end{cases}
\end{equation}
Thus, Stage~1 retains supervision for content tokens while deferring termination targets during new-modality alignment.
From Stage~2 onward, \texttt{[EOS]} supervision is restored to learn flexible-length generation under the same backbone and objective.
Deferring \texttt{[EOS]} supervision reduces premature termination in Stage~1; restoring it in Stage~2 improves ASR/TTS WER and reduces repetition and decoding-cap hits (Fig.~\ref{fig:padding_effect_wer}; Tab.~\ref{tab:training_recipe_effects}(c)).

\section{Evaluation}
\label{sec:evaluation}

We evaluate \method{} across textual reasoning, image understanding, generation, and editing, video understanding, speech recognition and synthesis (ASR/TTS), and joint omnimodal generation.
We include three groups of baselines: (i)~modality-specific experts, (ii)~perception-centric omnimodal models, and (iii)~unified models.
Among them, HyperCLOVAX~\citep{team2026hyperclova} and NExT-OMNI~\citep{next-omni} are the most direct baselines, as they support unified omnimodal understanding and generation.
Baseline comparison is provided in App.~\ref{sup:rw}.

\begin{table*}[t]
\centering
\tiny
\caption{Evaluation on reasoning and understanding benchmarks across text, image, and video.
{\setlength{\fboxsep}{0.2pt}\colorbox{gray!10}{Gray}} indicates unsupported modalities; ``--'' denotes supported but unreported or officially unused results.
$^{*}$~Video generation support.
$^{\dagger}$~Native speech input and output.
$^{\ddagger}$~Reproduced results.}
\label{tab:understanding}
\setlength{\tabcolsep}{2.75pt}
\renewcommand{\arraystretch}{0.9}
\begin{adjustbox}{max width=\linewidth}
\begin{tabular}{l cccc cccc cccc}
\toprule
& \multicolumn{4}{c}{\textbf{Text$\uparrow$}} 
& \multicolumn{4}{c}{\textbf{Image Understanding$\uparrow$}} 
& \multicolumn{4}{c}{\textbf{Video Understanding$\uparrow$}} \\
\cmidrule(lr){2-5} \cmidrule(lr){6-9} \cmidrule(lr){10-13}
\textbf{Model}
& MMLU & GSM8K & MATH & GPQA
& POPE & MME-P & MMMU & MMBench
& MVBench & Video-MME & NExT-QA & TempCompass \\
\midrule
\multicolumn{13}{l}{\textbf{Modality-specific Experts}} \\
\midrule
Llama-3-8B~\citep{grattafiori2024llama3}
  & 66.6 & 55.3 & 20.5 & 25.8
  & \cellcolor{gray!10} & \cellcolor{gray!10} & \cellcolor{gray!10} & \cellcolor{gray!10}
  & \cellcolor{gray!10} & \cellcolor{gray!10} & \cellcolor{gray!10} & \cellcolor{gray!10} \\
{LLaDA-8B-Base$^{\ddagger}$}~\citep{nie2025large}
  & {65.9} & {71.3} & {30.0} & {31.5}
  & \cellcolor{gray!10} & \cellcolor{gray!10} & \cellcolor{gray!10} & \cellcolor{gray!10}
  & \cellcolor{gray!10} & \cellcolor{gray!10} & \cellcolor{gray!10} & \cellcolor{gray!10} \\
{LLaDA-8B-Instruct$^{\ddagger}$}~\citep{nie2025large}
  & {65.4} & {68.7} & {29.8} & {31.8}
  & \cellcolor{gray!10} & \cellcolor{gray!10} & \cellcolor{gray!10} & \cellcolor{gray!10}
  & \cellcolor{gray!10} & \cellcolor{gray!10} & \cellcolor{gray!10} & \cellcolor{gray!10} \\
Qwen3-8B~\citep{yang2025qwen3}
  & 76.9 & 89.8 & 60.8 & 44.4
  & \cellcolor{gray!10} & \cellcolor{gray!10} & \cellcolor{gray!10} & \cellcolor{gray!10}
  & \cellcolor{gray!10} & \cellcolor{gray!10} & \cellcolor{gray!10} & \cellcolor{gray!10} \\
InternVL3-8B~\citep{zhu2025internvl3}
  & -- & -- & -- & --
  & 90.4 & 1748.4 & 62.2 & 83.6
  & 73.2 & 66.0 & 82.0$^{\ddagger}$ & 70.4$^{\ddagger}$ \\
Ovis2-8B~\citep{lu2024ovis}
  & -- & -- & -- & --
  & 88.6 & 1628.1 & 57.4 & 84.8
  & \cellcolor{gray!10} & \cellcolor{gray!10} & \cellcolor{gray!10} & \cellcolor{gray!10} \\
Qwen2.5-VL-7B~\citep{Qwen2.5-VL}
  & -- & -- & -- & --
  & 85.9 & 1698.1 & 58.0 & 83.2
  & 69.6 & 65.1 & -- & -- \\
LLaDA-V~\citep{you2025lladav}
  & -- & -- & -- & --
  & -- & 1507.0 & 48.6 & 82.9
  & 53.1 & 56.4 & -- & -- \\
LLaVA-OV~\citep{li2024llavaonevision}
  & -- & -- & -- & --
  & -- & 1580.0 & 48.8 & 80.8
  & 56.7 & 58.2 & 79.4 & 64.8 \\
InternVL2.5-8B~\citep{chen2024internvl}
  & -- & -- & -- & --
  & 90.6 & -- & 56.0 & 83.2
  & 72.0 & 64.2 & 84.1$^{\ddagger}$ & 67.7$^{\ddagger}$ \\
\midrule
\multicolumn{13}{l}{\textbf{Perception-centric}} \\
\midrule
OpenOmni~\citep{luo2025openomni}
  & -- & -- & -- & --
  & 86.4$^{\ddagger}$ & 1536.9$^{\ddagger}$ & 46.7 & 76.2
  & \cellcolor{gray!10} & \cellcolor{gray!10} & \cellcolor{gray!10} & \cellcolor{gray!10} \\
OmniVinci~\citep{ye2025omnivinci}
  & -- & -- & -- & --
  & 89.5$^{\ddagger}$ & 1651.0$^{\ddagger}$ & 49.7 & --
  & 70.6 & 68.2 & -- & -- \\
Baichuan-Omni-1.5~\citep{li2025baichuan}
  & 72.2 & -- & -- & --
  & 87.3$^{\ddagger}$ & 1632.1$^{\ddagger}$ & 53.9 & 85.6
  & 63.7 & 60.1 & -- & -- \\
MiniCPM-o-4.5~\citep{yao2024minicpm}
  & 77.0 & 94.5 & 77.0 & --
  & -- & -- & 67.6 & 87.6
  & -- & 70.4 & -- & -- \\
Qwen2.5-Omni-7B~\citep{Qwen2.5-Omni}
  & 71.8 & 88.7 & 71.5 & 30.8
  & 86.8$^{\ddagger}$ & 1481.3$^{\ddagger}$ & 59.2 & 81.8
  & 70.3 & 64.3 & 76.4$^{\ddagger}$ & 69.3$^{\ddagger}$ \\
\midrule
\multicolumn{13}{l}{\textbf{Unified}} \\
\midrule
Chameleon-7B~\citep{team2024chameleon}
  & 52.1 & 41.6 & 11.5 & --
  & -- & -- & 28.4 & 35.7
  & \cellcolor{gray!10} & \cellcolor{gray!10} & \cellcolor{gray!10} & \cellcolor{gray!10} \\
SEED-X~\citep{ge2024seed}
  & -- & -- & -- & --
  & 84.2 & 1435.7 & 35.6 & 70.1
  & \cellcolor{gray!10} & \cellcolor{gray!10} & \cellcolor{gray!10} & \cellcolor{gray!10} \\
Emu3$^{*}$~\citep{wang2024emu3}
  & -- & -- & -- & --
  & 85.2 & -- & 31.6 & 58.5
  & -- & -- & -- & -- \\
MMaDA~\citep{MMaDA}
  & 68.4 & 73.4 & 36.0 & 28.4
  & 86.1 & 1410.7 & 30.2 & 68.5
  & \cellcolor{gray!10} & \cellcolor{gray!10} & \cellcolor{gray!10} & \cellcolor{gray!10} \\
Fudoki~\citep{wang2025fudoki}
  & -- & -- & -- & --
  & 86.1 & 1485.4 & 34.3 & 73.9
  & \cellcolor{gray!10} & \cellcolor{gray!10} & \cellcolor{gray!10} & \cellcolor{gray!10} \\
Janus-Pro~\citep{chen2025januspro}
  & -- & -- & -- & --
  & 87.4 & 1567.1 & 41.0 & 79.2
  & \cellcolor{gray!10} & \cellcolor{gray!10} & \cellcolor{gray!10} & \cellcolor{gray!10} \\
BAGEL~\citep{bagle}
  & -- & -- & -- & --
  & -- & 1687.0 & 55.3 & 85.0
  & \cellcolor{gray!10} & \cellcolor{gray!10} & \cellcolor{gray!10} & \cellcolor{gray!10} \\
Lumina-DiMOO~\citep{xin2025luminadimooomnidiffusionlarge}
  & -- & -- & -- & --
  & 87.4 & 1534.2 & 58.6 & 83.1
  & \cellcolor{gray!10} & \cellcolor{gray!10} & \cellcolor{gray!10} & \cellcolor{gray!10} \\
Show-o2$^{*}$~\citep{xie2025showo2}
  & 70.7 & 75.3 & -- & 31.5
  & -- & 1620.5 & 48.9 & 79.3
  & 55.8 & 57.4 & 79.0 & -- \\
NExT-OMNI$^{*,\dagger}$~\citep{next-omni}
  & -- & -- & -- & --
  & 87.4 & 1537.8 & 43.7 & 78.9
  & -- & -- & -- & -- \\
HyperCLOVAX$^{\dagger}$~\citep{team2026hyperclova}
  & 75.7 & 87.3 & 43.4$^{\ddagger}$ & 28.6$^{\ddagger}$
  & 79.8$^{\ddagger}$ & 1307.0$^{\ddagger}$ & 31.0$^{\ddagger}$ & 65.9$^{\ddagger}$
  & 49.5$^{\ddagger}$ & 58.2 & -- & 58.3$^{\ddagger}$ \\
\rowcolor{lightblue}
\textbf{\method{}$^{\dagger}$ (Ours)}
  & \nipsver{75.9} & \nipsver{83.7} & \nipsver{47.4} & \nipsver{37.9}
  & \nipsver{88.0} & \nipsver{1672.9} & \nipsver{52.8} & \nipsver{74.7}
  & \nipsver{66.4} & \nipsver{63.4} & \nipsver{81.3} & \nipsver{66.7} \\
\bottomrule
\end{tabular}
\end{adjustbox}
\end{table*}

\subsection{Main Results}
\label{subsec:main_results}

Tab.~\ref{tab:understanding} and~\ref{tab:gen_speech} summarize representative benchmarks across reasoning, image understanding and synthesis, video understanding, and ASR/TTS.
Detailed inference configurations are provided in App.~\ref{app:inference_configs}.
Extended benchmark results, direct comparisons with text-diffusion and diffusion vision--language models, and evaluation uncertainty including perceptual TTS quality are provided in App.~\ref{app:additional_results}.

\paragraph{Comparison with modality-specific experts.}
Modality-specific experts achieve the strongest peak performance in their respective domains, including Qwen3-8B~\citep{yang2025qwen3} for text reasoning, InternVL-series models~\citep{chen2024internvl,zhu2025internvl3} for multimodal understanding, Qwen-Image~\citep{wu2025qwenimagetechnicalreport} for image synthesis, and dedicated ASR/TTS systems~\citep{Qwen3-ASR,hu2026qwen3tts}.
Nevertheless, \method{} remains competitive across domains: it approaches Qwen3-8B on MMLU, outperforms Qwen2.5-VL~\citep{Qwen2.5-VL} on POPE, approaches it on MME-P, matches Qwen-Image on GenEval, and achieves strong ASR and TTS performance.
While specialized models retain advantages on mathematical reasoning and SEED-TTS, \method{} provides balanced performance across these domains within a single backbone.

\paragraph{Comparison with perception-centric omnimodal models.}
Perception-centric omnimodal models, such as Qwen2.5-Omni~\citep{Qwen2.5-Omni}, MiniCPM-o~\citep{yao2024minicpm}, and OmniVinci~\citep{ye2025omnivinci}, achieve strong understanding and speech performance, but provide limited visual generation capability within a unified token-level generative process.
\method{} remains competitive with these models on reasoning, image understanding, video understanding, and speech benchmarks, while additionally supporting native image generation and editing under the same masked-diffusion objective.
This shows that \method{} extends omnimodal systems beyond perception-centric capability while retaining competitive understanding performance.

\begin{table}[t]
\centering
\tiny
\caption{Evaluation on image synthesis and speech benchmarks.
Notation follows Tab.~\ref{tab:understanding}.}
\label{tab:gen_speech}
\setlength{\tabcolsep}{6.75pt}
\renewcommand{\arraystretch}{0.9}
\begin{adjustbox}{max width=\linewidth}
\begin{tabular}{l ccc cc cc}
\toprule
& \multicolumn{3}{c}{\textbf{Image Generation / Editing$\uparrow$}} 
& \multicolumn{2}{c}{\textbf{ASR$\downarrow$}}
& \multicolumn{2}{c}{\textbf{TTS$\downarrow$}} \\
\cmidrule(lr){2-4} \cmidrule(lr){5-6} \cmidrule(lr){7-8}
\textbf{Model}
& GenEval & DPG-Bench & ImgEdit
& LibriSpeech\textsubscript{clean} & LibriSpeech\textsubscript{other}
& LibriSpeech\textsubscript{clean} & SEED-TTS \\
\midrule
\multicolumn{8}{l}{\textbf{Modality-specific Experts}} \\
\midrule
Seedream 3.0~\citep{gao2025seedream3}
  & 0.84 & 88.3 & \cellcolor{gray!10}
  & \cellcolor{gray!10} & \cellcolor{gray!10} 
  & \cellcolor{gray!10} & \cellcolor{gray!10} \\
GPT-Image 1~\citep{gptimage}
  & 0.84 & 85.2 & 4.28 
  & \cellcolor{gray!10} & \cellcolor{gray!10} 
  & \cellcolor{gray!10} & \cellcolor{gray!10} \\
Qwen-Image~\citep{wu2025qwenimagetechnicalreport}
  & 0.87 & 88.3 & 4.33 
  & \cellcolor{gray!10} & \cellcolor{gray!10} 
  & \cellcolor{gray!10} & \cellcolor{gray!10} \\
Flux.1 Kontext~\citep{labs2025flux1kontextflowmatching}
  & -- & -- & 3.59 
  & \cellcolor{gray!10} & \cellcolor{gray!10} 
  & \cellcolor{gray!10} & \cellcolor{gray!10} \\
Qwen2-Audio~\citep{Qwen2-Audio}
  & \cellcolor{gray!10} & \cellcolor{gray!10} & \cellcolor{gray!10} 
  & 1.6 & 3.6 
  & \cellcolor{gray!10} & \cellcolor{gray!10} \\
Qwen3-ASR~\citep{Qwen3-ASR}
  & \cellcolor{gray!10} & \cellcolor{gray!10} & \cellcolor{gray!10} 
  & 1.6 & 3.4 
  & \cellcolor{gray!10} & \cellcolor{gray!10} \\
MaskGCT~\citep{wang2024maskgct}
  & \cellcolor{gray!10} & \cellcolor{gray!10} & \cellcolor{gray!10} 
  & \cellcolor{gray!10} & \cellcolor{gray!10} 
  & 2.6 & 2.6 \\
Llasa-8B~\citep{ye2025llasa}
  & \cellcolor{gray!10} & \cellcolor{gray!10} & \cellcolor{gray!10} 
  & \cellcolor{gray!10} & \cellcolor{gray!10} 
  & 1.5 & 3.0 \\
Qwen3-TTS~\citep{hu2026qwen3tts}
  & \cellcolor{gray!10} & \cellcolor{gray!10} & \cellcolor{gray!10} 
  & \cellcolor{gray!10} & \cellcolor{gray!10} 
  & 6.6$^{\ddagger}$ & 1.2 \\
\midrule
\multicolumn{8}{l}{\textbf{Perception-centric}} \\
\midrule
EMOVA-7B~\citep{chen2024emova}
  & \cellcolor{gray!10} & \cellcolor{gray!10} & \cellcolor{gray!10} 
  & 4.0 & -- 
  & 3.6 & -- \\
OpenOmni~\citep{luo2025openomni}
  & \cellcolor{gray!10} & \cellcolor{gray!10} & \cellcolor{gray!10} 
  & 3.1 & 7.0 
  & 3.4 & -- \\
MiniCPM-o-4.5~\citep{yao2024minicpm}
  & \cellcolor{gray!10} & \cellcolor{gray!10} & \cellcolor{gray!10} 
  & 1.4 & 2.8 
  & -- & 2.4 \\
Qwen2.5-Omni-7B~\citep{Qwen2.5-Omni}
  & \cellcolor{gray!10} & \cellcolor{gray!10} & \cellcolor{gray!10} 
  & 1.8 & 3.4 
  & 7.0$^{\ddagger}$ & 2.3 \\
\midrule
\multicolumn{8}{l}{\textbf{Unified}} \\
\midrule
Emu3$^{*}$~\citep{wang2024emu3}
  & 0.66 & 81.6 & \cellcolor{gray!10}
  & \cellcolor{gray!10} & \cellcolor{gray!10} 
  & \cellcolor{gray!10} & \cellcolor{gray!10} \\
MMaDA~\citep{MMaDA}
  & 0.63 & 70.0 & \cellcolor{gray!10}
  & \cellcolor{gray!10} & \cellcolor{gray!10} 
  & \cellcolor{gray!10} & \cellcolor{gray!10} \\
Fudoki~\citep{wang2025fudoki}
  & 0.77 & 83.6 & \cellcolor{gray!10}
  & \cellcolor{gray!10} & \cellcolor{gray!10} 
  & \cellcolor{gray!10} & \cellcolor{gray!10} \\
Janus-Pro~\citep{chen2025januspro}
  & 0.80 & 84.2 & \cellcolor{gray!10}
  & \cellcolor{gray!10} & \cellcolor{gray!10} 
  & \cellcolor{gray!10} & \cellcolor{gray!10} \\
Show-o2$^{*}$~\citep{xie2025showo2}
  & 0.76 & 86.1 & \cellcolor{gray!10}
  & \cellcolor{gray!10} & \cellcolor{gray!10} 
  & \cellcolor{gray!10} & \cellcolor{gray!10} \\
Lumina-DiMOO~\citep{xin2025luminadimooomnidiffusionlarge}
  & 0.88 & 86.0 & --
  & \cellcolor{gray!10} & \cellcolor{gray!10} 
  & \cellcolor{gray!10} & \cellcolor{gray!10} \\
BAGEL~\citep{bagle}
  & 0.82 & 85.1 & 3.30 
  & \cellcolor{gray!10} & \cellcolor{gray!10} 
  & \cellcolor{gray!10} & \cellcolor{gray!10} \\
UniWorld-V1~\citep{lin2025uniworld}
  & 0.79 & -- & 3.30 
  & \cellcolor{gray!10} & \cellcolor{gray!10} 
  & \cellcolor{gray!10} & \cellcolor{gray!10} \\
LaViDa-O~\citep{li2025lavidao}
  & 0.77 & 81.8 & 3.81 
  & \cellcolor{gray!10} & \cellcolor{gray!10} 
  & \cellcolor{gray!10} & \cellcolor{gray!10} \\
NExT-OMNI$^{*,\dagger}$~\citep{next-omni}
  & 0.85 & 84.2 & \cellcolor{gray!10}
  & 3.1 & 7.0 
  & 3.1 & -- \\
HyperCLOVAX$^{\dagger}$~\citep{team2026hyperclova}
  & 0.64 & 60.8$^{\ddagger}$ & 3.83 
  & 1.9 & 4.5$^{\ddagger}$ 
  & 7.9$^{\ddagger}$ & 7.7$^{\ddagger}$ \\
\rowcolor{lightblue}
\textbf{\method{}$^{\dagger}$ (Ours)}
  & \nipsver{0.87} & \nipsver{85.5} & \nipsver{3.82}
  & \nipsver{2.1} & \nipsver{4.8} 
  & \nipsver{2.1} & \nipsver{4.3} \\
\bottomrule
\end{tabular}
\end{adjustbox}
\end{table}

\paragraph{Comparison with unified models.}
Unified models differ in both modality coverage and degree of architectural unification.
Compared with image--text unified models such as MMaDA~\citep{MMaDA} and BAGEL~\citep{bagle}, \method{} extends coverage to video understanding and speech input/output while remaining competitive or stronger on shared benchmarks, including MME-P, GenEval, and ImgEdit.
Compared with compositional omnimodal systems such as HyperCLOVAX-8B-Omni~\citep{team2026hyperclova}, which rely on external generators, \method{} achieves stronger performance on several shared benchmarks, while using a single native backbone.
These results show that native masked diffusion can scale to broader omnimodal modeling.

\begin{table*}[t]
\centering
\tiny
\caption{
Evaluation on omnimodal generation. Overall averages the six metrics, with 0--1 metrics scaled by 10.
Notation follows Tab.~\ref{tab:understanding}.}
\label{tab:omnimodal_generation}
\setlength{\tabcolsep}{1.5pt}
\renewcommand{\arraystretch}{0.95}

\begin{adjustbox}{max width=\linewidth}
\begin{tabular}{l c cc cccccc c}
\toprule
&
& \multicolumn{2}{c}{\textbf{Generator}}
& \multicolumn{2}{c}{\textbf{Text$\uparrow$}}
& \multicolumn{2}{c}{\textbf{Image$\uparrow$}}
& \multicolumn{2}{c}{\textbf{Speech$\uparrow$}}
& 
\\

\cmidrule(lr){3-4}
\cmidrule(lr){5-6}
\cmidrule(lr){7-8}
\cmidrule(lr){9-10}

\textbf{Orchestrator} & \textbf{Joint Gen.}
& {Image}
& {Speech}
& {BERTScore}
& {LLM Judge}
& {CLIP Sim.}
& {LLM Judge}
& {Metadata Acc.}
& {LLM Judge}
& \textbf{Overall$\uparrow$}
\\

\midrule
\multicolumn{11}{l}{\textbf{LLM}} \\
\midrule

InternVL3.5-4B~\citep{wang2025internvl3}
& $\times$
& Flux.1 K~\citep{labs2025flux1kontextflowmatching}
& Parler~\citep{parler}
& 0.87 & 7.44 & 0.76 & 5.43 & 0.55 & 8.06 & 7.12
\\

Qwen3.5-4B~\citep{qwen3.5}
& $\times$
& Flux.1 K~\citep{labs2025flux1kontextflowmatching}
& Parler~\citep{parler}
& 0.88 & 8.24 & 0.77 & 6.42 & 0.52 & 8.22 & 7.43
\\

\midrule
\multicolumn{11}{l}{\textbf{Omnimodal Perception}} \\
\midrule

EMOVA-7B~\citep{chen2024emova}
& $\times$
& Flux.1 K~\citep{labs2025flux1kontextflowmatching}
& --
& 0.87 & 7.78 & 0.78 & 4.73 & 0.52 & 8.23 & 7.06
\\

Qwen2.5-Omni-7B~\citep{Qwen2.5-Omni}
& $\times$
& Flux.1 K~\citep{labs2025flux1kontextflowmatching}
& --
& 0.88 & 7.97 & 0.76 & 5.57 & 0.54 & 8.07 & 7.24
\\

\midrule
\multicolumn{11}{l}{\textbf{Multimodal Unified}} \\
\midrule

Emu3$^{\*}$~\citep{wang2024emu3}
& $\times$
& --
& Parler~\citep{parler}
& 0.87 & 5.48 & 0.58 & 1.78 & 0.51 & 7.91 & 5.81
\\

MMaDA~\citep{MMaDA}
& $\times$
& --
& Parler~\citep{parler}
& 0.87 & 4.79 & 0.65 & 1.96 & 0.49 & 8.03 & 5.82
\\

\midrule
\multicolumn{11}{l}{\textbf{Omnimodal Unified}} \\
\midrule

AnyGPT~\citep{anygpt}
& $\times$
& --
& --
& 0.80 & 3.24 & 0.64 & 1.66 & 0.42 & 5.34 & 4.81
\\

HyperCLOVAX$^{\dagger}$~\citep{team2026hyperclova}
& $\times$
& --
& --
& 0.87 & 6.43 & 0.68 & 2.50 & 0.53 & 8.14 & 6.32
\\

\rowcolor{lightblue}
\textbf{\method{}$^{\dagger}$ (Ours)}
& $\checkmark$
& --
& --
& 0.92 & 6.78 & 0.86 & 6.40 & 0.61 & 8.70 & 7.63
\\

\bottomrule
\end{tabular}
\end{adjustbox}
\end{table*}

\subsection{Omnimodal Generation}
\label{subsec:omnimodal_generation}

We further evaluate \method{} on Omni-StoryBench~\citep{omnistorybench}, a story-grounded benchmark in which models generate coordinated narration, scene images, and spoken utterances from a given context.
Tab.~\ref{tab:omnimodal_generation} compares different approaches to constructing omnimodal generation systems.
The orchestration baselines employ strong external image and speech specialists, while existing unified models~\citep{MMaDA,anygpt,team2026hyperclova} generate different modalities through staged or separate inference calls.
In contrast, \method{} jointly generates interleaved text, image, and speech within a single masked-diffusion process without external generators.
Despite this fully native design, \method{} surpasses the strongest orchestration baseline in overall, with consistent performance after filtering against the omnimodal data used for training (App.~\ref{app:additional_results}).
These results establish native joint decoding as an effective alternative to orchestrated omnimodal generation.

\section{Analysis}
\label{sec:analysis}

\begin{table*}[t]
\centering
\caption{
Evaluation of the three-stage training recipe:
(a) benchmark performance across training stages;
(b) performance with and without merging after 10K Stage~2 updates; and
(c) the effect of \texttt{[EOS]} supervision within Stages~1 and~2.
``new'' denotes a capability absent from the backbone.
}
\label{tab:training_recipe_effects}

\tiny
\setlength{\tabcolsep}{3.5pt}
\renewcommand{\arraystretch}{1.05}

\begin{tabular*}{\linewidth}{
    @{\extracolsep{\fill}}
    lrrrrrr
    @{}
}
\multicolumn{7}{c}{
    (a) Benchmark performance across the three training stages.
} \\
\addlinespace[2pt]
\toprule
Checkpoint
& GSM8K$\uparrow$
& MME-P$\uparrow$
& GenEval$\uparrow$
& Video-MME$\uparrow$
& ASR WER$\downarrow$
& TTS WER$\downarrow$ \\
\midrule
Backbone
& 56.3
& 1410.7
& 0.63
& --
& --
& -- \\
Stage~1
& 4.2 (-52.1)
& 413.8 (-996.9)
& 0.12 (-0.51)
& 10.0 (new)
& 12.50 (new)
& 31.50 (new) \\
Stage~2
& 64.1 (+59.9)
& 1576.5 (+1162.7)
& 0.63 (+0.51)
& 52.0 (+42.0)
& 5.00 (-7.50)
& 4.17 (-27.33) \\
Stage~3
& \textbf{83.7 (+19.6)}
& \textbf{1672.9 (+96.4)}
& \textbf{0.87 (+0.24)}
& \textbf{63.4 (+11.4)}
& \textbf{2.10 (-2.90)}
& \textbf{2.10 (-2.07)} \\
\bottomrule
\end{tabular*}

\vspace{6pt}

\begin{tabularx}{\linewidth}{
    @{}l*{5}{>{\raggedleft\arraybackslash}X}@{}
}
\multicolumn{6}{c}{
    (b) Benchmark performance after 10K Stage~2 updates with and without modality-disentangled merging.
} \\
\addlinespace[2pt]
\toprule
& \multicolumn{3}{c}{\textbf{Original Modality}}
& \multicolumn{2}{c}{\textbf{New Modality}} \\
\cmidrule(lr){2-4}\cmidrule(lr){5-6}
Initialization
& GSM8K$\uparrow$
& POPE$\uparrow$
& MME-P$\uparrow$
& ASR WER$\downarrow$
& TTS WER$\downarrow$ \\
\midrule
No merge
& 18.0
& 50.0
& 820.0
& 10.50
& 20.00 \\
\rowcolor{lightblue}Modality-disentangled merging
& \textbf{38.0}
& \textbf{76.0}
& \textbf{1150.0}
& \textbf{6.46}
& \textbf{10.22} \\
\bottomrule
\end{tabularx}

\vspace{6pt}

\begin{tabularx}{\linewidth}{
    @{}ll*{6}{>{\raggedleft\arraybackslash}X}@{}
}
\multicolumn{8}{c}{
    \parbox{0.98\linewidth}{
        \centering
        (c) Speech and termination results after 10K updates with and without
        \texttt{[EOS]} supervision in each stage.
        \textit{Non-prem.}: first \texttt{[EOS]} at $\geq 80\%$ of reference length
        and before the decoding cap;
        \textit{Prem.}: earlier termination;
        \textit{Repeat}: three consecutive repeats of the same four-token sequence;
        \textit{Cap}: no \texttt{[EOS]} before the decoding limit.
    }
} \\
\addlinespace[2pt]
\toprule
Stage
& \texttt{[EOS]}
& ASR WER$\downarrow$
& TTS WER$\downarrow$
& Non-prem.$\uparrow$
& Prem.$\downarrow$
& Repeat$\downarrow$
& Cap$\downarrow$ \\
\midrule
1 & Enabled
& 131.80
& 55.00
& \textbf{35\%}
& 60\%
& \textbf{8\%}
& \textbf{5\%} \\
\rowcolor{lightblue}1 & {Disabled}
& \textbf{100.65}
& \textbf{35.07}
& 20\%
& \textbf{3\%}
& 60\%
& 77\% \\
\midrule
2 & Disabled
& 14.00
& 24.00
& 30\%
& \textbf{2\%}
& 55\%
& 68\% \\
\rowcolor{lightblue}2 & {Enabled}
& \textbf{6.46}
& \textbf{10.22}
& \textbf{94\%}
& 3\%
& \textbf{4\%}
& \textbf{3\%} \\
\bottomrule
\end{tabularx}

\end{table*}

\subsection{Training Recipe}
\label{subsec:training_recipe_analysis}

\paragraph{Stage-wise capability evolution.}
Tab.~\ref{tab:training_recipe_effects}(a) tracks capability evolution across training.
Stage~1 focuses adaptation on the new video and speech modalities, enabling efficient capability acquisition but causing forgetting of inherited text--image performance.
Stage~2 merges the adapted model with the original backbone, recovering these inherited capabilities while further improving video and speech.
Notably, even before the final scaling stage, Stage~2 already surpasses the backbone on GSM8K and MME-P and restores GenEval to its original level, showing that recovery does not come at the expense of the newly added modalities.
Stage~3 further improves all domains, with GSM8K, MME-P, and GenEval surpassing the original backbone despite adding video and speech.

\paragraph{Effect of modality-disentangled merging.}
Tab.~\ref{tab:training_recipe_effects}(b) isolates merging under matched data, optimizer, and update budgets.
As shown in Fig.~\ref{fig:merging_loss}, modality-disentangled merging lowers the initial Stage~2 loss and accelerates convergence across original and new-modality tasks.
After equal training, it also improves text, image, and speech results over training without merging (Tab.~\ref{tab:training_recipe_effects}(b)).
The ASR/TTS gains show that old-task recovery does not come at the expense of new-modality performance in this comparison.
The benefit therefore persists beyond initialization, helping Stage~2 consolidate both inherited and newly acquired capabilities.

\begin{figure*}[t]
\centering

\begin{minipage}[t]{0.485\linewidth}
    \vspace{0pt}
    \centering
    \includegraphics[width=\linewidth]{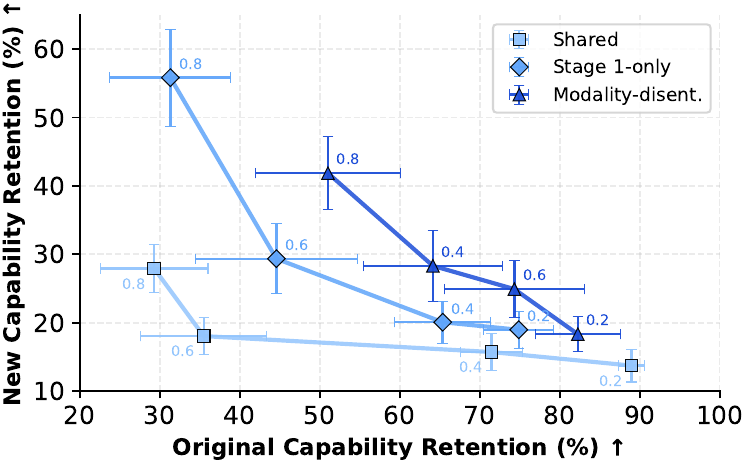}
    \captionof{figure}{
    Capability retention across merging strategies and interpolation ratios.
    }
    \label{fig:merging_strategies}
\end{minipage}
\hfill
\begin{minipage}[t]{0.485\linewidth}
    \vspace{0pt}
    \centering
    \includegraphics[width=\linewidth]{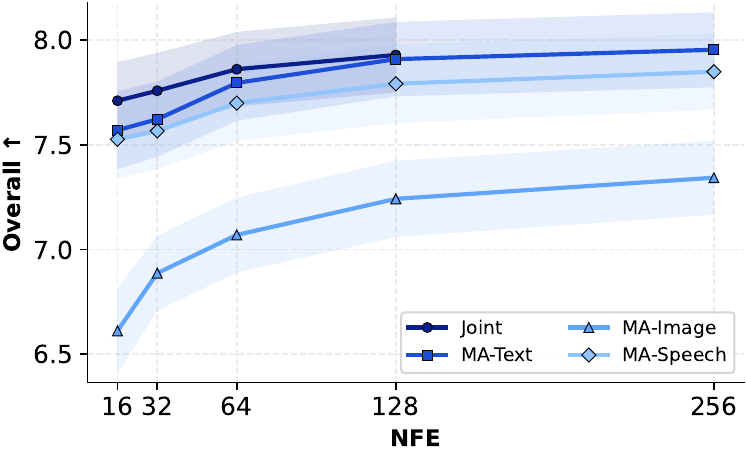}
    \captionof{figure}{
    Joint and modality-anchored omnimodal generation across NFE budgets. {Shaded bands show $\pm0.2$ standard deviations.}
    }
    \label{fig:omnimodal_generation}
\end{minipage}

\end{figure*}

\paragraph{Comparison of merging strategies.}
We compare modality-disentangled merging with two alternatives at the Stage~1-to-Stage~2 transition.
All three interpolate the shared Transformer parameters.
\textit{Shared Merging} interpolates existing-token rows and takes new-token rows from Stage~1, whereas \textit{Stage~1-Only} takes all vocabulary-dependent rows from Stage~1.
Complete formulations are provided in App.~\ref{sup:training_recipes}.
Fig.~\ref{fig:merging_strategies} sweeps $\alpha$ before Stage~2 training, comparing inherited-capability retention (GSM8K, POPE, and MME-P relative to the backbone) with new-modality retention (ASR and TTS relative to Stage~1).
Shared Merging retains neither group well, while Stage~1-Only favors new modalities at the expense of inherited text--image capabilities.
Modality-disentangled merging provides the best balance at $\alpha=0.4$ (backbone-to-Stage~1 ratio of $6{:}4$), showing the value of selecting token rows by their origin.
Across the sweep, the alternative token-row rules cannot match the same inherited/new-modality balance through $\alpha$ tuning alone.
{The matched post-training comparison in App.~\ref{app:adaptation-controls}, Table~\ref{tab:control-merge}, additionally tests whether the vocabulary-row advantage persists after integration. After the same 8,192 Stage~2 updates in the reduced-data 8B control, modality-disentangled merging has the best point estimates on all six reported metrics.}

\paragraph{Effect of scheduled padding supervision.}
Tab.~\ref{tab:training_recipe_effects}(c) isolates \texttt{[EOS]} supervision under matched data, optimizer, and update budgets; the comparisons also use the same initialization within each stage.
Our schedule disables \texttt{[EOS]} supervision in Stage~1 and enables it in Stage~2; disabling supervision does not prevent \texttt{[EOS]} generation at inference.

In Stage~1, disabling \texttt{[EOS]} supervision reduces WER and premature termination, but increases repetition and decoding-cap hits.
Enabling it in Stage~2 reduces WER, repetition, and cap hits while increasing non-premature termination.
Neither fixed supervision choice resolves both failure modes: early termination targets favor short outputs, whereas continued omission leaves length control undertrained.
These results support aligning new-modality content before introducing explicit length control.
Expanded merging controls and WER decompositions are provided in App.~\ref{app:additional_results}.

\begingroup
To compare  EOS supervision schedules, we  use a  reduced-data 8B  control with 256 speech--transcript pairs: 4,096 alignment updates followed by 2,048 multimodal integration updates, without model merging. The attenuation baseline  uses Omni-Diffusion's  rule ($\gamma=0.6$) during  alignment; all staged schedules  enable EOS supervision during  integration. As shown in Tab.~\ref{tab:main-eos-comparison}, Off$\rightarrow$On achieves the lowest ASR and TTS WER among the three staged schedules. Experimental details and additional controls are provided in App.~\ref{app:adaptation-controls}.

\begin{table}[t]
\begin{minipage}[t]{.49\linewidth}
\captionsetup{font=footnotesize}
\centering

\caption{{Whole-path comparison of EOS schedules.}}
\label{tab:main-eos-comparison}
\tiny
\setlength{\tabcolsep}{0pt}
\renewcommand{\arraystretch}{0.68}
\setlength{\aboverulesep}{0.5pt}
\setlength{\belowrulesep}{0.5pt}
\begin{tabular*}{\linewidth}{@{}>{\raggedright\arraybackslash}p{.46\linewidth}>{\raggedleft\arraybackslash}p{.27\linewidth}>{\raggedleft\arraybackslash}p{.27\linewidth}@{}}
\toprule
Schedule & ASR WER $\downarrow$ & TTS WER $\downarrow$ \\
\midrule
AlwaysOn & 71.29 & 16.41 \\
Attenuation$\rightarrow$On & 65.16 & 17.08 \\
\rowcolor{lightblue}Off$\rightarrow$On & {64.27} & {15.57} \\
\bottomrule
\end{tabular*}
\end{minipage}\hfill
\begin{minipage}[t]{.49\linewidth}
\captionsetup{font=footnotesize}

\centering
\caption{{Cross-modal attention intervention.}}
\label{tab:crossmodal-intervention}
\tiny
\setlength{\tabcolsep}{0pt}
\renewcommand{\arraystretch}{0.68}
\setlength{\aboverulesep}{0.5pt}
\setlength{\belowrulesep}{0.5pt}
\begin{tabular*}{\linewidth}{@{}>{\raggedright\arraybackslash}p{.24\linewidth}*{5}{>{\raggedleft\arraybackslash}p{.152\linewidth}}@{}}
\toprule
Condition & T-BERT $\uparrow$ & T-judge $\uparrow$ & I-CLIP $\uparrow$ & I-judge $\uparrow$ & S-WER $\downarrow$ \\
\midrule
Block text & 0.9108 & 7.160 & 0.8057 & 6.016 & 67.09 \\
Block image & 0.9222 & 7.223 & 0.7994 & 5.889 & 42.04 \\
Block speech & 0.9185 & 7.084 & 0.8185 & 6.152 & 62.64 \\
\rowcolor{lightblue}Full joint & {0.9342} & {7.629} & {0.8288} & {6.496} & {31.32} \\
\bottomrule
\end{tabular*}
\end{minipage}
\end{table}
\endgroup

\subsection{Joint Omnimodal Generation}
\label{subsec:omni_decoding_analysis}

Fig.~\ref{fig:omnimodal_generation} compares joint decoding with Modality Anchoring (MA) on the same Omni-StoryBench examples and checkpoint. Joint decoding refines all modalities in one pass; MA generates an anchor first and conditions the others on it in a second pass. We count total function evaluations (NFEs) across passes, so two 128-NFE MA passes require 256 NFEs.

At every matched budget, joint decoding scores higher than all MA variants on the five-metric Overall (Fig.~\ref{fig:omnimodal_generation}), with its largest margin over the strongest anchor at 16--32 NFEs. Joint-128 nearly matches the best two-stage result at 256 NFEs; joint decoding also leads the best anchor in Omni consistency at low budgets. Full results are in App.~\ref{app:additional_results}; measured decoding costs and paired uncertainty are in App.~\ref{app:control-costs}. {The separate paired analysis supports a low-budget gain.}

\begingroup

\paragraph{Cross-modal communication.}
On 64 inputs, we block cross-modal attention in both directions between each generated modality and the other two, retaining the input context, checkpoint, and  seeds with a 128-step refinement  budget. Full joint generation obtains the best mean on all reported text (T), image (I), and speech (S) measures (Tab.~\ref{tab:crossmodal-intervention}), supporting the role of cross-modal communication in mutual refinement. Speech WER measures agreement with each output's extractable intended utterance.  Intervention details are in App.~\ref{app:control-attention}.

\paragraph{Transfer to nonstory instructions.}
Without additional training, we evaluate 32 examples per domain in cooking, DIY installation/repair, and sports/body movement. \method{} achieves the highest text BERTScore among four systems: 0.8378, 0.8622, and 0.8599, respectively.  Its auxiliary Overall scores are 5.696/ 5.569/5.549, above Omni-Diffusion's  4.808/ 4.784/5.002 and MMaDA+Parler's 0.929/3.998/4.618; Qwen+Kontext  scores  higher at 6.796/6.889/6.511.  These results extend the evidence for native text--image--speech generation beyond stories. The auxiliary metric's definition and  full component results are in App.~\ref{app:control-nonstory}.
\endgroup

\section{Conclusion}

We introduced \method{}, an open-source 8B-scale masked-diffusion model supporting text and image understanding and generation, video understanding, and speech understanding and generation within one backbone.
{Its capability-preserving three-stage recipe combines  scheduled  EOS supervision and modality-disentangled merging to  recover inherited  strengths and  stabilize termination.  The resulting model is the first masked-diffusion model to natively generate interleaved text, image, and speech within a single backbone and reverse process.}
{It pairs this native joint-generation capability with strong individual-task results, outperforming prior unified systems on several shared benchmarks and orchestration systems on joint generation. }

\paragraph{{Limitations.}}
The computational cost of full-scale foundation-model training
limits exhaustive ablations of the training recipe. We therefore
complement the full-scale results with controlled studies that retain
the 8B backbone while reducing the data and update budgets. Our joint
generation experiments focus on text, image, and speech; video is
primarily evaluated as an input modality for understanding.
Furthermore, the fidelity of generated images and speech depends on
the underlying discrete tokenizers and detokenizers, including their
ability to preserve fine visual detail and speech prosody. As a
generalist model, \method{} also allocates its capacity across tasks,
whereas specialist models can devote their capacity to a single
modality. Future work could examine the recipe across a broader range
of data, compute, and model scales and extend joint generation to
temporally coherent video. Further discussion is provided in
App.~\ref{app:limitations}.

\FloatBarrier
\addtolength{\bibsep}{-0.4pt}
\bibliography{main}

\begin{thebibliography}{155}
\providecommand{\natexlab}[1]{#1}
\providecommand{\url}[1]{\texttt{#1}}
\expandafter\ifx\csname urlstyle\endcsname\relax
  \providecommand{\doi}[1]{doi: #1}\else
  \providecommand{\doi}{doi: \begingroup \urlstyle{rm}\Url}\fi

\bibitem[Agarwal et~al.(2025)Agarwal, Ahmad, Ai, Altman, Applebaum, Arbus,
  Arora, Bai, Baker, Bao, et~al.]{agarwal2025gptoss}
Sandhini Agarwal, Lama Ahmad, Jason Ai, Sam Altman, Andy Applebaum, Edwin
  Arbus, Rahul~K Arora, Yu~Bai, Bowen Baker, Haiming Bao, et~al.
\newblock gpt-oss-120b \& gpt-oss-20b model card.
\newblock \emph{arXiv preprint arXiv:2508.10925}, 2025.

\bibitem[Anastassiou et~al.(2024)Anastassiou, Chen, Chen, Chen, Chen, Chen,
  Cong, Deng, Ding, Gao, et~al.]{anastassiou2024seedtts}
Philip Anastassiou, Jiawei Chen, Jitong Chen, Yuanzhe Chen, Zhuo Chen, Ziyi
  Chen, Jian Cong, Lelai Deng, Chuang Ding, Lu~Gao, et~al.
\newblock Seed-tts: A family of high-quality versatile speech generation
  models.
\newblock \emph{arXiv preprint arXiv:2406.02430}, 2024.

\bibitem[Ardila et~al.(2020)Ardila, Branson, Davis, Kohler, Meyer, Henretty,
  Morais, Saunders, Tyers, and Weber]{ardila2020common}
Rosana Ardila, Megan Branson, Kelly Davis, Michael Kohler, Josh Meyer, Michael
  Henretty, Reuben Morais, Lindsay Saunders, Francis Tyers, and Gregor Weber.
\newblock Common voice: A massively-multilingual speech corpus.
\newblock In \emph{Proceedings of the twelfth language resources and evaluation
  conference}, pp.\  4218--4222, 2020.

\bibitem[Arnab et~al.(2021)Arnab, Dehghani, Heigold, Sun, Lu{\v{c}}i{\'c}, and
  Schmid]{arnab2021vivit}
Anurag Arnab, Mostafa Dehghani, Georg Heigold, Chen Sun, Mario Lu{\v{c}}i{\'c},
  and Cordelia Schmid.
\newblock Vivit: A video vision transformer.
\newblock In \emph{Proceedings of the IEEE/CVF international conference on
  computer vision}, pp.\  6836--6846, 2021.

\bibitem[Arriola et~al.(2025)Arriola, Gokaslan, Chiu, Yang, Qi, Han, Sahoo, and
  Kuleshov]{arriola2025block}
Marianne Arriola, Aaron Gokaslan, Justin~T Chiu, Zhihan Yang, Zhixuan Qi, Jiaqi
  Han, Subham~Sekhar Sahoo, and Volodymyr Kuleshov.
\newblock Block diffusion: Interpolating between autoregressive and diffusion
  language models.
\newblock \emph{arXiv preprint arXiv:2503.09573}, 2025.

\bibitem[Austin et~al.(2021)Austin, Johnson, Ho, Tarlow, and
  Berg]{austin_structured_2021}
Jacob Austin, Daniel~D. Johnson, Jonathan Ho, Daniel Tarlow, and Rianne van~den
  Berg.
\newblock Structured {Denoising} {Diffusion} {Models} in {Discrete}
  {State}-{Spaces}.
\newblock In \emph{{Proc. of Neural Information Processing Systems (NeurIPS)}},
  volume~34, 2021.

\bibitem[Bai et~al.(2025)Bai, Chen, Liu, Wang, Ge, Song, Dang, Wang, Wang,
  Tang, Zhong, Zhu, Yang, Li, Wan, Wang, Ding, Fu, Xu, Ye, Zhang, Xie, Cheng,
  Zhang, Yang, Xu, and Lin]{Qwen2.5-VL}
Shuai Bai, Keqin Chen, Xuejing Liu, Jialin Wang, Wenbin Ge, Sibo Song, Kai
  Dang, Peng Wang, Shijie Wang, Jun Tang, Humen Zhong, Yuanzhi Zhu, Mingkun
  Yang, Zhaohai Li, Jianqiang Wan, Pengfei Wang, Wei Ding, Zheren Fu, Yiheng
  Xu, Jiabo Ye, Xi~Zhang, Tianbao Xie, Zesen Cheng, Hang Zhang, Zhibo Yang,
  Haiyang Xu, and Junyang Lin.
\newblock Qwen2.5-vl technical report.
\newblock \emph{arXiv preprint arXiv:2502.13923}, 2025.

\bibitem[Bain et~al.(2021)Bain, Nagrani, Varol, and Zisserman]{webvid}
Max Bain, Arsha Nagrani, G{\"u}l Varol, and Andrew Zisserman.
\newblock Frozen in time: A joint video and image encoder for end-to-end
  retrieval.
\newblock In \emph{Proceedings of the IEEE/CVF international conference on
  computer vision}, pp.\  1728--1738, 2021.

\bibitem[Bercovich et~al.(2025)Bercovich, Levy, Golan, Dabbah, El-Yaniv, Puny,
  Galil, Moshe, Ronen, Nabwani, Shahaf, Tropp, Karpas, Zilberstein, Zeng,
  Singhal, Bukharin, Zhang, Konuk, Shen, Mahabaleshwarkar, Kartal, Suhara,
  Delalleau, Chen, Wang, Mosallanezhad, Renduchintala, Qian, Rekesh, Jia,
  Majumdar, Noroozi, Ahmad, Narenthiran, Ficek, Samadi, Huang, Jain, Gitman,
  Moshkov, Du, Toshniwal, Armstrong, Kisacanin, Novikov, Gitman, Bakhturina,
  Scowcroft, Kamalu, Su, Kong, Kliegl, Karimi, Lin, Satheesh, Parmar, Gundecha,
  Norick, Jennings, Prabhumoye, Akter, Patwary, Khattar, Narayanan, Waleffe,
  Zhang, Su, Huang, Kong, Chadha, Jain, Harvey, Segal, Huang, Kashirsky,
  McQueen, Putterman, Lam, Venkatesan, Wu, Nguyen, Kilaru, Wang, Warno,
  Somasamudramath, Bhaskar, Dong, Assaf, Mor, Argov, Junkin, Romanenko, Larroy,
  Katariya, Rovinelli, Balas, Edelman, Bhiwandiwalla, Subramaniam, Ithape,
  Ramamoorthy, Wu, Velury, Almog, Daw, Fridman, Galinkin, Evans, Luna,
  Derczynski, Pope, Long, Schneider, Siman, Grzegorzek, Ribalta, Katariya,
  Conway, Saar, Guan, Pawelec, Prayaga, Kuchaiev, Ginsburg, Olabiyi, Briski,
  Cohen, Catanzaro, Alben, Geifman, Chung, and
  Alexiuk]{bercovich2025llamanemotrone}
Akhiad Bercovich, Itay Levy, Izik Golan, Mohammad Dabbah, Ran El-Yaniv, Omri
  Puny, Ido Galil, Zach Moshe, Tomer Ronen, Najeeb Nabwani, Ido Shahaf, Oren
  Tropp, Ehud Karpas, Ran Zilberstein, Jiaqi Zeng, Soumye Singhal, Alexander
  Bukharin, Yian Zhang, Tugrul Konuk, Gerald Shen, Ameya~Sunil
  Mahabaleshwarkar, Bilal Kartal, Yoshi Suhara, Olivier Delalleau, Zijia Chen,
  Zhilin Wang, David Mosallanezhad, Adi Renduchintala, Haifeng Qian, Dima
  Rekesh, Fei Jia, Somshubra Majumdar, Vahid Noroozi, Wasi~Uddin Ahmad, Sean
  Narenthiran, Aleksander Ficek, Mehrzad Samadi, Jocelyn Huang, Siddhartha
  Jain, Igor Gitman, Ivan Moshkov, Wei Du, Shubham Toshniwal, George Armstrong,
  Branislav Kisacanin, Matvei Novikov, Daria Gitman, Evelina Bakhturina,
  Jane~Polak Scowcroft, John Kamalu, Dan Su, Kezhi Kong, Markus Kliegl, Rabeeh
  Karimi, Ying Lin, Sanjeev Satheesh, Jupinder Parmar, Pritam Gundecha, Brandon
  Norick, Joseph Jennings, Shrimai Prabhumoye, Syeda~Nahida Akter, Mostofa
  Patwary, Abhinav Khattar, Deepak Narayanan, Roger Waleffe, Jimmy Zhang,
  Bor-Yiing Su, Guyue Huang, Terry Kong, Parth Chadha, Sahil Jain, Christine
  Harvey, Elad Segal, Jining Huang, Sergey Kashirsky, Robert McQueen, Izzy
  Putterman, George Lam, Arun Venkatesan, Sherry Wu, Vinh Nguyen, Manoj Kilaru,
  Andrew Wang, Anna Warno, Abhilash Somasamudramath, Sandip Bhaskar, Maka Dong,
  Nave Assaf, Shahar Mor, Omer~Ullman Argov, Scot Junkin, Oleksandr Romanenko,
  Pedro Larroy, Monika Katariya, Marco Rovinelli, Viji Balas, Nicholas Edelman,
  Anahita Bhiwandiwalla, Muthu Subramaniam, Smita Ithape, Karthik Ramamoorthy,
  Yuting Wu, Suguna~Varshini Velury, Omri Almog, Joyjit Daw, Denys Fridman,
  Erick Galinkin, Michael Evans, Katherine Luna, Leon Derczynski, Nikki Pope,
  Eileen Long, Seth Schneider, Guillermo Siman, Tomasz Grzegorzek, Pablo
  Ribalta, Monika Katariya, Joey Conway, Trisha Saar, Ann Guan, Krzysztof
  Pawelec, Shyamala Prayaga, Oleksii Kuchaiev, Boris Ginsburg, Oluwatobi
  Olabiyi, Kari Briski, Jonathan Cohen, Bryan Catanzaro, Jonah Alben, Yonatan
  Geifman, Eric Chung, and Chris Alexiuk.
\newblock Llama-nemotron: Efficient reasoning models, 2025.
\newblock URL \url{https://arxiv.org/abs/2505.00949}.

\bibitem[Bertasius et~al.(2021)Bertasius, Wang, and
  Torresani]{bertasius2021space}
Gedas Bertasius, Heng Wang, and Lorenzo Torresani.
\newblock Is space-time attention all you need for video understanding?
\newblock In \emph{{Proc. of Int'l Conf. on Machine Learning (ICML)}},
  volume~2, pp.\ ~4, 2021.

\bibitem[Bethune et~al.(2026)Bethune, Turrisi, Mlodozeniec, Lopez, Boominathan,
  Bhendawade, Shidani, Pelemans, Olausson, Hjelm, Dixon, Monteiro, Ablin,
  Banna, Blaas, Henderson, Noriy, Busbridge, Susskind, Cuturi, Belousova,
  Zappella, Webb, and Ramapuram]{bethune2026trimodal}
Louis Bethune, Victor Turrisi, Bruno~Kacper Mlodozeniec, Pau~Rodriguez Lopez,
  Lokesh Boominathan, Nikhil Bhendawade, Amitis Shidani, Joris Pelemans,
  Theo~X. Olausson, Devon Hjelm, Paul Dixon, Joao Monteiro, Pierre Ablin,
  Vishnu Banna, Arno Blaas, Nick Henderson, Kari Noriy, Dan Busbridge, Josh
  Susskind, Marco Cuturi, Irina Belousova, Luca Zappella, Russ Webb, and Jason
  Ramapuram.
\newblock The design space of tri-modal masked diffusion models, 2026.
\newblock URL \url{https://arxiv.org/abs/2602.21472}.

\bibitem[Bie et~al.(2025)Bie, Cao, Chen, Du, Gong, Gong, Gu, Hu, Huang, Lan,
  et~al.]{bie2025llada2}
Tiwei Bie, Maosong Cao, Kun Chen, Lun Du, Mingliang Gong, Zhuochen Gong, Yanmei
  Gu, Jiaqi Hu, Zenan Huang, Zhenzhong Lan, et~al.
\newblock Llada2.0: Scaling up diffusion language models to 100b.
\newblock \emph{arXiv preprint arXiv:2512.15745}, 2025.

\bibitem[Cai et~al.(2025)Cai, Chen, Chen, Li, Long, Pan, Qiu, Zhang, Gao, Xu,
  et~al.]{hidreami1technicalreport}
Qi~Cai, Jingwen Chen, Yang Chen, Yehao Li, Fuchen Long, Yingwei Pan, Zhaofan
  Qiu, Yiheng Zhang, Fengbin Gao, Peihan Xu, et~al.
\newblock Hidream-i1: A high-efficient image generative foundation model with
  sparse diffusion transformer.
\newblock \emph{arXiv preprint arXiv:2505.22705}, 2025.

\bibitem[Chang et~al.(2022)Chang, Zhang, Jiang, Liu, and
  Freeman]{Chang_2022_CVPR}
Huiwen Chang, Han Zhang, Lu~Jiang, Ce~Liu, and William~T. Freeman.
\newblock Maskgit: Masked generative image transformer.
\newblock In \emph{Proceedings of the IEEE/CVF Conference on Computer Vision
  and Pattern Recognition (CVPR)}, pp.\  11315--11325, June 2022.

\bibitem[Chen et~al.(2024{\natexlab{a}})Chen, Du, Fang,
  et~al.]{chen2024naivemc}
Chi Chen, Yiyang Du, Zheng Fang, et~al.
\newblock Model composition for multimodal large language models.
\newblock In \emph{Proceedings of the 62nd Annual Meeting of the Association
  for Computational Linguistics}, 2024{\natexlab{a}}.
\newblock URL \url{https://aclanthology.org/2024.acl-long.606/}.

\bibitem[Chen et~al.(2021)Chen, Chai, Wang, Du, Zhang, Weng, Su, Povey, Trmal,
  Zhang, et~al.]{chen2021gigaspeech}
Guoguo Chen, Shuzhou Chai, Guanbo Wang, Jiayu Du, Wei-Qiang Zhang, Chao Weng,
  Dan Su, Daniel Povey, Jan Trmal, Junbo Zhang, et~al.
\newblock Gigaspeech: An evolving, multi-domain asr corpus with 10,000 hours of
  transcribed audio.
\newblock \emph{arXiv preprint arXiv:2106.06909}, 2021.

\bibitem[Chen et~al.(2024{\natexlab{b}})Chen, Gou, Huang, Liu, Tan, Xu, Wang,
  Zhu, Zeng, Yang, et~al.]{chen2024emova}
Kai Chen, Yunhao Gou, Runhui Huang, Zhili Liu, Daxin Tan, Jing Xu, Chunwei
  Wang, Yi~Zhu, Yihan Zeng, Kuo Yang, et~al.
\newblock Emova: Empowering language models to see, hear and speak with vivid
  emotions.
\newblock \emph{arXiv preprint arXiv:2409.18042}, 2024{\natexlab{b}}.

\bibitem[Chen et~al.(2024{\natexlab{c}})Chen, Wei, Li, Dong, Zhang, Zang, Chen,
  Duan, Tang, Yuan, et~al.]{chen2024sharegpt4video}
Lin Chen, Xilin Wei, Jinsong Li, Xiaoyi Dong, Pan Zhang, Yuhang Zang, Zehui
  Chen, Haodong Duan, Zhenyu Tang, Li~Yuan, et~al.
\newblock Sharegpt4video: Improving video understanding and generation with
  better captions.
\newblock \emph{Advances in Neural Information Processing Systems},
  37:\penalty0 19472--19495, 2024{\natexlab{c}}.

\bibitem[Chen et~al.(2025{\natexlab{a}})Chen, Chen, Chen, Chen, Chen, Deng, Du,
  Gao, Gao, Gao, et~al.]{chen2025minmo}
Qian Chen, Yafeng Chen, Yanni Chen, Mengzhe Chen, Yingda Chen, Chong Deng,
  Zhihao Du, Ruize Gao, Changfeng Gao, Zhifu Gao, et~al.
\newblock Minmo: A multimodal large language model for seamless voice
  interaction.
\newblock \emph{arXiv preprint arXiv:2501.06282}, 2025{\natexlab{a}}.

\bibitem[Chen et~al.(2025{\natexlab{b}})Chen, Wu, Liu, Pan, Liu, Xie, Yu, and
  Ruan]{chen2025januspro}
Xiaokang Chen, Zhiyu Wu, Xingchao Liu, Zizheng Pan, Wen Liu, Zhenda Xie,
  Xingkai Yu, and Chong Ruan.
\newblock Janus-pro: Unified multimodal understanding and generation with data
  and model scaling.
\newblock \emph{arXiv preprint arXiv:2501.17811}, 2025{\natexlab{b}}.

\bibitem[Chen et~al.(2024{\natexlab{d}})Chen, Wu, Wang, Su, Chen, Xing, Zhong,
  Zhang, Zhu, Lu, et~al.]{chen2024internvl}
Zhe Chen, Jiannan Wu, Wenhai Wang, Weijie Su, Guo Chen, Sen Xing, Muyan Zhong,
  Qinglong Zhang, Xizhou Zhu, Lewei Lu, et~al.
\newblock Internvl: Scaling up vision foundation models and aligning for
  generic visual-linguistic tasks.
\newblock In \emph{Proceedings of the IEEE/CVF Conference on Computer Vision
  and Pattern Recognition}, pp.\  24185--24198, 2024{\natexlab{d}}.

\bibitem[Cheng et~al.(2025)Cheng, Jiang, Zhou, Liu, Tao, Zhang, Qi, and
  Zhou]{cheng2025sdarvl}
Shuang Cheng, Yuhua Jiang, Zineng Zhou, Dawei Liu, Wang Tao, Linfeng Zhang,
  Biqing Qi, and Bowen Zhou.
\newblock {SDAR-VL}: Stable and efficient block-wise diffusion for
  vision-language understanding, 2025.
\newblock URL \url{https://arxiv.org/abs/2512.14068}.

\bibitem[Chu et~al.(2024)Chu, Xu, Yang, Wei, Wei, Guo, Leng, Lv, He, Lin, Zhou,
  and Zhou]{Qwen2-Audio}
Yunfei Chu, Jin Xu, Qian Yang, Haojie Wei, Xipin Wei, Zhifang Guo, Yichong
  Leng, Yuanjun Lv, Jinzheng He, Junyang Lin, Chang Zhou, and Jingren Zhou.
\newblock Qwen2-audio technical report.
\newblock \emph{arXiv preprint arXiv:2407.10759}, 2024.

\bibitem[Clark et~al.(2018)Clark, Cowhey, Etzioni, Khot, Sabharwal, Schoenick,
  and Tafjord]{arc}
Peter Clark, Isaac Cowhey, Oren Etzioni, Tushar Khot, Ashish Sabharwal, Carissa
  Schoenick, and Oyvind Tafjord.
\newblock Think you have solved question answering? try arc, the ai2 reasoning
  challenge.
\newblock \emph{arXiv:1803.05457v1}, 2018.

\bibitem[Cobbe et~al.(2021)Cobbe, Kosaraju, Bavarian, Chen, Jun, Kaiser,
  Plappert, Tworek, Hilton, Nakano, Hesse, and Schulman]{cobbe2021gsm8k}
Karl Cobbe, Vineet Kosaraju, Mohammad Bavarian, Mark Chen, Heewoo Jun, Lukasz
  Kaiser, Matthias Plappert, Jerry Tworek, Jacob Hilton, Reiichiro Nakano,
  Christopher Hesse, and John Schulman.
\newblock Training verifiers to solve math word problems.
\newblock \emph{arXiv preprint arXiv:2110.14168}, 2021.

\bibitem[Cui et~al.(2025)Cui, Chen, Deng, Huang, Li, Liu, Liu, Luo, Wang, Wang,
  et~al.]{cui2025emu3_5}
Yufeng Cui, Honghao Chen, Haoge Deng, Xu~Huang, Xinghang Li, Jirong Liu, Yang
  Liu, Zhuoyan Luo, Jinsheng Wang, Wenxuan Wang, et~al.
\newblock Emu3.5: Native multimodal models are world learners.
\newblock \emph{arXiv preprint arXiv:2510.26583}, 2025.

\bibitem[Das et~al.(2024)Das, Dingliwal, Ronanki, Paturi, Huang, Mathur, Yuan,
  Bekal, Niu, Jayanthi, et~al.]{das2024speechverse}
Nilaksh Das, Saket Dingliwal, Srikanth Ronanki, Rohit Paturi, Zhaocheng Huang,
  Prashant Mathur, Jie Yuan, Dhanush Bekal, Xing Niu, Sai~Muralidhar Jayanthi,
  et~al.
\newblock Speechverse: A large-scale generalizable audio language model.
\newblock \emph{arXiv preprint arXiv:2405.08295}, 2024.

\bibitem[DeepSeek-AI(2025)]{qwendeepseek}
DeepSeek-AI.
\newblock Deepseek-r1: Incentivizing reasoning capability in llms via
  reinforcement learning, 2025.
\newblock URL \url{https://arxiv.org/abs/2501.12948}.

\bibitem[Deng et~al.(2025)Deng, Zhu, Li, Gou, Li, Wang, Zhong, Yu, Nie, Song,
  et~al.]{bagle}
Chaorui Deng, Deyao Zhu, Kunchang Li, Chenhui Gou, Feng Li, Zeyu Wang, Shu
  Zhong, Weihao Yu, Xiaonan Nie, Ziang Song, et~al.
\newblock Emerging properties in unified multimodal pretraining.
\newblock \emph{arXiv preprint arXiv:2505.14683}, 2025.

\bibitem[Dong et~al.(2023)Dong, Han, Peng, Qi, Ge, Yang, Zhao, Sun, Zhou, Wei,
  et~al.]{dong2023dreamllm}
Runpei Dong, Chunrui Han, Yuang Peng, Zekun Qi, Zheng Ge, Jinrong Yang, Liang
  Zhao, Jianjian Sun, Hongyu Zhou, Haoran Wei, et~al.
\newblock Dreamllm: Synergistic multimodal comprehension and creation.
\newblock \emph{arXiv preprint arXiv:2309.11499}, 2023.

\bibitem[Face(2025)]{openr1}
Hugging Face.
\newblock Open r1: A fully open reproduction of deepseek-r1, January 2025.
\newblock URL \url{https://github.com/huggingface/open-r1}.

\bibitem[Fang et~al.(2025)Fang, Yu, Duan, Huang, Bai, Cai, Wang, Liu, Liu, and
  Li]{fang2025fluxreason}
Rongyao Fang, Aldrich Yu, Chengqi Duan, Linjiang Huang, Shuai Bai, Yuxuan Cai,
  Kun Wang, Si~Liu, Xihui Liu, and Hongsheng Li.
\newblock Flux-reason-6m \& prism-bench: A million-scale text-to-image
  reasoning dataset and comprehensive benchmark.
\newblock \emph{arXiv preprint arXiv:2509.09680}, 2025.

\bibitem[Fu et~al.(2023)Fu, Chen, Shen, Qin, Zhang, Lin, Yang, Zheng, Li, Sun,
  et~al.]{fu2023mme}
Chaoyou Fu, Peixian Chen, Yunhang Shen, Yulei Qin, Mengdan Zhang, Xu~Lin,
  Jinrui Yang, Xiawu Zheng, Ke~Li, Xing Sun, et~al.
\newblock Mme: A comprehensive evaluation benchmark for multimodal large
  language models.
\newblock \emph{arXiv preprint arXiv:2306.13394}, 2023.

\bibitem[Fu et~al.(2024)Fu, Dai, Luo, Li, Ren, Zhang, Wang, Zhou, Shen, Zhang,
  Chen, Li, Lin, Zhao, Li, Xu, Zheng, Chen, Shan, He, and Sun]{fu2024videomme}
Chaoyou Fu, Yuhan Dai, Yongdong Luo, Lei Li, Shuhuai Ren, Renrui Zhang, Zihan
  Wang, Chenyu Zhou, Yunhang Shen, Mengdan Zhang, Peixian Chen, Yanwei Li,
  Shaohui Lin, Sirui Zhao, Ke~Li, Tong Xu, Xiawu Zheng, Enhong Chen, Caifeng
  Shan, Ran He, and Xing Sun.
\newblock Video-mme: The first-ever comprehensive evaluation benchmark of
  multi-modal llms in video analysis.
\newblock \emph{arXiv preprint arXiv:2405.21075}, 2024.

\bibitem[Fu et~al.(2025)Fu, Lin, Wang, Zhang, Shen, Liu, Cao, Long, Gao, Li,
  et~al.]{fu2025vita}
Chaoyou Fu, Haojia Lin, Xiong Wang, Yi-Fan Zhang, Yunhang Shen, Xiaoyu Liu,
  Haoyu Cao, Zuwei Long, Heting Gao, Ke~Li, et~al.
\newblock Vita-1.5: Towards gpt-4o level real-time vision and speech
  interaction.
\newblock \emph{arXiv preprint arXiv:2501.01957}, 2025.

\bibitem[Gadre et~al.(2023)Gadre, Ilharco, Fang, Hayase, Smyrnis, Nguyen,
  Marten, Wortsman, Ghosh, Zhang, et~al.]{gadre2023datacomp}
Samir~Yitzhak Gadre, Gabriel Ilharco, Alex Fang, Jonathan Hayase, Georgios
  Smyrnis, Thao Nguyen, Ryan Marten, Mitchell Wortsman, Dhruba Ghosh, Jieyu
  Zhang, et~al.
\newblock Datacomp: In search of the next generation of multimodal datasets.
\newblock \emph{Advances in Neural Information Processing Systems},
  36:\penalty0 27092--27112, 2023.

\bibitem[Gao et~al.(2025)Gao, Gong, Guo, Hou, Lai, Li, Li, Lian, Liao, Liu,
  et~al.]{gao2025seedream3}
Yu~Gao, Lixue Gong, Qiushan Guo, Xiaoxia Hou, Zhichao Lai, Fanshi Li, Liang Li,
  Xiaochen Lian, Chao Liao, Liyang Liu, et~al.
\newblock Seedream 3.0 technical report.
\newblock \emph{arXiv preprint arXiv:2504.11346}, 2025.

\bibitem[Ge et~al.(2024)Ge, Zhao, Zhu, Ge, Yi, Song, Li, Ding, and
  Shan]{ge2024seed}
Yuying Ge, Sijie Zhao, Jinguo Zhu, Yixiao Ge, Kun Yi, Lin Song, Chen Li,
  Xiaohan Ding, and Ying Shan.
\newblock Seed-x: Multimodal models with unified multi-granularity
  comprehension and generation.
\newblock \emph{arXiv preprint arXiv:2404.14396}, 2024.

\bibitem[Ghosh et~al.(2023)Ghosh, Hajishirzi, and Schmidt]{ghosh2023geneval}
Dhruba Ghosh, Hannaneh Hajishirzi, and Ludwig Schmidt.
\newblock Geneval: An object-focused framework for evaluating text-to-image
  alignment.
\newblock \emph{Advances in Neural Information Processing Systems},
  36:\penalty0 52132--52152, 2023.

\bibitem[Gong et~al.(2023)Gong, Li, Feng, Wu, and Kong]{gong_diffuseq_2022}
Shansan Gong, Mukai Li, Jiangtao Feng, Zhiyong Wu, and Lingpeng Kong.
\newblock {DiffuSeq}: {Sequence} to {Sequence} {Text} {Generation} with
  {Diffusion} {Models}.
\newblock In \emph{{Proc. of Int'l Conf. on Learning Representations (ICLR) }},
  2023.

\bibitem[Grattafiori et~al.(2024)Grattafiori, Dubey, Jauhri, Pandey, Kadian,
  Al-Dahle, Letman, Mathur, Schelten, Vaughan, et~al.]{grattafiori2024llama3}
Aaron Grattafiori, Abhimanyu Dubey, Abhinav Jauhri, Abhinav Pandey, Abhishek
  Kadian, Ahmad Al-Dahle, Aiesha Letman, Akhil Mathur, Alan Schelten, Alex
  Vaughan, et~al.
\newblock The llama 3 herd of models.
\newblock \emph{arXiv preprint arXiv:2407.21783}, 2024.

\bibitem[He et~al.(2023)He, Sun, Tang, Wang, Huang, and
  Qiu]{he_diffusionbert_2023}
Zhengfu He, Tianxiang Sun, Qiong Tang, Kuanning Wang, Xuanjing Huang, and
  Xipeng Qiu.
\newblock {DiffusionBERT}: {Improving} {Generative} {Masked} {Language}
  {Models} with {Diffusion} {Models}.
\newblock In \emph{{Proc. of Annual Meeting of the Association for
  Computational Linguistics (ACL)}}, 2023.

\bibitem[Hendrycks et~al.(2021{\natexlab{a}})Hendrycks, Burns, Basart, Zou,
  Mazeika, Song, and Steinhardt]{mmlu}
Dan Hendrycks, Collin Burns, Steven Basart, Andy Zou, Mantas Mazeika, Dawn
  Song, and Jacob Steinhardt.
\newblock Measuring massive multitask language understanding.
\newblock \emph{Proceedings of the International Conference on Learning
  Representations (ICLR)}, 2021{\natexlab{a}}.

\bibitem[Hendrycks et~al.(2021{\natexlab{b}})Hendrycks, Burns, Kadavath, Arora,
  Basart, Tang, Song, and Steinhardt]{hendrycksmath2021}
Dan Hendrycks, Collin Burns, Saurav Kadavath, Akul Arora, Steven Basart, Eric
  Tang, Dawn Song, and Jacob Steinhardt.
\newblock Measuring mathematical problem solving with the math dataset.
\newblock \emph{arXiv preprint arXiv:2103.03874}, 2021{\natexlab{b}}.

\bibitem[hexgard(2025)]{kokoro}
hexgard.
\newblock Kokoro-82m.
\newblock https://github.com/hexgrad/kokoro, 2025.

\bibitem[Ho \& Salimans(2022)Ho and Salimans]{ho2022classifier}
Jonathan Ho and Tim Salimans.
\newblock Classifier-free diffusion guidance.
\newblock \emph{arXiv preprint arXiv:2207.12598}, 2022.

\bibitem[Hu et~al.(2026)Hu, Zhu, He, Guo, Zhang, Wang, Guo, Jiang, Hao, Guo,
  et~al.]{hu2026qwen3tts}
Hangrui Hu, Xinfa Zhu, Ting He, Dake Guo, Bin Zhang, Xiong Wang, Zhifang Guo,
  Ziyue Jiang, Hongkun Hao, Zishan Guo, et~al.
\newblock Qwen3-tts technical report.
\newblock \emph{arXiv preprint arXiv:2601.15621}, 2026.

\bibitem[Hu et~al.(2024)Hu, Wang, Fang, Fu, Cheng, and Yu]{hu2024ella_dpgbench}
Xiwei Hu, Rui Wang, Yixiao Fang, Bin Fu, Pei Cheng, and Gang Yu.
\newblock Ella: Equip diffusion models with llm for enhanced semantic
  alignment.
\newblock \emph{arXiv preprint arXiv:2403.05135}, 2024.

\bibitem[Huang et~al.(2022)Huang, Zhang, Yeung, Jiang, and
  Liu]{huang2022spiral}
Wenyong Huang, Zhenhe Zhang, Yu~Ting Yeung, Xin Jiang, and Qun Liu.
\newblock {SPIRAL}: Self-supervised perturbation-invariant representation
  learning for speech pre-training.
\newblock In \emph{International Conference on Learning Representations}, 2022.
\newblock URL \url{https://openreview.net/forum?id=TBpg4PnXhYH}.

\bibitem[Hudson \& Manning(2019)Hudson and Manning]{hudson2019gqa}
Drew~A Hudson and Christopher~D Manning.
\newblock Gqa: A new dataset for real-world visual reasoning and compositional
  question answering.
\newblock In \emph{Proceedings of the IEEE/CVF conference on computer vision
  and pattern recognition}, pp.\  6700--6709, 2019.

\bibitem[Huggingface(2025)]{mixtureofthought}
Huggingface.
\newblock Mixture of thought dataset.
\newblock https://huggingface.co/datasets/open-r1/Mixture-of-Thoughts, 2025.

\bibitem[Hui et~al.(2024)Hui, Yang, Zhao, Shi, Wang, Wang, Zhou, and
  Xie]{hui2024hqedit}
Mude Hui, Siwei Yang, Bingchen Zhao, Yichun Shi, Heng Wang, Peng Wang, Yuyin
  Zhou, and Cihang Xie.
\newblock Hq-edit: A high-quality dataset for instruction-based image editing.
\newblock \emph{arXiv preprint arXiv:2404.09990}, 2024.

\bibitem[Hyeon et~al.(2026)Hyeon, Lee, Lim, Kim, Kim, and Do]{omnistorybench}
Sieun Hyeon, Yejoon Lee, Mintaek Lim, Woojin Kim, Jaeik Kim, and Jaeyoung Do.
\newblock {What Comes Next? Omni-StoryBench for Evaluating Story-Grounded
  Omnimodal Generation}, 2026.
\newblock URL \url{https://arxiv.org/abs/2609.37317}.

\bibitem[Jiang et~al.(2024)Jiang, He, Zeng, Wei, Ku, Liu, and
  Chen]{Jiang2024MANTISIM}
Dongfu Jiang, Xuan He, Huaye Zeng, Cong Wei, Max~W.F. Ku, Qian Liu, and Wenhu
  Chen.
\newblock Mantis: Interleaved multi-image instruction tuning.
\newblock \emph{Transactions on Machine Learning Research}, 2024, 2024.
\newblock URL \url{https://openreview.net/forum?id=skLtdUVaJa}.

\bibitem[Khanna et~al.(2025)Khanna, Kharbanda, Li, Varma, Wang, Birnbaum, Luo,
  Miraoui, Palrecha, Ermon, et~al.]{khanna2025mercury}
Samar Khanna, Siddhant Kharbanda, Shufan Li, Harshit Varma, Eric Wang, Sawyer
  Birnbaum, Ziyang Luo, Yanis Miraoui, Akash Palrecha, Stefano Ermon, et~al.
\newblock Mercury: Ultra-fast language models based on diffusion.
\newblock \emph{arXiv e-prints}, pp.\  arXiv--2506, 2025.

\bibitem[Kim et~al.(2021)Kim, Kong, and Son]{kim2021vits}
Jaehyeon Kim, Jungil Kong, and Juhee Son.
\newblock Conditional variational autoencoder with adversarial learning for
  end-to-end text-to-speech.
\newblock In Marina Meila and Tong Zhang (eds.), \emph{Proceedings of the 38th
  International Conference on Machine Learning}, volume 139 of
  \emph{Proceedings of Machine Learning Research}, pp.\  5530--5540. PMLR,
  18--24 Jul 2021.
\newblock URL \url{https://proceedings.mlr.press/v139/kim21f.html}.

\bibitem[Kim et~al.(2025)Kim, Shah, Kontonis, Kakade, and Chen]{kim2025train}
Jaeyeon Kim, Kulin Shah, Vasilis Kontonis, Sham Kakade, and Sitan Chen.
\newblock Train for the worst, plan for the best: Understanding token ordering
  in masked diffusions.
\newblock \emph{{Proc. of Int'l Conf. on Machine Learning (ICML)}}, 2025.

\bibitem[Kim \& Do(2025)Kim and Do]{kim2025dont}
Woojin Kim and Jaeyoung Do.
\newblock Don{\textquoteright}t let it fade: Preserving edits in diffusion
  language models via token timestep allocation.
\newblock In \emph{The Thirty-ninth Annual Conference on Neural Information
  Processing Systems}, 2025.
\newblock URL \url{https://openreview.net/forum?id=LOVuux2kvs}.

\bibitem[Kirkpatrick et~al.(2017)Kirkpatrick, Pascanu, Rabinowitz, Veness,
  Desjardins, Rusu, Milan, Quan, Ramalho, Grabska-Barwinska,
  et~al.]{kirkpatrick2017overcoming}
James Kirkpatrick, Razvan Pascanu, Neil Rabinowitz, Joel Veness, Guillaume
  Desjardins, Andrei~A Rusu, Kieran Milan, John Quan, Tiago Ramalho, Agnieszka
  Grabska-Barwinska, et~al.
\newblock Overcoming catastrophic forgetting in neural networks.
\newblock \emph{Proceedings of the national academy of sciences}, 114\penalty0
  (13):\penalty0 3521--3526, 2017.

\bibitem[Kirstain et~al.(2023)Kirstain, Polyak, Singer, Matiana, Penna, and
  Levy]{kirstain2023pickapick}
Yuval Kirstain, Adam Polyak, Uriel Singer, Shahbuland Matiana, Joe Penna, and
  Omer Levy.
\newblock Pick-a-pic: An open dataset of user preferences for text-to-image
  generation.
\newblock \emph{Advances in neural information processing systems},
  36:\penalty0 36652--36663, 2023.

\bibitem[Labs(2024)]{flux2024}
Black~Forest Labs.
\newblock Flux.
\newblock \url{https://github.com/black-forest-labs/flux}, 2024.

\bibitem[Labs(2025)]{flux-2-2025}
Black~Forest Labs.
\newblock {FLUX.2: Frontier Visual Intelligence}.
\newblock \url{https://bfl.ai/blog/flux-2}, 2025.

\bibitem[Labs et~al.(2025)Labs, Batifol, Blattmann, Boesel, Consul, Diagne,
  Dockhorn, English, English, Esser, Kulal, Lacey, Levi, Li, Lorenz, Müller,
  Podell, Rombach, Saini, Sauer, and Smith]{labs2025flux1kontextflowmatching}
Black~Forest Labs, Stephen Batifol, Andreas Blattmann, Frederic Boesel, Saksham
  Consul, Cyril Diagne, Tim Dockhorn, Jack English, Zion English, Patrick
  Esser, Sumith Kulal, Kyle Lacey, Yam Levi, Cheng Li, Dominik Lorenz, Jonas
  Müller, Dustin Podell, Robin Rombach, Harry Saini, Axel Sauer, and Luke
  Smith.
\newblock Flux.1 kontext: Flow matching for in-context image generation and
  editing in latent space, 2025.
\newblock URL \url{https://arxiv.org/abs/2506.15742}.

\bibitem[Lee et~al.(2023)Lee, Hunter, and Ruiz]{lee2023platypus}
Ariel~N Lee, Cole~J Hunter, and Nataniel Ruiz.
\newblock Platypus: Quick, cheap, and powerful refinement of llms.
\newblock \emph{arXiv preprint arXiv:2308.07317}, 2023.

\bibitem[Li et~al.(2024{\natexlab{a}})Li, Zhang, Guo, Zhang, Li, Zhang, Zhang,
  Zhang, Li, Liu, et~al.]{li2024llavaonevision}
Bo~Li, Yuanhan Zhang, Dong Guo, Renrui Zhang, Feng Li, Hao Zhang, Kaichen
  Zhang, Peiyuan Zhang, Yanwei Li, Ziwei Liu, et~al.
\newblock Llava-onevision: Easy visual task transfer.
\newblock \emph{arXiv preprint arXiv:2408.03326}, 2024{\natexlab{a}}.

\bibitem[Li et~al.(2025{\natexlab{a}})Li, Du, Li, Goh, Wang, Wang, Liu, Tang,
  Alharbi, He, and Zhang]{li2025mmer}
Junlin Li, Guodong Du, Jing Li, Sim~Kuan Goh, Wenya Wang, Yequan Wang, Fangming
  Liu, Ho-Kin Tang, Saleh Alharbi, Daojing He, and Min Zhang.
\newblock Multi-modality expansion and retention for llms through parameter
  merging and decoupling, 2025{\natexlab{a}}.
\newblock URL \url{https://arxiv.org/abs/2505.17110}.

\bibitem[Li et~al.(2024{\natexlab{b}})Li, Wang, He, Li, Wang, Liu, Wang, Xu,
  Chen, Luo, et~al.]{li2024mvbench}
Kunchang Li, Yali Wang, Yinan He, Yizhuo Li, Yi~Wang, Yi~Liu, Zun Wang, Jilan
  Xu, Guo Chen, Ping Luo, et~al.
\newblock Mvbench: A comprehensive multi-modal video understanding benchmark.
\newblock In \emph{Proceedings of the IEEE/CVF Conference on Computer Vision
  and Pattern Recognition}, pp.\  22195--22206, 2024{\natexlab{b}}.

\bibitem[Li et~al.(2026)Li, Long, Shen, Gao, Cao, Sun, Shan, He, and
  Fu]{li2026omnidiffusion}
Lijiang Li, Zuwei Long, Yunhang Shen, Heting Gao, Haoyu Cao, Xing Sun, Caifeng
  Shan, Ran He, and Chaoyou Fu.
\newblock Omni-diffusion: Unified multimodal understanding and generation with
  masked discrete diffusion, 2026.
\newblock URL \url{https://arxiv.org/abs/2603.06577}.
\newblock Accepted at ICML 2026.

\bibitem[Li et~al.(2025{\natexlab{b}})Li, Gu, Liu, Lin, Wei, Grover, and
  Kuen]{li2025lavidao}
Shufan Li, Jiuxiang Gu, Kangning Liu, Zhe Lin, Zijun Wei, Aditya Grover, and
  Jason Kuen.
\newblock Lavida-o: Elastic large masked diffusion models for unified
  multimodal understanding and generation.
\newblock \emph{arXiv preprint arXiv:2509.19244}, 2025{\natexlab{b}}.

\bibitem[Li et~al.(2025{\natexlab{c}})Li, Kallidromitis, Bansal, Gokul, Kato,
  Kozuka, Kuen, Lin, Chang, and Grover]{li2025lavida}
Shufan Li, Konstantinos Kallidromitis, Hritik Bansal, Akash Gokul, Yusuke Kato,
  Kazuki Kozuka, Jason Kuen, Zhe Lin, Kai-Wei Chang, and Aditya Grover.
\newblock Lavida: A large diffusion language model for multimodal
  understanding, 2025{\natexlab{c}}.
\newblock URL \url{https://arxiv.org/abs/2505.16839}.

\bibitem[Li et~al.(2022)Li, Thickstun, Gulrajani, Liang, and
  Hashimoto]{li_diffusion-lm_2022}
Xiang~Lisa Li, John Thickstun, Ishaan Gulrajani, Percy Liang, and Tatsunori
  Hashimoto.
\newblock Diffusion-{LM} {Improves} {Controllable} {Text} {Generation}.
\newblock In \emph{{Proc. of Neural Information Processing Systems (NeurIPS)}},
  volume~35, 2022.

\bibitem[Li et~al.(2025{\natexlab{d}})Li, Liu, Zhang, Chen, Li, Li, Liu, Ming,
  Dong, Pan, et~al.]{li2025baichuan}
Yadong Li, Jun Liu, Tao Zhang, Song Chen, Tianpeng Li, Zehuan Li, Lijun Liu,
  Lingfeng Ming, Guosheng Dong, Da~Pan, et~al.
\newblock Baichuan-omni-1.5 technical report.
\newblock \emph{arXiv preprint arXiv:2501.15368}, 2025{\natexlab{d}}.

\bibitem[Li et~al.(2023)Li, Du, Zhou, Wang, Zhao, and Wen]{pope}
Yifan Li, Yifan Du, Kun Zhou, Jinpeng Wang, Wayne~Xin Zhao, and Ji-Rong Wen.
\newblock Evaluating object hallucination in large vision-language models.
\newblock In \emph{Proceedings of the 2023 conference on empirical methods in
  natural language processing}, pp.\  292--305, 2023.

\bibitem[Lin et~al.(2025)Lin, Li, Cheng, Niu, Ye, He, Yuan, Yu, Wang, Ge,
  et~al.]{lin2025uniworld}
Bin Lin, Zongjian Li, Xinhua Cheng, Yuwei Niu, Yang Ye, Xianyi He, Shenghai
  Yuan, Wangbo Yu, Shaodong Wang, Yunyang Ge, et~al.
\newblock Uniworld-v1: High-resolution semantic encoders for unified visual
  understanding and generation.
\newblock \emph{arXiv preprint arXiv:2506.03147}, 2025.

\bibitem[Lin et~al.(2023)Lin, Gong, Shen, Wu, Fan, Lin, Duan, and
  Chen]{lin_text_2023}
Zhenghao Lin, Yeyun Gong, Yelong Shen, Tong Wu, Zhihao Fan, Chen Lin, Nan Duan,
  and Weizhu Chen.
\newblock Text generation with diffusion language models: a pre-training
  approach with continuous paragraph denoise.
\newblock In \emph{{Proc. of Int'l Conf. on Machine Learning (ICML)}}, 2023.

\bibitem[Liu et~al.(2026)Liu, Khandelwal, Subramanian, Jouault, Rastogi,
  Sad{\'e}, Jeffares, Jiang, Cahill, Gavaudan, et~al.]{liu2026ministral3}
Alexander~H Liu, Kartik Khandelwal, Sandeep Subramanian, Victor Jouault,
  Abhinav Rastogi, Adrien Sad{\'e}, Alan Jeffares, Albert Jiang, Alexandre
  Cahill, Alexandre Gavaudan, et~al.
\newblock Ministral 3.
\newblock \emph{arXiv preprint arXiv:2601.08584}, 2026.

\bibitem[Liu et~al.(2025{\natexlab{a}})Liu, Han, Xing, Yin, Wang, Cheng, Liao,
  Wang, Fu, Han, et~al.]{liu2025step1x}
Shiyu Liu, Yucheng Han, Peng Xing, Fukun Yin, Rui Wang, Wei Cheng, Jiaqi Liao,
  Yingming Wang, Honghao Fu, Chunrui Han, et~al.
\newblock Step1x-edit: A practical framework for general image editing.
\newblock \emph{arXiv preprint arXiv:2504.17761}, 2025{\natexlab{a}}.

\bibitem[Liu et~al.(2023)Liu, Duan, Zhang, Li, Zhang, Zhao, Yuan, Wang, He,
  Liu, Chen, and Lin]{MMBench}
Yuan Liu, Haodong Duan, Yuanhan Zhang, Bo~Li, Songyang Zhang, Wangbo Zhao, Yike
  Yuan, Jiaqi Wang, Conghui He, Ziwei Liu, Kai Chen, and Dahua Lin.
\newblock Mmbench: Is your multi-modal model an all-around player?
\newblock \emph{arXiv preprint arXiv:2307.06281}, 2023.

\bibitem[Liu et~al.(2024)Liu, Li, Liu, Wang, Ren, Li, Chen, Sun, and
  Hou]{liu2024tempcompass}
Yuanxin Liu, Shicheng Li, Yi~Liu, Yuxiang Wang, Shuhuai Ren, Lei Li, Sishuo
  Chen, Xu~Sun, and Lu~Hou.
\newblock Tempcompass: Do video llms really understand videos?
\newblock In \emph{Findings of the Association for Computational Linguistics:
  ACL 2024}, pp.\  8731--8772, 2024.

\bibitem[Liu et~al.(2025{\natexlab{b}})Liu, Yang, Zhang, Chen, Zou, Wei, Wang,
  and Zhang]{liu2025dllmcacheacceleratingdiffusionlarge}
Zhiyuan Liu, Yicun Yang, Yaojie Zhang, Junjie Chen, Chang Zou, Qingyuan Wei,
  Shaobo Wang, and Linfeng Zhang.
\newblock dllm-cache: Accelerating diffusion large language models with
  adaptive caching, 2025{\natexlab{b}}.
\newblock URL \url{https://arxiv.org/abs/2506.06295}.

\bibitem[Lozhkov et~al.(2025)Lozhkov, Kydlíček, Allal, Penedo, Beeching,
  Gallouédec, Habib, Tunstall, and von Werra]{lozhkov2025openr1math220k}
Anton Lozhkov, Hynek Kydlíček, Loubna~Ben Allal, Guilherme Penedo, Edward
  Beeching, Quentin Gallouédec, Nathan Habib, Lewis Tunstall, and Leandro von
  Werra.
\newblock Openr1-math-220k.
\newblock \url{https://huggingface.co/datasets/open-r1/OpenR1-Math-220k}, 2025.

\bibitem[Lu et~al.(2024{\natexlab{a}})Lu, Clark, Lee, Zhang, Khosla, Marten,
  Hoiem, and Kembhavi]{lu2024unifiedio2}
Jiasen Lu, Christopher Clark, Sangho Lee, Zichen Zhang, Savya Khosla, Ryan
  Marten, Derek Hoiem, and Aniruddha Kembhavi.
\newblock Unified-io 2: Scaling autoregressive multimodal models with vision
  language audio and action.
\newblock In \emph{Proceedings of the IEEE/CVF Conference on Computer Vision
  and Pattern Recognition}, pp.\  26439--26455, 2024{\natexlab{a}}.

\bibitem[Lu et~al.(2024{\natexlab{b}})Lu, Li, Chen, Xu, Luo, Zhang, and
  Ye]{lu2024ovis}
Shiyin Lu, Yang Li, Qing-Guo Chen, Zhao Xu, Weihua Luo, Kaifu Zhang, and
  Han-Jia Ye.
\newblock Ovis: Structural embedding alignment for multimodal large language
  model.
\newblock \emph{arXiv:2405.20797}, 2024{\natexlab{b}}.

\bibitem[Luo et~al.(2025{\natexlab{a}})Luo, Lin, Zhang, Wu, Liu, Yang, Li,
  Chen, Li, Zhang, et~al.]{luo2025openomni}
Run Luo, Ting-En Lin, Haonan Zhang, Yuchuan Wu, Xiong Liu, Min Yang, Yongbin
  Li, Longze Chen, Jiaming Li, Lei Zhang, et~al.
\newblock Openomni: Advancing open-source omnimodal large language models with
  progressive multimodal alignment and real-time self-aware emotional speech
  synthesis.
\newblock \emph{arXiv preprint arXiv:2501.04561}, 2025{\natexlab{a}}.

\bibitem[Luo et~al.(2025{\natexlab{b}})Luo, Xia, Wang, Chen, Shan, Luo, Yang,
  and Chua]{next-omni}
Run Luo, Xiaobo Xia, Lu~Wang, Longze Chen, Renke Shan, Jing Luo, Min Yang, and
  Tat-Seng Chua.
\newblock Next-omni: Towards any-to-any omnimodal foundation models with
  discrete flow matching, 2025{\natexlab{b}}.
\newblock URL \url{https://arxiv.org/abs/2510.13721}.

\bibitem[Lyth \& King(2024)Lyth and King]{parler}
Dan Lyth and Simon King.
\newblock Natural language guidance of high-fidelity text-to-speech with
  synthetic annotations.
\newblock \emph{arXiv preprint arXiv:2402.01912}, 2024.

\bibitem[Ma et~al.(2025{\natexlab{a}})Ma, Yu, Fang, and Wang]{ma2025dkvcache}
Xinyin Ma, Runpeng Yu, Gongfan Fang, and Xinchao Wang.
\newblock d{KV}-cache: The cache for diffusion language models.
\newblock In \emph{The Thirty-ninth Annual Conference on Neural Information
  Processing Systems}, 2025{\natexlab{a}}.
\newblock URL \url{https://openreview.net/forum?id=Gppo2JImHs}.

\bibitem[Ma et~al.(2025{\natexlab{b}})Ma, Du, Wei, Chen, Xu, Wang, Feng, Lu,
  Liu, Qi, Zhang, Tao, Feng, Jiang, Xu, Huang, Zhuang, Xu, Hu, Lan, Zhao, Li,
  and Zheng]{ma2025dinferefficientinferenceframework}
Yuxin Ma, Lun Du, Lanning Wei, Kun Chen, Qian Xu, Kangyu Wang, Guofeng Feng,
  Guoshan Lu, Lin Liu, Xiaojing Qi, Xinyuan Zhang, Zhen Tao, Haibo Feng, Ziyun
  Jiang, Ying Xu, Zenan Huang, Yihong Zhuang, Haokai Xu, Jiaqi Hu, Zhenzhong
  Lan, Junbo Zhao, Jianguo Li, and Da~Zheng.
\newblock dinfer: An efficient inference framework for diffusion language
  models, 2025{\natexlab{b}}.
\newblock URL \url{https://arxiv.org/abs/2510.08666}.

\bibitem[Moshkov et~al.(2025)Moshkov, Hanley, Sorokin, Toshniwal, Henkel,
  Schifferer, Du, and Gitman]{openmathreasoning}
Ivan Moshkov, Darragh Hanley, Ivan Sorokin, Shubham Toshniwal, Christof Henkel,
  Benedikt Schifferer, Wei Du, and Igor Gitman.
\newblock Aimo-2 winning solution: Building state-of-the-art mathematical
  reasoning models with openmathreasoning dataset.
\newblock \emph{arXiv preprint arXiv:2504.16891}, 2025.

\bibitem[Nan et~al.(2024)Nan, Xie, Zhou, Fan, Yang, Chen, Li, Yang, and
  Tai]{nan2024openvid1m}
Kepan Nan, Rui Xie, Penghao Zhou, Tiehan Fan, Zhenheng Yang, Zhijie Chen, Xiang
  Li, Jian Yang, and Ying Tai.
\newblock Openvid-1m: A large-scale high-quality dataset for text-to-video
  generation.
\newblock \emph{arXiv preprint arXiv:2407.02371}, 2024.

\bibitem[Nie et~al.(2025)Nie, Zhu, You, Zhang, Ou, Hu, Zhou, Lin, Wen, and
  Li]{nie2025large}
Shen Nie, Fengqi Zhu, Zebin You, Xiaolu Zhang, Jingyang Ou, Jun Hu, Jun Zhou,
  Yankai Lin, Ji-Rong Wen, and Chongxuan Li.
\newblock Large language diffusion models.
\newblock \emph{arXiv preprint arXiv:2502.09992}, 2025.

\bibitem[OpenAI(2023)]{gpt4}
OpenAI.
\newblock {GPT4} technical report.
\newblock \emph{arXiv preprint arXiv:2303.08774}, 2023.

\bibitem[OpenAI(2025)]{gptimage}
OpenAI.
\newblock Gpt-image-1.
\newblock https://openai.com/index/introducing-4o-image-generation/., 2025.

\bibitem[Panayotov et~al.(2015)Panayotov, Chen, Povey, and
  Khudanpur]{panayotov2015librispeech}
Vassil Panayotov, Guoguo Chen, Daniel Povey, and Sanjeev Khudanpur.
\newblock Librispeech: an asr corpus based on public domain audio books.
\newblock In \emph{2015 IEEE international conference on acoustics, speech and
  signal processing (ICASSP)}, pp.\  5206--5210. IEEE, 2015.

\bibitem[Qian et~al.(2025)Qian, Bocek-Rivele, Song, Tong, Yang, Lu, Hu, and
  Gan]{qian2025picobanana400k}
Yusu Qian, Eli Bocek-Rivele, Liangchen Song, Jialing Tong, Yinfei Yang, Jiasen
  Lu, Wenze Hu, and Zhe Gan.
\newblock Pico-banana-400k: A large-scale dataset for text-guided image
  editing, 2025.
\newblock URL \url{https://arxiv.org/abs/2510.19808}.

\bibitem[{Qwen Team}(2026)]{qwen3.5}
{Qwen Team}.
\newblock {Qwen3.5}: Towards native multimodal agents, February 2026.
\newblock URL \url{https://qwen.ai/blog?id=qwen3.5}.

\bibitem[Radford et~al.(2022)Radford, Kim, Xu, Brockman, McLeavey, and
  Sutskever]{radford2022whisper}
Alec Radford, Jong~Wook Kim, Tao Xu, Greg Brockman, Christine McLeavey, and
  Ilya Sutskever.
\newblock Robust speech recognition via large-scale weak supervision, 2022.
\newblock URL \url{https://arxiv.org/abs/2212.04356}.

\bibitem[Rein et~al.(2024)Rein, Hou, Stickland, Petty, Pang, Dirani, Michael,
  and Bowman]{rein2024gpqa}
David Rein, Betty~Li Hou, Asa~Cooper Stickland, Jackson Petty, Richard~Yuanzhe
  Pang, Julien Dirani, Julian Michael, and Samuel~R. Bowman.
\newblock {GPQA}: A graduate-level google-proof q\&a benchmark.
\newblock In \emph{First Conference on Language Modeling}, 2024.
\newblock URL \url{https://openreview.net/forum?id=Ti67584b98}.

\bibitem[Sahoo et~al.(2024)Sahoo, Arriola, Gokaslan, Marroquin, Rush, Schiff,
  Chiu, and Kuleshov]{sahoo2024simple}
Subham~Sekhar Sahoo, Marianne Arriola, Aaron Gokaslan, Edgar~Mariano Marroquin,
  Alexander~M Rush, Yair Schiff, Justin~T Chiu, and Volodymyr Kuleshov.
\newblock Simple and effective masked diffusion language models.
\newblock In \emph{{Proc. of Neural Information Processing Systems (NeurIPS)}},
  2024.

\bibitem[Shi et~al.(2025)Shi, Bai, Zhao, Chai, Yu, Wu, Song, Tong, Li, Li,
  et~al.]{shi2025muddit}
Qingyu Shi, Jinbin Bai, Zhuoran Zhao, Wenhao Chai, Kaidong Yu, Jianzong Wu,
  Shuangyong Song, Yunhai Tong, Xiangtai Li, Xuelong Li, et~al.
\newblock Muddit: Liberating generation beyond text-to-image with a unified
  discrete diffusion model.
\newblock \emph{arXiv preprint arXiv:2505.23606}, 2025.

\bibitem[Shi et~al.(2026)Shi, Wang, Guo, Wang, Zhang, Zhang, Guo, Hao, Xi,
  Yang, Xu, Zhou, and Lin]{Qwen3-ASR}
Xian Shi, Xiong Wang, Zhifang Guo, Yongqi Wang, Pei Zhang, Xinyu Zhang, Zishan
  Guo, Hongkun Hao, Yu~Xi, Baosong Yang, Jin Xu, Jingren Zhou, and Junyang Lin.
\newblock Qwen3-asr technical report.
\newblock \emph{arXiv preprint}, 2026.

\bibitem[Song et~al.(2025)Song, Zhang, Luo, Gao, Xia, Luo, Li, Yang, Yu, Qu,
  Fu, Su, Zhang, Huang, Wang, Yan, Jia, Liu, Ma, Zhang, Wu, and
  Zhou]{song2025seeddiffusionlargescalediffusion}
Yuxuan Song, Zheng Zhang, Cheng Luo, Pengyang Gao, Fan Xia, Hao Luo, Zheng Li,
  Yuehang Yang, Hongli Yu, Xingwei Qu, Yuwei Fu, Jing Su, Ge~Zhang, Wenhao
  Huang, Mingxuan Wang, Lin Yan, Xiaoying Jia, Jingjing Liu, Wei-Ying Ma,
  Ya-Qin Zhang, Yonghui Wu, and Hao Zhou.
\newblock Seed diffusion: A large-scale diffusion language model with
  high-speed inference, 2025.
\newblock URL \url{https://arxiv.org/abs/2508.02193}.

\bibitem[Sun et~al.(2023)Sun, Pan, Ge, Li, Duan, Wu, Zhang, Zhou, Qin, Wang,
  et~al.]{sun2023journeydb}
Keqiang Sun, Junting Pan, Yuying Ge, Hao Li, Haodong Duan, Xiaoshi Wu, Renrui
  Zhang, Aojun Zhou, Zipeng Qin, Yi~Wang, et~al.
\newblock Journeydb: A benchmark for generative image understanding.
\newblock \emph{Advances in neural information processing systems},
  36:\penalty0 49659--49678, 2023.

\bibitem[Tang et~al.(2024)Tang, Yang, Khademi, Liu, Zhu, and Bansal]{CoDi-2}
Zineng Tang, Ziyi Yang, Mahmoud Khademi, Yang Liu, Chenguang Zhu, and Mohit
  Bansal.
\newblock Codi-2: In-context interleaved and interactive any-to-any generation.
\newblock In \emph{Proceedings of the IEEE/CVF Conference on Computer Vision
  and Pattern Recognition (CVPR)}, pp.\  27425--27434, June 2024.

\bibitem[Team(2024)]{team2024chameleon}
Chameleon Team.
\newblock Chameleon: Mixed-modal early-fusion foundation models, 2024.
\newblock \emph{URL https://arxiv. org/abs/2405.09818}, 9\penalty0 (8), 2024.

\bibitem[Team(2026)]{team2026hyperclova}
NAVER Cloud HyperCLOVA~X Team.
\newblock Hyperclova x 8b omni.
\newblock \emph{arXiv preprint arXiv:2601.01792}, 2026.

\bibitem[Team(2025)]{team2025zimage}
Z-Image Team.
\newblock Z-image: An efficient image generation foundation model with
  single-stream diffusion transformer.
\newblock \emph{arXiv preprint arXiv:2511.22699}, 2025.

\bibitem[Teknium(2023)]{OpenHermes2_5}
Teknium.
\newblock Openhermes 2.5: An open dataset of synthetic data for generalist llm
  assistants, 2023.
\newblock URL \url{https://huggingface.co/datasets/teknium/OpenHermes-2.5}.

\bibitem[Tong et~al.(2024)Tong, Brown, Wu, Woo, Iyer, Akula, Yang, Yang,
  Middepogu, Wang, et~al.]{tong2024cambrian}
Peter Tong, Ellis Brown, Penghao Wu, Sanghyun Woo, Adithya Jairam~Vedagiri
  Iyer, Sai~Charitha Akula, Shusheng Yang, Jihan Yang, Manoj Middepogu, Ziteng
  Wang, et~al.
\newblock Cambrian-1: A fully open, vision-centric exploration of multimodal
  llms.
\newblock \emph{Advances in Neural Information Processing Systems},
  37:\penalty0 87310--87356, 2024.

\bibitem[TrillionLabs(2026)]{trida}
TrillionLabs.
\newblock Trida-7b: Block diffusion language model.
\newblock https://huggingface.co/trillionlabs/Trida-7B, 2026.

\bibitem[Wan et~al.(2025)Wan, Wang, Ai, Wen, Mao, Xie, Chen, Yu, Zhao, Yang,
  Zeng, Wang, Zhang, Zhou, Wang, Chen, Zhu, Zhao, Yan, Huang, Feng, Zhang, Li,
  Wu, Chu, Feng, Zhang, Sun, Fang, Wang, Gui, Weng, Shen, Lin, Wang, Wang,
  Zhou, Wang, Shen, Yu, Shi, Huang, Xu, Kou, Lv, Li, Liu, Wang, Zhang, Huang,
  Li, Wu, Liu, Pan, Zheng, Hong, Shi, Feng, Jiang, Han, Wu, and Liu]{wan2025}
Team Wan, Ang Wang, Baole Ai, Bin Wen, Chaojie Mao, Chen-Wei Xie, Di~Chen,
  Feiwu Yu, Haiming Zhao, Jianxiao Yang, Jianyuan Zeng, Jiayu Wang, Jingfeng
  Zhang, Jingren Zhou, Jinkai Wang, Jixuan Chen, Kai Zhu, Kang Zhao, Keyu Yan,
  Lianghua Huang, Mengyang Feng, Ningyi Zhang, Pandeng Li, Pingyu Wu, Ruihang
  Chu, Ruili Feng, Shiwei Zhang, Siyang Sun, Tao Fang, Tianxing Wang, Tianyi
  Gui, Tingyu Weng, Tong Shen, Wei Lin, Wei Wang, Wei Wang, Wenmeng Zhou, Wente
  Wang, Wenting Shen, Wenyuan Yu, Xianzhong Shi, Xiaoming Huang, Xin Xu, Yan
  Kou, Yangyu Lv, Yifei Li, Yijing Liu, Yiming Wang, Yingya Zhang, Yitong
  Huang, Yong Li, You Wu, Yu~Liu, Yulin Pan, Yun Zheng, Yuntao Hong, Yupeng
  Shi, Yutong Feng, Zeyinzi Jiang, Zhen Han, Zhi-Fan Wu, and Ziyu Liu.
\newblock Wan: Open and advanced large-scale video generative models.
\newblock \emph{arXiv preprint arXiv:2503.20314}, 2025.

\bibitem[Wang et~al.(2025{\natexlab{a}})Wang, Lai, Li, Zhang, Sun, Kang, Wu,
  Li, and Luo]{wang2025fudoki}
Jin Wang, Yao Lai, Aoxue Li, Shifeng Zhang, Jiacheng Sun, Ning Kang, Chengyue
  Wu, Zhenguo Li, and Ping Luo.
\newblock Fudoki: Discrete flow-based unified understanding and generation via
  kinetic-optimal velocities.
\newblock \emph{arXiv preprint arXiv:2505.20147}, 2025{\natexlab{a}}.

\bibitem[Wang et~al.(2025{\natexlab{b}})Wang, Gao, Gu, Pu, Cui, Wei, Liu, Jing,
  Ye, Shao, et~al.]{wang2025internvl3}
Weiyun Wang, Zhangwei Gao, Lixin Gu, Hengjun Pu, Long Cui, Xingguang Wei,
  Zhaoyang Liu, Linglin Jing, Shenglong Ye, Jie Shao, et~al.
\newblock Internvl3.5: Advancing open-source multimodal models in versatility,
  reasoning, and efficiency.
\newblock \emph{arXiv preprint arXiv:2508.18265}, 2025{\natexlab{b}}.

\bibitem[Wang et~al.(2024{\natexlab{a}})Wang, Zhang, Luo, Sun, Cui, Wang,
  Zhang, Wang, Li, Yu, et~al.]{wang2024emu3}
Xinlong Wang, Xiaosong Zhang, Zhengxiong Luo, Quan Sun, Yufeng Cui, Jinsheng
  Wang, Fan Zhang, Yueze Wang, Zhen Li, Qiying Yu, et~al.
\newblock Emu3: Next-token prediction is all you need.
\newblock \emph{arXiv preprint arXiv:2409.18869}, 2024{\natexlab{a}}.

\bibitem[Wang et~al.(2024{\natexlab{b}})Wang, Zhan, Liu, Zeng, Guo, Zheng,
  Zhang, Zhang, Zhang, and Wu]{wang2024maskgct}
Yuancheng Wang, Haoyue Zhan, Liwei Liu, Ruihong Zeng, Haotian Guo, Jiachen
  Zheng, Qiang Zhang, Xueyao Zhang, Shunsi Zhang, and Zhizheng Wu.
\newblock Maskgct: Zero-shot text-to-speech with masked generative codec
  transformer.
\newblock \emph{arXiv preprint arXiv:2409.00750}, 2024{\natexlab{b}}.

\bibitem[Wei et~al.(2026)Wei, Cheng, Jin, et~al.]{wei2026optmerge}
Yongxian Wei, Runxi Cheng, Weike Jin, et~al.
\newblock Optmerge: Unifying multimodal llm capabilities and modalities via
  model merging.
\newblock In \emph{International Conference on Learning Representations}, 2026.
\newblock URL \url{https://arxiv.org/abs/2505.19892}.

\bibitem[Wu et~al.(2025{\natexlab{a}})Wu, Li, Zhou, Lin, Gao, Yan, ming Yin,
  Bai, Xu, Chen, Chen, Tang, Zhang, Wang, Yang, Yu, Cheng, Liu, Li, Zhang,
  Meng, Wei, Ni, Chen, Cao, Peng, Qu, Wu, Wang, Yu, Wen, Feng, Xu, Wang, Zhang,
  Zhu, Wu, Cai, and Liu]{wu2025qwenimagetechnicalreport}
Chenfei Wu, Jiahao Li, Jingren Zhou, Junyang Lin, Kaiyuan Gao, Kun Yan, Sheng
  ming Yin, Shuai Bai, Xiao Xu, Yilei Chen, Yuxiang Chen, Zecheng Tang, Zekai
  Zhang, Zhengyi Wang, An~Yang, Bowen Yu, Chen Cheng, Dayiheng Liu, Deqing Li,
  Hang Zhang, Hao Meng, Hu~Wei, Jingyuan Ni, Kai Chen, Kuan Cao, Liang Peng,
  Lin Qu, Minggang Wu, Peng Wang, Shuting Yu, Tingkun Wen, Wensen Feng,
  Xiaoxiao Xu, Yi~Wang, Yichang Zhang, Yongqiang Zhu, Yujia Wu, Yuxuan Cai, and
  Zenan Liu.
\newblock Qwen-image technical report, 2025{\natexlab{a}}.
\newblock URL \url{https://arxiv.org/abs/2508.02324}.

\bibitem[Wu et~al.(2026)Wu, Zhang, Xue, Liu, Diao, Zhu, Luo, Han, and
  Xie]{wu2026fastdllm}
Chengyue Wu, Hao Zhang, Shuchen Xue, Zhijian Liu, Shizhe Diao, Ligeng Zhu, Ping
  Luo, Song Han, and Enze Xie.
\newblock Fast-d{LLM}: Training-free acceleration of diffusion {LLM} by
  enabling {KV} cache and parallel decoding.
\newblock In \emph{The Fourteenth International Conference on Learning
  Representations}, 2026.
\newblock URL \url{https://openreview.net/forum?id=3Z3Is6hnOT}.

\bibitem[Wu et~al.(2025{\natexlab{b}})Wu, Zheng, Yan, Xiao, Luo, Wang, Li,
  Jiang, Liu, Zhou, et~al.]{wu2025omnigen2}
Chenyuan Wu, Pengfei Zheng, Ruiran Yan, Shitao Xiao, Xin Luo, Yueze Wang, Wanli
  Li, Xiyan Jiang, Yexin Liu, Junjie Zhou, et~al.
\newblock Omnigen2: Exploration to advanced multimodal generation.
\newblock \emph{arXiv preprint arXiv:2506.18871}, 2025{\natexlab{b}}.

\bibitem[Wu et~al.(2024)Wu, Fei, Qu, Ji, and Chua]{next-gpt}
Shengqiong Wu, Hao Fei, Leigang Qu, Wei Ji, and Tat-Seng Chua.
\newblock {NE}x{T}-{GPT}: Any-to-any multimodal {LLM}.
\newblock In \emph{Proceedings of the International Conference on Machine
  Learning}, pp.\  53366--53397, 2024.

\bibitem[Xiao et~al.(2021)Xiao, Shang, Yao, and Chua]{xiao2021nextqa}
Junbin Xiao, Xindi Shang, Angela Yao, and Tat-Seng Chua.
\newblock Next-qa: Next phase of question-answering to explaining temporal
  actions.
\newblock In \emph{Proceedings of the IEEE/CVF conference on computer vision
  and pattern recognition}, pp.\  9777--9786, 2021.

\bibitem[Xie et~al.(2024)Xie, Mao, Bai, Zhang, Wang, Lin, Gu, Chen, Yang, and
  Shou]{xie2024showo1}
Jinheng Xie, Weijia Mao, Zechen Bai, David~Junhao Zhang, Weihao Wang,
  Kevin~Qinghong Lin, Yuchao Gu, Zhijie Chen, Zhenheng Yang, and Mike~Zheng
  Shou.
\newblock Show-o: One single transformer to unify multimodal understanding and
  generation.
\newblock \emph{arXiv preprint arXiv:2408.12528}, 2024.

\bibitem[Xie et~al.(2025)Xie, Yang, and Shou]{xie2025showo2}
Jinheng Xie, Zhenheng Yang, and Mike~Zheng Shou.
\newblock Show-o2: Improved native unified multimodal models.
\newblock \emph{arXiv preprint arXiv:2506.15564}, 2025.

\bibitem[Xin et~al.(2025)Xin, Qin, Luo, Zhu, Yan, Tai, Lei, Cao, Wang, Wang,
  Bai, Yu, Jiang, Pu, Chen, Zhuo, He, Luo, Li, Hu, Ye, Ye, Zhang, Xu, Wang, Li,
  Zhai, Xue, Fu, Liu, Qiao, and Liu]{xin2025luminadimooomnidiffusionlarge}
Yi~Xin, Qi~Qin, Siqi Luo, Kaiwen Zhu, Juncheng Yan, Yan Tai, Jiayi Lei, Yuewen
  Cao, Keqi Wang, Yibin Wang, Jinbin Bai, Qian Yu, Dengyang Jiang, Yuandong Pu,
  Haoxing Chen, Le~Zhuo, Junjun He, Gen Luo, Tianbin Li, Ming Hu, Jin Ye,
  Shenglong Ye, Bo~Zhang, Chang Xu, Wenhai Wang, Hongsheng Li, Guangtao Zhai,
  Tianfan Xue, Bin Fu, Xiaohong Liu, Yu~Qiao, and Yihao Liu.
\newblock Lumina-dimoo: An omni diffusion large language model for multi-modal
  generation and understanding, 2025.
\newblock URL \url{https://arxiv.org/abs/2510.06308}.

\bibitem[Xu et~al.(2023)Xu, Sun, Zheng, Geng, Zhao, Feng, Tao, and
  Jiang]{evloinstruct}
Can Xu, Qingfeng Sun, Kai Zheng, Xiubo Geng, Pu~Zhao, Jiazhan Feng, Chongyang
  Tao, and Daxin Jiang.
\newblock Wizardlm: Empowering large language models to follow complex
  instructions.
\newblock \emph{arXiv preprint arXiv:2304.12244}, 2023.

\bibitem[Xu et~al.(2025{\natexlab{a}})Xu, Guo, He, Hu, He, Bai, Chen, Wang,
  Fan, Dang, Zhang, Wang, Chu, and Lin]{Qwen2.5-Omni}
Jin Xu, Zhifang Guo, Jinzheng He, Hangrui Hu, Ting He, Shuai Bai, Keqin Chen,
  Jialin Wang, Yang Fan, Kai Dang, Bin Zhang, Xiong Wang, Yunfei Chu, and
  Junyang Lin.
\newblock Qwen2.5-omni technical report.
\newblock \emph{arXiv preprint arXiv:2503.20215}, 2025{\natexlab{a}}.

\bibitem[Xu et~al.(2025{\natexlab{b}})Xu, Guo, Hu, Chu, Wang, He, Wang, Shi,
  He, Zhu, et~al.]{xu2025qwen3omni}
Jin Xu, Zhifang Guo, Hangrui Hu, Yunfei Chu, Xiong Wang, Jinzheng He, Yuxuan
  Wang, Xian Shi, Ting He, Xinfa Zhu, et~al.
\newblock Qwen3-omni technical report.
\newblock \emph{arXiv preprint arXiv:2509.17765}, 2025{\natexlab{b}}.

\bibitem[Xu et~al.(2024)Xu, Jiang, Niu, Deng, Poovendran, Choi, and
  Lin]{xu2024magpie}
Zhangchen Xu, Fengqing Jiang, Luyao Niu, Yuntian Deng, Radha Poovendran, Yejin
  Choi, and Bill~Yuchen Lin.
\newblock Magpie: Alignment data synthesis from scratch by prompting aligned
  llms with nothing.
\newblock \emph{arXiv preprint arXiv:2406.08464}, 2024.

\bibitem[Yagcioglu et~al.(2018)Yagcioglu, Erdem, Erdem, and
  Ikizler-Cinbis]{yagcioglu2018recipeqa}
Semih Yagcioglu, Aykut Erdem, Erkut Erdem, and Nazli Ikizler-Cinbis.
\newblock Recipeqa: A challenge dataset for multimodal comprehension of cooking
  recipes.
\newblock In \emph{Proceedings of EMNLP}, 2018.
\newblock URL \url{https://aclanthology.org/D18-1166/}.

\bibitem[Yang et~al.(2025{\natexlab{a}})Yang, Li, Yang, Zhang, Hui, Zheng, Yu,
  Gao, Huang, Lv, et~al.]{yang2025qwen3}
An~Yang, Anfeng Li, Baosong Yang, Beichen Zhang, Binyuan Hui, Bo~Zheng, Bowen
  Yu, Chang Gao, Chengen Huang, Chenxu Lv, et~al.
\newblock Qwen3 technical report.
\newblock \emph{arXiv preprint arXiv:2505.09388}, 2025{\natexlab{a}}.

\bibitem[Yang et~al.(2022)Yang, Chen, Jiang, Chen, Wang, and
  Zhao]{yang2022visualhow}
Jinhui Yang, Xianyu Chen, Ming Jiang, Shi Chen, Louis Wang, and Qi~Zhao.
\newblock Visualhow: Multimodal problem solving.
\newblock In \emph{CVPR}, 2022.
\newblock URL \url{https://www-users.cse.umn.edu/~qzhao/visualhow.html}.

\bibitem[Yang et~al.(2025{\natexlab{b}})Yang, Tian, Li, Zhang, Shen, Tong, and
  Wang]{MMaDA}
Ling Yang, Ye~Tian, Bowen Li, Xinchen Zhang, Ke~Shen, Yunhai Tong, and Mengdi
  Wang.
\newblock {MM}a{DA}: Multimodal large diffusion language models.
\newblock In \emph{The Thirty-ninth Annual Conference on Neural Information
  Processing Systems}, 2025{\natexlab{b}}.
\newblock URL \url{https://openreview.net/forum?id=wczmXLuLGd}.

\bibitem[Yao et~al.(2024)Yao, Yu, Zhang, Wang, Cui, Zhu, Cai, Li, Zhao, He,
  et~al.]{yao2024minicpm}
Yuan Yao, Tianyu Yu, Ao~Zhang, Chongyi Wang, Junbo Cui, Hongji Zhu, Tianchi
  Cai, Haoyu Li, Weilin Zhao, Zhihui He, et~al.
\newblock Minicpm-v: A gpt-4v level mllm on your phone.
\newblock \emph{arXiv preprint arXiv:2408.01800}, 2024.

\bibitem[Ye et~al.(2025{\natexlab{a}})Ye, Yang, Goel, Huang, Zhu, Su, Lin,
  Cheng, Wan, Tian, et~al.]{ye2025omnivinci}
Hanrong Ye, Chao-Han~Huck Yang, Arushi Goel, Wei Huang, Ligeng Zhu, Yuanhang
  Su, Sean Lin, An-Chieh Cheng, Zhen Wan, Jinchuan Tian, et~al.
\newblock Omnivinci: Enhancing architecture and data for omni-modal
  understanding llm.
\newblock \emph{arXiv preprint arXiv:2510.15870}, 2025{\natexlab{a}}.

\bibitem[Ye et~al.(2025{\natexlab{b}})Ye, Gong, Gao, Fan, Wu, Bi, Bai, Shang,
  and Kong]{ye2025dreamvl}
Jiacheng Ye, Shansan Gong, Jiahui Gao, Junming Fan, Shuang Wu, Wei Bi, Haoli
  Bai, Lifeng Shang, and Lingpeng Kong.
\newblock {Dream-VL} \& {Dream-VLA}: Open vision-language and
  vision-language-action models with diffusion language model backbone,
  2025{\natexlab{b}}.
\newblock URL \url{https://arxiv.org/abs/2512.22615}.

\bibitem[Ye et~al.(2025{\natexlab{c}})Ye, Xie, Zheng, Gao, Wu, Jiang, Li, and
  Kong]{dream2025}
Jiacheng Ye, Zhihui Xie, Lin Zheng, Jiahui Gao, Zirui Wu, Xin Jiang, Zhenguo
  Li, and Lingpeng Kong.
\newblock Dream 7b, 2025{\natexlab{c}}.
\newblock URL \url{https://hkunlp.github.io/blog/2025/dream}.

\bibitem[Ye et~al.(2025{\natexlab{d}})Ye, He, Li, Lin, Yuan, Yan, Hou, and
  Yuan]{ye2025imgedit}
Yang Ye, Xianyi He, Zongjian Li, Bin Lin, Shenghai Yuan, Zhiyuan Yan, Bohan
  Hou, and Li~Yuan.
\newblock Imgedit: A unified image editing dataset and benchmark.
\newblock \emph{arXiv preprint arXiv:2505.20275}, 2025{\natexlab{d}}.

\bibitem[Ye et~al.(2025{\natexlab{e}})Ye, Zhu, Chan, Wang, Tan, Lei, Peng, Liu,
  Jin, Dai, et~al.]{ye2025llasa}
Zhen Ye, Xinfa Zhu, Chi-Min Chan, Xinsheng Wang, Xu~Tan, Jiahe Lei, Yi~Peng,
  Haohe Liu, Yizhu Jin, Zheqi Dai, et~al.
\newblock Llasa: Scaling train-time and inference-time compute for llama-based
  speech synthesis.
\newblock \emph{arXiv preprint arXiv:2502.04128}, 2025{\natexlab{e}}.

\bibitem[Yin et~al.(2026)Yin, Zhu, Gao, Zheng, Huang, Zhou, Yang, Liu, Chen,
  Guo, et~al.]{yin2026vllmomni}
Peiqi Yin, Jiangyun Zhu, Han Gao, Chenguang Zheng, Yongxiang Huang, Taichang
  Zhou, Ruirui Yang, Weizhi Liu, Weiqing Chen, Canlin Guo, et~al.
\newblock vllm-omni: Fully disaggregated serving for any-to-any multimodal
  models.
\newblock \emph{arXiv preprint arXiv:2602.02204}, 2026.

\bibitem[You et~al.(2025)You, Nie, Zhang, Hu, Zhou, Lu, Wen, and
  Li]{you2025lladav}
Zebin You, Shen Nie, Xiaolu Zhang, Jun Hu, Jun Zhou, Zhiwu Lu, Ji-Rong Wen, and
  Chongxuan Li.
\newblock Llada-v: Large language diffusion models with visual instruction
  tuning, 2025.
\newblock URL \url{https://arxiv.org/abs/2505.16933}.

\bibitem[You et~al.(2026)You, Zhang, Zhou, Li, and Wen]{you2026lladao}
Zebin You, Xiaolu Zhang, Jun Zhou, Chongxuan Li, and Ji-Rong Wen.
\newblock {LLaDA-o}: An effective and length-adaptive omni diffusion model,
  2026.
\newblock URL \url{https://arxiv.org/abs/2603.01068}.

\bibitem[Yu et~al.(2023)Yu, Lezama, Gundavarapu, Versari, Sohn, Minnen, Cheng,
  Birodkar, Gupta, Gu, et~al.]{yu2023language}
Lijun Yu, Jos{\'e} Lezama, Nitesh~B Gundavarapu, Luca Versari, Kihyuk Sohn,
  David Minnen, Yong Cheng, Vighnesh Birodkar, Agrim Gupta, Xiuye Gu, et~al.
\newblock Language model beats diffusion--tokenizer is key to visual
  generation.
\newblock \emph{arXiv preprint arXiv:2310.05737}, 2023.

\bibitem[Yu et~al.(2025)Yu, Ma, and Wang]{yu2025dimple}
Runpeng Yu, Xinyin Ma, and Xinchao Wang.
\newblock Dimple: Discrete diffusion multimodal large language model with
  parallel decoding, 2025.
\newblock URL \url{https://arxiv.org/abs/2505.16990}.

\bibitem[Yu et~al.(2019)Yu, Xu, Yu, Yu, Zhao, Zhuang, and
  Tao]{yu2019activitynet}
Zhou Yu, Dejing Xu, Jun Yu, Ting Yu, Zhou Zhao, Yueting Zhuang, and Dacheng
  Tao.
\newblock Activitynet-qa: A dataset for understanding complex web videos via
  question answering.
\newblock In \emph{Proceedings of the AAAI conference on artificial
  intelligence}, volume~33, pp.\  9127--9134, 2019.

\bibitem[Yue et~al.(2023)Yue, Ni, Zhang, Zheng, Liu, Zhang, Stevens, Jiang,
  Ren, Sun, et~al.]{yue2023mmmu}
Xiang Yue, Yuansheng Ni, Kai Zhang, Tianyu Zheng, Ruoqi Liu, Ge~Zhang, Samuel
  Stevens, Dongfu Jiang, Weiming Ren, Yuxuan Sun, et~al.
\newblock Mmmu: A massive multi-discipline multimodal understanding and
  reasoning benchmark for expert agi.
\newblock \emph{arXiv preprint arXiv:2311.16502}, 2023.

\bibitem[Zeng et~al.(2025)Zeng, Lv, Zheng, Hou, Chen, Xie, Wang, Yin, Zeng,
  Zhang, et~al.]{zeng2025glm}
Aohan Zeng, Xin Lv, Qinkai Zheng, Zhenyu Hou, Bin Chen, Chengxing Xie, Cunxiang
  Wang, Da~Yin, Hao Zeng, Jiajie Zhang, et~al.
\newblock Glm-4.5: Agentic, reasoning, and coding (arc) foundation models.
\newblock \emph{arXiv preprint arXiv:2508.06471}, 2025.

\bibitem[Zhan et~al.(2024)Zhan, Dai, Ye, Zhou, Zhang, Liu, Zhang, Yuan, Zhang,
  Li, Yan, Fu, Gui, Sun, Jiang, and Qiu]{anygpt}
Jun Zhan, Junqi Dai, Jiasheng Ye, Yunhua Zhou, Dong Zhang, Zhigeng Liu, Xin
  Zhang, Ruibin Yuan, Ge~Zhang, Linyang Li, Hang Yan, Jie Fu, Tao Gui,
  Tianxiang Sun, Yu-Gang Jiang, and Xipeng Qiu.
\newblock {A}ny{GPT}: Unified multimodal {LLM} with discrete sequence modeling.
\newblock In Lun-Wei Ku, Andre Martins, and Vivek Srikumar (eds.),
  \emph{Proceedings of the 62nd Annual Meeting of the Association for
  Computational Linguistics (Volume 1: Long Papers)}, pp.\  9637--9662,
  Bangkok, Thailand, August 2024. Association for Computational Linguistics.
\newblock \doi{10.18653/v1/2024.acl-long.521}.
\newblock URL \url{https://aclanthology.org/2024.acl-long.521/}.

\bibitem[Zhang et~al.(2025)Zhang, Guo, Yang, Yu, Zhang, Lei, Mai, Yan, Yang,
  Yang, et~al.]{zhang2025minimax}
Bowen Zhang, Congchao Guo, Geng Yang, Hang Yu, Haozhe Zhang, Heidi Lei, Jialong
  Mai, Junjie Yan, Kaiyue Yang, Mingqi Yang, et~al.
\newblock Minimax-speech: Intrinsic zero-shot text-to-speech with a learnable
  speaker encoder.
\newblock \emph{arXiv preprint arXiv:2505.07916}, 2025.

\bibitem[Zhang et~al.(2024{\natexlab{a}})Zhang, Dong, Zang, Cao, Qian, Chen,
  Guo, Duan, Wang, Ouyang, et~al.]{icx2_5}
Pan Zhang, Xiaoyi Dong, Yuhang Zang, Yuhang Cao, Rui Qian, Lin Chen, Qipeng
  Guo, Haodong Duan, Bin Wang, Linke Ouyang, et~al.
\newblock Internlm-xcomposer-2.5: A versatile large vision language model
  supporting long-contextual input and output.
\newblock \emph{arXiv preprint arXiv:2407.03320}, 2024{\natexlab{a}}.

\bibitem[Zhang et~al.(2024{\natexlab{b}})Zhang, Wu, Li, Li, Ma, Liu, and
  Li]{zhang2024llavavideo}
Yuanhan Zhang, Jinming Wu, Wei Li, Bo~Li, Zejun Ma, Ziwei Liu, and Chunyuan Li.
\newblock Llava-video: Video instruction tuning with synthetic data.
\newblock \emph{arXiv preprint arXiv:2410.02713}, 2024{\natexlab{b}}.

\bibitem[Zhao et~al.(2024)Zhao, Ma, Chen, Si, Wu, An, Yu, Zhang, Li, and
  Chang]{zhao2024ultraedit}
Haozhe Zhao, Xiaojian~Shawn Ma, Liang Chen, Shuzheng Si, Rujie Wu, Kaikai An,
  Peiyu Yu, Minjia Zhang, Qing Li, and Baobao Chang.
\newblock Ultraedit: Instruction-based fine-grained image editing at scale.
\newblock \emph{Advances in Neural Information Processing Systems},
  37:\penalty0 3058--3093, 2024.

\bibitem[Zhao et~al.(2025)Zhao, Zhang, Guo, Hu, Duan, Fu, Chng, Wang, Chen, Xu,
  et~al.]{zhao2025unified}
Shanshan Zhao, Xinjie Zhang, Jintao Guo, Jiakui Hu, Lunhao Duan, Minghao Fu,
  Yong~Xien Chng, Guo-Hua Wang, Qing-Guo Chen, Zhao Xu, et~al.
\newblock Unified multimodal understanding and generation models: Advances,
  challenges, and opportunities.
\newblock \emph{arXiv preprint arXiv:2505.02567}, 2025.

\bibitem[Zheng et~al.(2024)Zheng, Yuan, Yu, and
  Kong]{zheng_reparameterized_2024}
Lin Zheng, Jianbo Yuan, Lei Yu, and Lingpeng Kong.
\newblock A {Reparameterized} {Discrete} {Diffusion} {Model} for {Text}
  {Generation}.
\newblock \emph{arXiv preprint arXiv:2302.05737}, 2024.

\bibitem[Zhou et~al.(2024)Zhou, Yu, Babu, Tirumala, Yasunaga, Shamis, Kahn, Ma,
  Zettlemoyer, and Levy]{zhou2024transfusion}
Chunting Zhou, Lili Yu, Arun Babu, Kushal Tirumala, Michihiro Yasunaga, Leonid
  Shamis, Jacob Kahn, Xuezhe Ma, Luke Zettlemoyer, and Omer Levy.
\newblock Transfusion: Predict the next token and diffuse images with one
  multi-modal model.
\newblock \emph{arXiv preprint arXiv:2408.11039}, 2024.

\bibitem[Zhu et~al.(2025)Zhu, Wang, Chen, Liu, Ye, Gu, Tian, Duan, Su, Shao,
  et~al.]{zhu2025internvl3}
Jinguo Zhu, Weiyun Wang, Zhe Chen, Zhaoyang Liu, Shenglong Ye, Lixin Gu, Hao
  Tian, Yuchen Duan, Weijie Su, Jie Shao, et~al.
\newblock Internvl3: Exploring advanced training and test-time recipes for
  open-source multimodal models.
\newblock \emph{arXiv preprint arXiv:2504.10479}, 2025.

\end{thebibliography}
\bibliographystyle{dynin_refs}

\appendix
\newpage
\part{Appendix}
\etocsetnexttocdepth{subsection}
\localtableofcontents

\newpage
\clearpage

\section{Limitations and Discussions}
\subsection{Limitations}
\label{app:limitations}
\paragraph{Scope of native generation.}
\method{} represents text, image, video, and speech within a shared token space,
but our native joint generation experiments focus on interleaved text, image, and
speech outputs. Video is currently evaluated primarily as an input modality for
understanding rather than as dense video output generation. Extending the same
formulation to temporally coherent video generation is an important direction for
future work. We view this as a scope limitation rather than an architectural one:
video tokens are already processed by the same backbone and objective, while
native video generation mainly requires longer sequence budgets, stronger video
tokenization/detokenization, and large-scale temporal generation data.

\paragraph{Generalist-specialist trade-off.}
\method{} is not intended to replace the strongest modality-specific experts on
every individual benchmark. Models specialized for text reasoning, visual
understanding, image synthesis, ASR, or TTS can allocate their full capacity,
training data, and objectives to a single domain. In contrast, \method{} uses a
single 8B-scale backbone and a unified masked-diffusion objective across
modalities, without relying on external modality-specific generators. Therefore,
remaining gaps to expert models on some tasks should be interpreted as the
expected cost of native unification. The main result is that a single native
backbone can achieve strong and balanced performance across modalities while
supporting joint omnimodal generation.

\paragraph{Dependence on discrete tokenizers.}
Our formulation relies on pretrained discrete tokenizers and deterministic
detokenizers for non-text modalities. These modules define the discrete
representation space and map generated tokens back to pixels or waveforms, but
they do not perform conditional generation. All generative decisions are made by
the unified masked-diffusion backbone, which predicts tokens in the shared
vocabulary under a single objective, without modality-specific generators,
prediction heads, or reconstruction losses. Nevertheless, final output fidelity
can still be bounded by the quality of the underlying tokenizers and
detokenizers. For example, fine visual details, speech prosody, or long-range
temporal consistency may benefit from stronger tokenization schemes. This
limitation is orthogonal to the proposed backbone: improved tokenizers can be
substituted while preserving the same unified masked-diffusion formulation.

\paragraph{Training-scale ablations.}
A complete ablation of all training factors is difficult at the scale of
\method{}. Since the model is an 8B-scale omnimodal foundation model, retraining
it across many combinations of model sizes, data mixtures, stage schedules, and
merging ratios would require substantial compute beyond the scope of this work.
This is a practical constraint shared by large-scale unified and omnimodal
foundation-model studies, where resource constraints often limit training and
validation to a fixed large-scale setting~\citep{next-omni}. Accordingly, we
provide targeted analyses of the components most relevant to our training
challenges, rather than exhaustive training sweeps. For modality-disentangled
merging, we compare alternative merging strategies and interpolation ratios to
analyze capability preservation under vocabulary expansion. For scheduled
padding supervision, we provide training diagnostics, including qualitative
examples of early \texttt{[EOS]} shortcut behavior and stage-wise ASR/TTS WER
changes after enabling \texttt{[EOS]} supervision. These analyses support the
effectiveness of the proposed components, while broader scaling-law, data-mixture,
and stage-schedule studies remain valuable directions for future work.

\paragraph{Backbone generality.}
Our training-recipe controls evaluate the extension from MMaDA to \method{} and therefore establish effectiveness for this pretrained backbone and modality-expansion setting.
Applying the same recipe to other masked-diffusion or autoregressive backbones would require additional full training runs and remains an important direction for testing cross-backbone generality.

\subsection{Broader Impacts}
\label{app:impacts}
\method{} may benefit applications that require unified interaction across text, vision, and speech, such as accessible assistants, educational tools, creative storytelling, and multimodal content production. At the same time, stronger omnimodal generation can increase risks of synthetic media misuse, impersonation, misinformation, biased outputs, and inappropriate generated content. We therefore encourage responsible deployment with provenance tracking, content filtering, human oversight in high-stakes settings, and clear disclosure when generated media is used.

\section{Discrete Diffusion Modeling}
\label{app:discrete_diffusion}

In this subsection, we describe the masked diffusion framework underlying \method{}, providing formal definitions of the forward and reverse processes, the training objective, and the inference procedure.

\subsection{Forward and Reverse Processes}

Given a discrete token sequence $\mathbf{x}_0 = (x_1, \ldots, x_L)$ drawn from the joint vocabulary $\mathcal{V}$, the forward process independently corrupts each token by replacing it with a special \texttt{[MASK]} token $m$.
At masking ratio $t \in [0,1]$, the marginal distribution over the corrupted sequence $\mathbf{x}_t$ factorizes across positions as:
\begin{equation}
    q(\mathbf{x}_t \mid \mathbf{x}_0) = \prod_{i=1}^{L} q(x_{t,i} \mid x_{0,i}), \qquad
    q(x_{t,i} \mid x_{0,i}) = \begin{cases} 1 - t, & x_{t,i} = x_{0,i}, \\ t, & x_{t,i} = m, \end{cases}
    \label{eq:forward}
\end{equation}
so each token is independently masked with probability $t$ and left unchanged with probability $1-t$. At $t=1$ all tokens are masked; at $t=0$ the sequence is fully observed.

The reverse process learns to recover $\mathbf{x}_0$ from $\mathbf{x}_t$.
For $0 \le s < t \le 1$, the true reverse posterior factorizes across positions and is given by:
\begin{equation}
    q(x_{s,i} \mid x_{t,i},\, x_{0,i}) =
    \begin{cases}
        \mathrm{Cat}\!\left(x_{s,i};\; x_{t,i}\right),
        & x_{t,i} \neq m, \\[6pt]
        \mathrm{Cat}\!\left(x_{s,i};\; \dfrac{(1-\alpha_s)\,m + (\alpha_s - \alpha_t)\,x_{0,i}}{1 - \alpha_t}\right),
        & x_{t,i} = m,
    \end{cases}
    \label{eq:reverse_posterior}
\end{equation}
where $\alpha_t = 1 - t$ under the linear schedule.
For an unmasked token the posterior is deterministic ($x_{s,i} = x_{t,i}$); for a masked token it interpolates between remaining masked (weight $(1-\alpha_s)/(1-\alpha_t)$) and recovering the clean token (weight $(\alpha_s-\alpha_t)/(1-\alpha_t)$).
Since $x_{0,i}$ is unknown at inference, the parameterized reverse model replaces it with the model's prediction:
\begin{equation}
    p_\theta(x_{s,i} \mid \mathbf{x}_t) =
    \begin{cases}
        \mathrm{Cat}\!\left(x_{s,i};\; x_{t,i}\right),
        & x_{t,i} \neq m, \\[6pt]
        \mathrm{Cat}\!\left(x_{s,i};\; \dfrac{(1-\alpha_s)\,m + (\alpha_s - \alpha_t)\,p_\theta(x_{0,i} \mid \mathbf{x}_t)}{1 - \alpha_t}\right),
        & x_{t,i} = m,
    \end{cases}
    \label{eq:reverse_param}
\end{equation}
where $p_\theta(x_{0,i} \mid \mathbf{x}_t)$ is the predicted distribution over $\mathcal{V}$ for position $i$ conditioned on the full partially masked sequence $\mathbf{x}_t$~\citep{sahoo2024simple, nie2025large}.

\subsection{Training}

The backbone Transformer $p_\theta$ is trained as a mask predictor: given a corrupted sequence $\mathbf{x}_t$ sampled from the forward process, it predicts the original token $x_{0,i}$ at every masked position simultaneously.
Following prior work~\citep{sahoo2024simple,nie2025large}, the training objective is a cross-entropy loss computed only over masked positions, derived as a variational upper bound on the negative log-likelihood:
\begin{equation}
    \mathcal{L}(\theta)
    = -\mathbb{E}_{t,\, \mathbf{x}_0,\, \mathbf{x}_t}
    \left[
        \frac{1}{t}
        \sum_{i=1}^{L}
        \mathbb{I}\!\left[x_{t,i} = m\right]
        \log p_\theta(x_{0,i} \mid \mathbf{x}_t)
    \right],
    \label{eq:training_obj}
\end{equation}
where $t \sim \mathcal{U}(0,1)$, $\mathbf{x}_0$ is sampled from the training data, and $\mathbf{x}_t \sim q(\mathbf{x}_t \mid \mathbf{x}_0)$ from Eq.~\eqref{eq:forward}.
The weighting factor $1/t$ arises from the continuous-time variational lower bound~\citep{sahoo2024simple}, and it has been shown that $\mathcal{L}(\theta)$ upper-bounds the negative log-likelihood of the model distribution:
\begin{equation}
    -\mathbb{E}_{\mathbf{x}_0 \sim p_\mathrm{data}}\!\left[\log p_\theta(\mathbf{x}_0)\right] \leq \mathcal{L}(\theta).
\end{equation}
Unmasked tokens do not contribute to the loss, as their predictions are trivially correct due to the carry-over property of the reverse process (Eq.~\eqref{eq:reverse_param}).
This objective is shared uniformly across all modalities in the joint vocabulary $\mathcal{V}$.

\subsection{Inference}
At inference time, generation begins from a fully masked target sequence $\mathbf{x}_1 = (m, \ldots, m)$ of a pre-specified length, and the reverse process is discretized into $T$ steps from $t=1$ to $t=0$. At each step from time $t$ to $s = t - 1/T$, the backbone predicts a categorical distribution $p_\theta(x_{0,i} \mid \mathbf{x}_t)$ over $\mathcal{V}$ for all masked positions in parallel, and a subset of tokens are committed while the rest are remasked for the next step. In principle, a fraction $s/t$ of predicted tokens should be remasked at random to align the reverse process with the forward process~\citep{nie2025large}; in practice, \textit{low-confidence remasking}~\citep{Chang_2022_CVPR,nie2025large} is adopted, where the $\lfloor L(1-s) \rfloor$ currently masked tokens with the lowest predicted probability are remasked while the remaining high-confidence predictions are committed and held fixed.
To further respect sequential structure, \textit{block-wise decoding}~\citep{arriola2025block} divides the target sequence into contiguous blocks that are decoded left to right, with masked tokens within each block refined in parallel over multiple diffusion steps conditioned on all previously committed blocks.

\section{Related Work}
\label{sup:rw}

\begin{table*}[t]
\footnotesize
\centering
\caption{
Supported modality capabilities and architectural characteristics of representative perception-centric and unified perception--generation models.
\textit{Diff.} denotes diffusion-based generation and \textit{FM} denotes flow-matching-based generation.
Shaded cells indicate omnimodal capability (supporting text, vision, and speech concurrently) within each task group (Understanding or Generation). Entries written in parentheses, such as (+\,Diff.), indicate auxiliary external generation modules attached to the primary architecture.
For Omni-Diffusion, the image-editing entry denotes reported image inpainting rather than instruction-driven editing.
}
\label{tab:model_modalities_integrated}
\resizebox{\linewidth}{!}{
\begin{tabular}{llccccccccc}
\toprule
& & \multicolumn{4}{c}{Understanding Tasks} & \multicolumn{3}{c}{Generation Tasks} \\
\cmidrule(lr){3-6} \cmidrule(lr){7-9}
Methods & Modeling & Textual Reasoning & Image & Video & Speech & Image & Image Editing & Speech \\
\midrule
\multicolumn{9}{l}{\textbf{Perception-centric}} \\
\midrule
EMOVA-7B~\citep{chen2024emova} & AR &
\cellcolor{undomni} &
\cellcolor{undomni}\checkmark &
\cellcolor{undomni} &
\cellcolor{undomni}\checkmark &
 &  & \checkmark \\

VITA-1.5~\citep{fu2025vita} & AR &
\cellcolor{undomni} &
\cellcolor{undomni}\checkmark &
\cellcolor{undomni}\checkmark &
\cellcolor{undomni}\checkmark &
 &  & \checkmark \\

MiniCPM-o-2.6~\citep{yao2024minicpm} & AR &
\cellcolor{undomni} &
\cellcolor{undomni}\checkmark &
\cellcolor{undomni}\checkmark &
\cellcolor{undomni}\checkmark &
 &  & \checkmark \\

Baichuan-Omni-1.5~\citep{li2025baichuan} & AR &
\cellcolor{undomni}\checkmark &
\cellcolor{undomni}\checkmark &
\cellcolor{undomni}\checkmark &
\cellcolor{undomni}\checkmark &
 &  & \checkmark \\

OpenOmni~\citep{luo2025openomni} & AR &
\cellcolor{undomni} &
\cellcolor{undomni}\checkmark &
\cellcolor{undomni}\checkmark &
\cellcolor{undomni}\checkmark &
 &  & \checkmark \\

OmniVinci~\citep{ye2025omnivinci} & AR &
\cellcolor{undomni} &
\cellcolor{undomni}\checkmark &
\cellcolor{undomni}\checkmark &
\cellcolor{undomni}\checkmark &
 &  & \checkmark \\

Qwen2.5-Omni-7B~\citep{Qwen2.5-Omni} & AR &
\cellcolor{undomni}\checkmark &
\cellcolor{undomni}\checkmark &
\cellcolor{undomni}\checkmark &
\cellcolor{undomni}\checkmark &
 &  & \checkmark \\
\midrule
\multicolumn{9}{l}{\textbf{Compositional Unified}} \\
\midrule

NextGPT~\citep{next-gpt} & AR (+ Diff.) &
\cellcolor{undomni} &
\cellcolor{undomni}\checkmark &
\cellcolor{undomni}\checkmark &
\cellcolor{undomni}\checkmark &
\cellcolor{genomni}\checkmark &
\cellcolor{genomni} &
\cellcolor{genomni}\checkmark \\

CoDi-2~\citep{CoDi-2} & AR (+ Diff.) &
\cellcolor{undomni} &
\cellcolor{undomni}\checkmark &
\cellcolor{undomni}\checkmark &
\cellcolor{undomni}\checkmark &
\cellcolor{genomni}\checkmark &
\cellcolor{genomni}\checkmark &
\cellcolor{genomni} \\

DreamLLM~\citep{dong2023dreamllm} & AR (+ Diff.) &
 & \checkmark &  &  &
\checkmark &  &  \\

SEED-X~\citep{ge2024seed} & AR (+ Diff.) &
 & \checkmark &  &  &
\checkmark & \checkmark &  \\

UniWorld-V1~\citep{lin2025uniworld} & AR (+ Diff.) &
 & \checkmark &  &  &
\checkmark & \checkmark &  \\

HyperCLOVAX-8B-Omni~\citep{team2026hyperclova} & AR (+ Diff.) &
\cellcolor{undomni}\checkmark &
\cellcolor{undomni}\checkmark &
\cellcolor{undomni}\checkmark &
\cellcolor{undomni}\checkmark &
\cellcolor{genomni}\checkmark &
\cellcolor{genomni}\checkmark &
\cellcolor{genomni}\checkmark \\

\midrule
\multicolumn{9}{l}{\textbf{Native Unified}} \\
\midrule
AnyGPT~\citep{anygpt} & AR &
\cellcolor{undomni} &
\cellcolor{undomni}\checkmark &
\cellcolor{undomni} &
\cellcolor{undomni}\checkmark &
\cellcolor{genomni}\checkmark &
\cellcolor{genomni} &
\cellcolor{genomni}\checkmark \\

Chameleon-7B~\citep{team2024chameleon} & AR &
\checkmark & \checkmark &  &  & \checkmark &  &  \\

Emu3~\citep{wang2024emu3} & AR &
 & \checkmark & \checkmark &  & \checkmark &  &  \\

Janus-Pro~\citep{chen2025januspro} & AR &
 & \checkmark &  &  & \checkmark &  &  \\

BAGEL~\citep{bagle} & AR + FM &
 & \checkmark &  &  & \checkmark & \checkmark &  \\

Show-o2~\citep{xie2025showo2} & AR + FM &
\checkmark & \checkmark & \checkmark &  & \checkmark &  &  \\

MMaDA~\citep{MMaDA} & Discrete Diff. &
\checkmark & \checkmark &  &  & \checkmark &  &  \\

Omni-Diffusion~\citep{li2026omnidiffusion} & Discrete Diff. &
\cellcolor{undomni} & \cellcolor{undomni}\checkmark &
\cellcolor{undomni} & \cellcolor{undomni}\checkmark &
\cellcolor{genomni}\checkmark & \cellcolor{genomni}\checkmark &
\cellcolor{genomni}\checkmark \\

Fudoki~\citep{wang2025fudoki} & Discrete FM &
 & \checkmark &  &  & \checkmark &  &  \\

NExT-OMNI~\citep{next-omni} & Discrete FM &
\cellcolor{undomni} &
\cellcolor{undomni}\checkmark &
\cellcolor{undomni}\checkmark &
\cellcolor{undomni}\checkmark &
\cellcolor{genomni}\checkmark &
\cellcolor{genomni}\checkmark &
\cellcolor{genomni}\checkmark \\

LaViDa-O~\citep{li2025lavidao} & Discrete Diff. &
 & \checkmark &  &  & \checkmark & \checkmark &  \\

Lumina-DiMOO~\citep{xin2025luminadimooomnidiffusionlarge} & Discrete Diff. &
 & \checkmark &  &  & \checkmark & \checkmark &  \\

\rowcolor{oursrow}
\textbf{\method{} (Ours)} & Discrete Diff. &
\cellcolor{oursrow}\checkmark &
\cellcolor{oursrow}\checkmark &
\cellcolor{oursrow}\checkmark &
\cellcolor{oursrow}\checkmark &
\cellcolor{oursrow}\checkmark &
\cellcolor{oursrow}\checkmark &
\cellcolor{oursrow}\checkmark \\
\bottomrule
\end{tabular}
}
\end{table*}

\subsection{Diffusion Language Models}

Diffusion language models (DLMs) have recently emerged as an alternative to autoregressive (AR) language models by replacing left-to-right decoding with iterative denoising over token sequences~\citep{austin_structured_2021, li_diffusion-lm_2022, gong_diffuseq_2022, lin_text_2023}. By relaxing strict causal ordering, DLMs enable any-order generation, flexible intermediate editing, and improved global coherence~\citep{kim2025train, khanna2025mercury, song2025seeddiffusionlargescalediffusion}, while also supporting controllable generation and parallel token updates for enhanced efficiency~\citep{li_diffusion-lm_2022, kim2025dont, wu2026fastdllm, liu2025dllmcacheacceleratingdiffusionlarge, ma2025dkvcache}.
Among various formulations, discrete diffusion models instantiate DLMs directly over categorical token spaces, avoiding continuous relaxations while remaining compatible with standard language modeling objectives~\citep{he_diffusionbert_2023, zheng_reparameterized_2024, sahoo2024simple}. In particular, masked (absorbing-state) diffusion models formulate generation as iterative mask prediction, where subsets of tokens are replaced with \texttt{[MASK]} and reconstructed conditioned on the remaining context~\citep{he_diffusionbert_2023, sahoo2024simple, nie2025large}. This discrete denoising paradigm naturally supports bidirectional context modeling and parallel token refinement, making it well-suited for flexible and globally consistent generation.
At scale, LLaDA~\citep{nie2025large} and Dream~\citep{dream2025} demonstrate that masked diffusion can be trained into a strong general-purpose language model competitive with AR baselines. Building on masked diffusion foundations, recent works extend diffusion-based modeling to multimodal settings. For example, LLaDA-V~\citep{you2025lladav}, Dimple~\citep{yu2025dimple}, and LaViDa~\citep{li2025lavida} explore discrete diffusion for vision-language modeling. MMaDA~\citep{MMaDA} and Lumina-DiMOO~\citep{xin2025luminadimooomnidiffusionlarge} propose multimodal large diffusion language models for unified understanding and generation. Tri-modal masked-diffusion work includes text--image--audio modeling studies~\citep{bethune2026trimodal} and Omni-Diffusion~\citep{li2026omnidiffusion}, which supports text, image, and speech tasks under one masked-token objective and paired text--speech responses. Our focus is capability-preserving expansion to video understanding and broad reasoning, vision, and speech tasks, together with native interleaved text--image--speech output.

\subsection{Perception-centric and Unified Models}

We review three lines of related work: 
(1) perception-centric models, 
(2) compositional unified models, and 
(3) native unified models. Detailed model-wise comparisons are provided in Tab.~\ref{tab:model_modalities_integrated}.

\begin{figure}[t]
    \centering
    \includegraphics[width=1\linewidth]{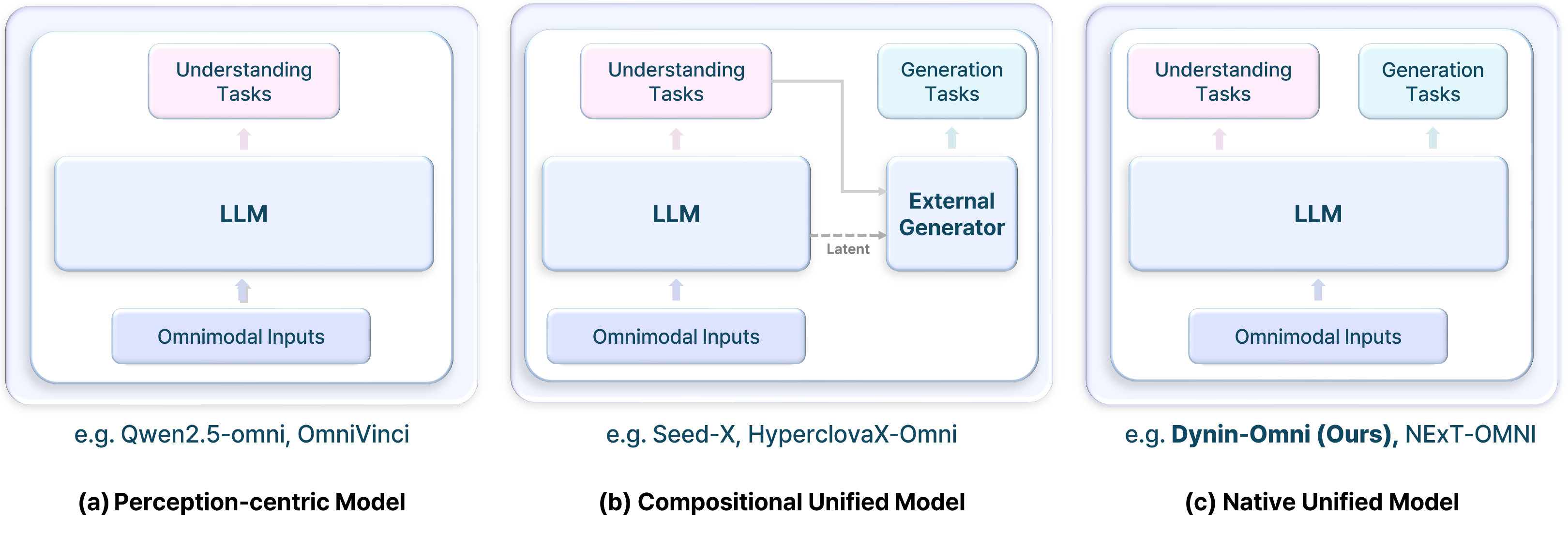}
    \caption{
    Architectural comparison of omnimodal modeling paradigms.
    (a) Perception-centric models focus on multimodal understanding with limited generation.
    (b) Compositional unified models rely on external modality-specific generators.
    (c) \method{} adopts a native unified formulation with shared token-level modeling.
    }
    \label{fig:arch_comparison}
\end{figure}

\paragraph{\textbf{Perception-centric models.}} 
Beyond traditional vision–language models~\citep{chen2024internvl,lu2024ovis,li2024llavaonevision} designed for visual perception tasks such as visual question answering (VQA)~\citep{yue2023mmmu,pope}, recent systems—including EMOVA~\citep{chen2024emova}, VITA-1.5~\citep{fu2025vita}, MiniCPM-o~\citep{yao2024minicpm},  Baichuan-Omni~\citep{li2025baichuan}, OpenOmni~\citep{luo2025openomni}, OmniVinci~\citep{ye2025omnivinci}, and Qwen-Omni~\citep{Qwen2.5-Omni,xu2025qwen3omni}—extend multimodal perception beyond vision and language to additional modalities such as video, speech, and audio (Fig.~\ref{fig:arch_comparison}(a)). 
These models support omnimodal inputs while also generating speech or audio responses. However, their architectures remain largely optimized for perception and understanding. In most cases, speech generation is implemented through auxiliary decoders or modality-specific pipelines (\eg codec or TTS-style modules) rather than through a unified generative modeling framework. Consequently, these systems primarily function as omnimodal understanding models rather than fully unified omnimodal models.

\paragraph{\textbf{Compositional unified models.}}
Compositional unified models integrate multimodal perception and generation 
by coordinating a shared backbone with modality-specific generators (Fig.~\ref{fig:arch_comparison}(b)). 
Rather than modeling all modalities within a single generative process, 
these systems delegate generation to auxiliary modules specialized for 
each modality.
Early compositional unified models such as DreamLLM~\citep{dong2023dreamllm}, 
SEED-X~\citep{ge2024seed}, and UniWorld-V1~\citep{lin2025uniworld} 
combine a multimodal backbone with external modality-specific generators, 
typically focusing on visual generation tasks such as image synthesis or editing.
Another line of work extends this compositional design toward broader 
omnimodal capabilities. For example, NeXT-GPT~\citep{next-gpt} and CoDi-2~\citep{CoDi-2} 
support both visual and speech generation by coupling a multimodal language 
model with multiple diffusion-based generators.
More recently, HyperCLOVAX-8B-Omni~\citep{team2026hyperclova} further advances 
this paradigm by enabling text, vision, and speech as both inputs and outputs, 
forming an omnimodal system. However, its generation still relies on heavy 
external diffusion generators such as FLUX~\citep{flux2024} rather than being 
natively modeled within the backbone, which increases system complexity and 
can introduce orchestration failures during inference.

\paragraph{\textbf{Native unified models.}}
Native unified models aim to directly model multimodal perception and generation 
within a single backbone without relying on external generators 
(Fig.~\ref{fig:arch_comparison}(c)). 
Early native unified models primarily focus on image--text modeling using 
autoregressive formulations. For example, AnyGPT~\citep{anygpt}, 
Chameleon~\citep{team2024chameleon}, Emu3~\citep{wang2024emu3}, and 
Janus-Pro~\citep{chen2025januspro} extend language modeling to multimodal tokens, 
allowing the backbone to autoregressively generate text and visual 
representations within a shared token space.
Subsequent works explore non-autoregressive generative paradigms to improve 
efficiency and scalability. Models such as BAGEL~\citep{bagle} and 
Show-o2~\citep{xie2025showo2} combine autoregressive reasoning with diffusion 
or flow-matching-based generation, while MMaDA~\citep{MMaDA}, 
Fudoki~\citep{wang2025fudoki}, LaViDa-O~\citep{li2025lavidao}, and 
Lumina-DiMOO~\citep{xin2025luminadimooomnidiffusionlarge} adopt fully discrete 
diffusion or flow-based formulations to directly model multimodal token 
generation. Omni-Diffusion~\citep{li2026omnidiffusion} also models text, image, and speech with a unified masked-token objective.
While recent work such as NExT-OMNI~\citep{next-omni} moves toward unified omnimodal perception and generation across multiple modalities, its training objective combines flow-matching generation with additional modality-specific reconstruction losses, introducing heterogeneous objectives that partially weaken the simplicity of fully unified modeling.
Like Omni-Diffusion, \method{} models modality-specific discrete codes with a single masked-token objective and backbone.
This unifies sequence-level modeling and generation across modalities while leaving their low-level tokenization spaces modality-specific.
Our contribution additionally examines retention of inherited reasoning and visual capabilities during expansion to video and speech, and native interleaved text--image--speech output.

\section{Architecture}
\label{sec:architecture}

\subsection{Backbone}
\label{subsubsec:backbone}

\method{} builds upon the masked-diffusion paradigm introduced in LLaDA~\citep{nie2025large}
and adopts MMaDA~\citep{MMaDA} as its architectural foundation, extending it toward full omnimodal capability.
Beyond the original vision–language modeling scope of MMaDA,
we incorporate video understanding, speech understanding and generation,
and introduce image editing,
all within a unified masked token prediction framework.
To support this expansion, we extend the discrete vocabulary to incorporate speech tokens and adapt the embedding lookup table and language modeling heads accordingly, while retaining a unified Transformer backbone without introducing modality-specific subnetworks. This architectural continuity enables scalable training through model merging and stage-wise optimization (\S\ref{sec:training_recipe}) without backbone fragmentation across modalities.

\begin{table}[t]
\centering
\footnotesize
\setlength{\tabcolsep}{2pt}
\renewcommand{\arraystretch}{1.5}
\caption{
Input--output template families across training stages.
{\color{red}Red} tokens denote masked target responses that are supervised throughout the stages.
{\color{blue}Blue} tokens indicate the global termination token \texttt{<EOS>},
which also serves as padding token in masked diffusion language models,
and is supervised only from Stage~2 onward.
}
\label{tab:global_templates}
\begin{adjustbox}{max width=\linewidth}
\begin{tabular}{l | ccc | l}
\toprule
\textbf{Template Family} & \textbf{S1} & \textbf{S2} & \textbf{S3} & \textbf{Input-Output Template} \\
\midrule
Video $\rightarrow$ Text &
\checkmark & \checkmark & \checkmark &
$\underbrace{\texttt{<image> \ldots <image>}}_{\text{video frames}}
\;\texttt{<|user|>}\;\{\text{prompt}\}\;
\texttt{<|assistant|>}\;{\color{red}\{\text{response}\}}\;
{\color{blue}\texttt{<EOS>}}$ \\

Speech $\rightarrow$ Text &
\checkmark & \checkmark & \checkmark &
$\texttt{<|startofspeech|>}\;\{\text{speech}\}\; \texttt{<|endofspeech|>}
\;\texttt{<|startoftext|>}\;{\color{red}\{\text{text}\}}\;
{\color{blue}\texttt{<EOS>}}$ \\

Text $\rightarrow$ Speech &
\checkmark & \checkmark & \checkmark &
$\texttt{<|startoftext|>}\;\{\text{text}\}\;\texttt{<|endoftext|>}
\;\texttt{<|startofspeech|>}\;{\color{red}\{\text{speech}\}}\; \texttt{<|endofspeech|>}\;
{\color{blue}\texttt{<EOS>}}$ \\

Text Chat &
 & \checkmark & \checkmark &
$\texttt{<|user|>}\;\{\text{prompt}\}\;
\texttt{<|assistant|>}\;
{\color{red}\{\text{response}\}}\;
{\color{blue}\texttt{<EOS>}}$ \\

Image $\rightarrow$ Text &
 & \checkmark & \checkmark &
$\texttt{<image>}\;
\texttt{<|user|>}\;\{\text{prompt}\}\;
\texttt{<|assistant|>};{\color{red}\{\text{response}\}}\;
{\color{blue}\texttt{<EOS>}}$ \\

Text $\rightarrow$ Image &
 & \checkmark & \checkmark &
$\texttt{<|startoftext|>}\;\{\text{instruction}\}\;
\texttt{<|endoftext|>}$
${\color{red}\text{\{image\}}}$ \\

Image $\rightarrow$ Image &
 & \checkmark & \checkmark &
$\texttt{<image>}\;
\texttt{<|startoftext|>}\;\{\text{instruction}\}\;
\texttt{<|endoftext|>}$
${\color{red}\text{\{image\}}}$ \\

Omni Gen. &
&
&
$\checkmark$ &
\begin{tabular}[c]{@{}l@{}}
$\texttt{<image>}\;\texttt{<|startoftext|>}\;\{\text{instruction}\}\;\texttt{<|endoftext|>}$ \\
${\color{red}\{\text{image}\}}\;
 \texttt{<|startoftext|>}{\color{red}\{\text{Text}\}}{\color{blue}\texttt{<EOS>}}\;
 \texttt{<|startofspeech|>}{\color{red}\{\text{Speech}\}}\texttt{<|endofspeech|>}{\color{blue}\texttt{<EOS>}}$
\end{tabular} \\

Thinking-mode &
 &  & \checkmark &
$\texttt{<|user|>}\;\{\text{prompt}\}\;
{\color{darkgreen}\texttt{\textbackslash think}\;|\;\texttt{\textbackslash no\_think}}
\;\texttt{<|assistant|>}\;
{\color{red}\{\text{response}\}}\;
{\color{blue}\texttt{<EOS>}}$ \\

\bottomrule
\end{tabular}
\end{adjustbox}
\end{table}

\subsection{Omnimodal Tokenization}
\label{app:omnimodal_tokenization}
\method{} supports textual, visual, and speech inputs within a unified token-based formulation. Specifically, images, videos, and speech signals are mapped into a shared discrete token space using modality-specific pretrained tokenizers. This enables all modalities to be processed under a single masked diffusion backbone, sharing the same embedding lookup table and Transformer architecture.
Throughout training, both tokenizers and detokenizers remain frozen to preserve stable representations. 
Importantly, the detokenizer is a lightweight reconstruction module that deterministically maps discrete tokens back to modality space, without performing iterative generation or introducing additional learnable generative stages. By representing all modalities in the same token space, omnimodal interactions are modeled directly at the token level under a unified objective. This streamlined design reduces the need for separate generation pipelines or complex modality-specific objectives often found in existing omnimodal models~\citep{next-omni,team2026hyperclova}.

\paragraph{\textbf{Text.}}
Built upon the LLaDA~\citep{nie2025large} architecture, \method{} retains its original text tokenizer (vocabulary size 126,080) to preserve the backbone’s tokenization scheme, while introducing a small set of task-oriented special tokens (\eg \texttt{<think>} and modality delimiters) to support structured reasoning and multimodal interaction. Beyond modality-specific tokenization, the central design of \method{} lies in mapping all modality tokens into a shared discrete vocabulary and processing them through a single embedding matrix and Transformer backbone. This unified token space enables multimodal interactions to occur directly at the token level under a single masked diffusion objective, integrating cross-modal reasoning and generation without relying on late-stage fusion or modality-specific orchestration.

\paragraph{\textbf{Image.}}
For image inputs, we follow Show-o~\citep{xie2024showo1} and MMaDA~\citep{MMaDA} and adopt a MAGVIT-v2~\citep{yu2023language}-style
vector-quantized image tokenizer pretrained jointly with Show-o~\citep{xie2024showo1}.
The tokenizer consists of a convolutional encoder, a discrete codebook,
and a corresponding decoder, and is trained to reconstruct images
from discrete latent representations.
Given an input image $x^I \in \mathbb{R}^{H \times W \times C}$,
the encoder downsamples the image by a fixed factor $f$
(using strided convolutions) and maps it into a latent feature map
of spatial resolution $\frac{H}{f} \times \frac{W}{f}$.
Each latent vector is then quantized to its nearest entry
in a shared visual codebook of size $|\mathcal{V}_{\text{vision}}| = 8192$,
resulting in a grid of discrete visual tokens.
This grid is flattened in raster-scan order to form a 1D token sequence
of length $\frac{H}{f} \cdot \frac{W}{f}$, which is subsequently processed
by the backbone language model together with tokens from other modalities.
For image generation and editing, a pretrained reconstruction module from MAGVIT-v2
maps the predicted discrete visual tokens back to pixel space to produce the output image.
Importantly, this module is frozen and serves solely as a token-to-image reconstruction component;
all sequence-level generative modeling is performed within the masked diffusion backbone.

\paragraph{\textbf{Video.}}
To maintain a unified visual token space across images and videos and to avoid introducing modality-specific tokenization branches, we design the video tokenizer by reusing the pretrained image tokenizer. Rather than learning a separate video tokenizer, we treat a video as a temporally ordered sequence of images while preserving the same vision token space $\mathcal{V}_{\text{vision}}$. 
Following previous works~\citep{bertasius2021space,arnab2021vivit}, we apply the pretrained image tokenizer independently to sampled video frames and aggregate the resulting tokens into a single sequence.
Given a video, we uniformly sample $T$ frames $\{x^I_t\}_{t=1}^{T}$,
where each frame $x^I_t \in \mathbb{R}^{H \times W \times C}$.
Each frame is tokenized independently using the same visual codebook as for images,
producing a discrete token sequence of length $\frac{H}{f} \cdot \frac{W}{f}$.
The per-frame token sequences are then concatenated in temporal order
to form a video token sequence of length
$T \cdot \frac{H}{f} \cdot \frac{W}{f}$.
This design yields a unified discrete representation for both images and videos
within a shared latent space.
Empirically, we observe that improvements in image understanding
consistently translate to enhanced video understanding performance,
suggesting effective knowledge transfer enabled by the shared visual tokenization.

\paragraph{\textbf{Speech.}}
For speech, we adopt the EMOVA speech-to-unit (S2U) tokenizer~\citep{chen2024emova},
which encodes an input waveform into discrete speech units.
Concretely, EMOVA uses a SPIRAL-based speech encoder~\citep{huang2022spiral} to capture phonetic and tonal cues,
followed by a finite scalar quantizer (FSQ) that discretizes the semantic content into a
codebook of size $|\mathcal{V}_{\text{speech}}| = 4096$ at a rate of $25$ tokens per second
(\ie one unit every $\sim$40\,ms).
We integrate speech into our token-based backbone by concatenating
the speech codebook indices to the original vocabulary.
Speech tokenizer further adopts a semantic--acoustic disentanglement design, where only the semantic
embedding is quantized into speech units, while acoustic style is modeled separately.
For detokenization, we use EMOVA’s unit-to-speech (U2S) detokenizer built on VITS~\citep{kim2021vits},
which conditions a speech decoder on (i) unit embeddings derived from generated speech units
and (ii) a style embedding extracted from a reference style prototype.
In particular, the tokenizer constructs $24$ style prototypes by combining two genders
(male/female), four emotions (neutral/happy/sad/angry), and three pitches (normal/high/low),
enabling controllable speech generation across these styles. Similar to the image modality, the speech modality also relies on a pretrained and frozen reconstruction module, with all sequence-level generative modeling performed within the masked diffusion backbone.

\section{Data Composition}
\label{sec:data_composition}
\begin{figure}
    \centering
\includegraphics[width=1\linewidth,page=6]{figs/main_figures.pdf}
    \caption{Distribution of training data across stages and tasks.
    Each stage is represented by a pair of horizontal bars: the upper bar shows the composition of training tasks
    (\eg Text→Text, Image→Text), while the lower bar depicts the proportion of real and synthetic data used in that stage.
    Percentages indicate the relative contribution of each task within a stage.
    }
    \label{fig:data_composition}
\end{figure}

Fig.~\ref{fig:data_composition} illustrates the data allocation across the three training stages, 
and Tab.~\ref{tab:data_composition} summarizes the dataset statistics.
Our data composition follows a progressive scaling strategy: 
Stage~1 prioritizes modality-specific datasets to align newly introduced modalities with the backbone; 
Stage~2 substantially increases cross-modal instruction data to enable balanced joint omnimodal supervision; 
and Stage~3 emphasizes high-quality reasoning traces and curated multimodal samples for capability refinement. 
This staged allocation ensures stable modality integration in early training, 
broad omnimodal coverage in intermediate training, 
and targeted capability enhancement in later stages.

\paragraph{\textbf{Training Stage~1 (Modality Alignment).}}
Stage~1 aims to align newly introduced modalities
with text as the semantic anchor.
For video-to-text alignment, we leverage WebVid-10M~\citep{webvid},
a large-scale video captioning corpus containing $10.7$M video–text pairs
covering approximately $52$K hours of video.
For speech alignment (used for both ASR and TTS),
we combine multiple large-scale speech datasets,
including GigaSpeech ($10$K hours)~\citep{chen2021gigaspeech},
LibriSpeech ($360$ hours)~\citep{panayotov2015librispeech},
and CommonVoice ($3$K hours)~\citep{ardila2020common},
resulting in approximately $14$K hours of speech data.
The same speech datasets are consistently used for both ASR and TTS
to ensure stable semantic grounding across modalities.

\paragraph{\textbf{Training Stage~2 (Omnimodal SFT).}}
Stage~2 performs joint supervised fine-tuning across
text, image, video, and speech to establish unified cross-modal capability.
For text, we train on large-scale human–assistant dialogue data
(\eg Evol-Instruct~\citep{evloinstruct} and Magpie-Pro~\citep{xu2024magpie})
to improve conversational ability and general knowledge,
and additionally include mathematical reasoning datasets
(\eg Open-Platypus~\citep{lee2023platypus},
OpenR1-Math~\citep{openr1},
OpenHermes-2.5~\citep{OpenHermes2_5}),
where supervision is applied only to concise final answers
without explicit chain-of-thought traces.
For image understanding, we use Cambrian10M~\citep{tong2024cambrian}
to cover diverse and complex visual formats.
Image generation and editing are trained using a mixture of
high-quality datasets, including JourneyDB~\citep{sun2023journeydb},
FLUX-Reason-6M~\citep{fang2025fluxreason},
PickaPic~\citep{kirstain2023pickapick},
UltraEdit~\citep{zhao2024ultraedit},
HQEdit~\citep{hui2024hqedit}, and Pico-Banana-400K~\citep{qian2025picobanana400k}.

{For Stage~2 video understanding, we combine LLaVA-Video-178K~\citep{zhang2024llavavideo} and OpenVid1M~\citep{nan2024openvid1m} with approximately 2M synthetic video--text pairs. We construct generation prompts from captions and instructions in the public video datasets, remove near-duplicate and low-information prompts, and generate the corresponding clips using Wan2.2~\citep{wan2025}. The generated clips are filtered for decoding failures, insufficient motion, poor visual quality, and low agreement between the conditioning text and video content using InternVL-based consistency checks~\citep{chen2024internvl}. Near-duplicate clips and samples flagged by the benchmark-overlap checks are also removed. The retained clips are paired with their originating descriptions and converted into video captioning and instruction-following examples. Together with the public datasets, this produces the 3.2M video-to-text examples used in Stage~2.}
For speech, we reuse the ASR and TTS datasets employed in Stage~1,
ensuring continuity between modality alignment and joint training.

\paragraph{\textbf{Training Stage~3 (Capability Scaling).}}
Stage~3 focuses on enhancing advanced reasoning,
long-form generation quality, and multimodal robustness.
For textual reasoning, we introduce explicit chain-of-thought (CoT)
supervision with thinking-mode control tokens.
We leverage publicly available reasoning datasets,
including Llama-Nemotron Post-Training~\citep{bercovich2025llamanemotrone},
OpenR1-Math-220K~\citep{lozhkov2025openr1math220k},
Mixture-of-Thoughts~\citep{mixtureofthought},
and OpenMathReasoning~\citep{openmathreasoning}.

{For textual reasoning, we supplement the public reasoning datasets with traces generated by GLM-4.5-Air~\citep{zeng2025glm}, GPT-OSS-120B~\citep{agarwal2025gptoss}, Qwen3-Next-80B-A3B-Thinking~\citep{yang2025qwen3}, and DeepSeek-R1-Distill-Qwen-32B~\citep{qwendeepseek}. Starting from approximately 10M candidate traces, we verify final answers using task-specific checkers where available, remove traces whose reasoning is inconsistent with the final answer, and discard incomplete, repetitive, or truncated generations. We additionally balance the retained examples across subjects and difficulty levels, resulting in 200K synthetic reasoning examples.}

{For image understanding, we augment Cambrian10M~\citep{tong2024cambrian} with synthetic captioning, question-answering, and multi-turn instruction-following examples. InternVL~\citep{chen2024internvl} generates questions and answers grounded in the source images, after which a separate verification pass checks whether each answer is supported by visible image content. We remove examples with ungrounded answers, ambiguous questions, OCR corruption, or low image--text consistency. The resulting public and synthetic mixture contains 0.4M image-understanding examples.}

{For text-to-image generation, we construct prompts from the public image--text mixture and expand them with variations in composition, attributes, spatial relations, and rendering style. We generate candidate images using FLUX~\citep{flux2024}, FLUX2~\citep{flux-2-2025}, Z-Image~\citep{team2025zimage}, and Qwen-Image~\citep{wu2025qwenimagetechnicalreport}. The candidates are filtered using InternVL~\citep{chen2024internvl} for prompt--image consistency and instruction satisfaction, together with rule-based checks for resolution, malformed outputs, duplicated images, embedded text, and low visual quality. After filtering and balancing prompt categories, we retain 0.5M text-to-image pairs.}

{For image editing, we construct source--instruction--target triplets using FLUX-Kontext~\citep{labs2025flux1kontextflowmatching} and Qwen-Image~\citep{wu2025qwenimagetechnicalreport}. Candidate targets are retained only when InternVL confirms that the requested modification is present while unrelated source content is preserved. We additionally remove pairs with excessive global changes, weak source--target correspondence, or ambiguous editing instructions. The final image-editing mixture contains 0.5M examples.}

{For video understanding, we combine LLaVA-Video-178K~\citep{zhang2024llavavideo} and ShareGPT4Video~\citep{chen2024sharegpt4video} with additional question--answer pairs constructed from the training videos. Qwen2.5-VL~\citep{Qwen2.5-VL} generates questions requiring object, action, temporal-order, and multi-frame reasoning together with answers grounded in the corresponding clips. We discard questions that can be answered without observing the video, answers unsupported by the sampled frames, and examples for which independent teacher predictions disagree. The resulting video-understanding mixture contains 0.8M examples.}

{We construct a candidate pool of 300K long-form speech examples using Kokoro~\citep{kokoro}. The input texts are sampled from public transcripts and the filtered text-instruction mixture and are synthesized with diverse voices and speaking styles, covering utterances up to 21 seconds. We remove clips containing excessive silence, clipping, decoding failures, or durations outside the training range. We further transcribe each generated clip with Qwen3-ASR~\citep{Qwen3-ASR} and retain only samples whose transcription agrees with the conditioning text below a 10\% word-error threshold. After filtering, 0.2M speech examples are used for Stage~3 ASR and TTS training.}

To train native joint generation, we construct 10K text--image-to-text--image--speech examples from public text--image instructional data and synthetic procedural examples. For the public portion, consecutive instructions and their accompanying images form the context and continuation. For the synthetic portion, GLM-4.5-Air~\citep{zeng2025glm} generates procedural contexts, narration, scene descriptions, and spoken utterances. We create or edit candidate scene images using the image-generation and editing models described above---FLUX~\citep{flux2024}, FLUX2~\citep{flux-2-2025}, Z-Image~\citep{team2025zimage}, Qwen-Image~\citep{wu2025qwenimagetechnicalreport}, and FLUX-Kontext~\citep{labs2025flux1kontextflowmatching}---and synthesize speech with Kokoro~\citep{kokoro}. We retain examples that pass contextual-consistency, image--text alignment, ASR-agreement, output-length, and duplicate-content checks. Each retained sequence supervises interleaved text, image, and speech generation.

\paragraph{Data Construction Materials.}
We will release the prompts, generator configurations, filtering code, and manifests used to construct the synthetic datasets above. For source media that cannot be redistributed, we will provide source identifiers, hashes, and reconstruction scripts.

\section{Training Recipes}
\label{sup:training_recipes}
\subsection{Dataset Curation}

We report the datasets used at each stage of \method{}, including their sources and sizes, in Tab.~\ref{tab:data_composition}.

\begin{table}[t]
\centering
\footnotesize
\setlength{\tabcolsep}{16.5pt}
\renewcommand{\arraystretch}{1.2}
\caption{Detailed data composition across training stages for \method{}.
Tasks are described using unified input--output directions
(e.g., V$\rightarrow$T denotes video-to-text).}
\label{tab:data_composition}
\begin{adjustbox}{max width=\linewidth}
\begin{tabular}{c l l c}
\toprule
\textbf{Stage} & \textbf{Task} & \textbf{Dataset Source} & \textbf{Scale (\#~pair)} \\
\midrule
\multirow[c]{3}{*}{Stage 1}
& V$\rightarrow$T &
WebVid-10M~\citep{webvid} &
10.7M  \\
\cmidrule{2-4}
& S$\rightarrow$T &
GigaSpeech~\citep{chen2021gigaspeech}; LibriSpeech~\citep{panayotov2015librispeech}; Common Voice~\citep{ardila2020common} &
\multirow[c]{2}{*}{10M} \\
\cmidrule{2-3}
& T$\rightarrow$S &
GigaSpeech~\citep{chen2021gigaspeech}; LibriSpeech~\citep{panayotov2015librispeech}; Common Voice~\citep{ardila2020common} &
\\
\midrule
\multirow[c]{7}{*}{Stage 2}
& T$\rightarrow$T &
\makecell[l]{Evol-Instruct~\citep{evloinstruct}; Magpie-Pro~\citep{xu2024magpie};
Open-Platypus~\citep{lee2023platypus}; 
\\ OpenR1-Math~\citep{lozhkov2025openr1math220k}; OpenHermes-2.5~\citep{OpenHermes2_5}} &
1.6M \\
\cmidrule{2-4}
& I$\rightarrow$T &
Mantis-Instruct~\citep{Jiang2024MANTISIM}; Cambrian10M~\citep{tong2024cambrian} &
8M \\
\cmidrule{2-4}
& T$\rightarrow$I &
\makecell[l]{JourneyDB~\citep{sun2023journeydb}; FLUX-Reason-6M~\citep{fang2025fluxreason}; PickaPic~\citep{kirstain2023pickapick}} &
10M \\
\cmidrule{2-4}
& I$\rightarrow$I &
\makecell[l]{UltraEdit~\citep{zhao2024ultraedit}; HQEdit~\citep{hui2024hqedit}; Pico-Banana-400K~\citep{qian2025picobanana400k}} &
1.1M \\
\cmidrule{2-4}
& V$\rightarrow$T &
\makecell[l]{LLaVA-Video-178K~\citep{zhang2024llavavideo}; OpenVid1M~\citep{nan2024openvid1m}; Synthetic Inst. Data} &
3.2M  \\
\cmidrule{2-4}
& S$\rightarrow$T &
GigaSpeech~\citep{chen2021gigaspeech}; LibriSpeech~\citep{panayotov2015librispeech}; Common Voice~\citep{ardila2020common} &
\multirow[c]{2}{*}{10M} \\
\cmidrule{2-3}
& T$\rightarrow$S &
GigaSpeech~\citep{chen2021gigaspeech}; LibriSpeech~\citep{panayotov2015librispeech}; Common Voice~\citep{ardila2020common} &
\\
\midrule
\multirow[c]{7}{*}{Stage 3}
& T$\rightarrow$T &
\makecell[l]{Llama-Nemotron Post-Training~\citep{bercovich2025llamanemotrone}; OpenR1-Math-220K~\citep{lozhkov2025openr1math220k}; \\
Mixture-of-Thoughts~\citep{mixtureofthought}; OpenMathReasoning~\citep{openmathreasoning}; Synthetic Data} &
0.5M \\
\cmidrule{2-4}
& I$\rightarrow$T &
\makecell[l]{Cambrian10M~\citep{tong2024cambrian}; Synthetic Data} &
0.4M \\
\cmidrule{2-4}
& T$\rightarrow$I &
FLUX-Reason-6M~\citep{fang2025fluxreason}; Synthetic Data &
0.5M \\
\cmidrule{2-4}
& I$\rightarrow$I &
Pico-Banana-400K~\citep{qian2025picobanana400k}; Synthetic Data &
0.5M \\
\cmidrule{2-4}
& V$\rightarrow$T &
LLaVA-Video-178K~\citep{zhang2024llavavideo}; ShareGPT4Video~\citep{chen2024sharegpt4video}; Synthetic QA Data &
0.8M \\
\cmidrule{2-4}
& S$\rightarrow$T &
LibriSpeech~\citep{panayotov2015librispeech}; Synthetic Data&
\multirow[c]{2}{*}{0.2M} \\
\cmidrule{2-3}
& T$\rightarrow$S &LibriSpeech~\citep{panayotov2015librispeech}; Synthetic Data & \\
\cmidrule{2-4}
& TI$\rightarrow$TIS & Public + Synthetic Omnimodal Data & 0.01M
\\
\bottomrule
\end{tabular}
\end{adjustbox}
\end{table}

\begin{figure*}[t]
    \centering
    \color{black}
    \includegraphics[width=0.94\linewidth,page=5]{figs/main_figures.pdf}
    \caption{Three-stage training recipe for stable omnimodal extension of \method{}.}
    \label{fig:training_stages_main}
\end{figure*}

\subsection{Training Details}
\label{app:training_details}

Tab.~\ref{tab:training_hyperparams} summarizes the stage-wise training configurations of \method{}, including optimization hyperparameters, compute budget, modality-specific resolutions, and speech duration.
All stages use AdamW with cosine learning-rate scheduling, weight decay $0.1$, gradient norm clipping $1.0$, and batch size $128$.
Stage~1 uses a learning rate of $1.0\times10^{-5}$ on $32$ H100 GPUs, while Stages~2 and~3 use $2.0\times10^{-5}$ on $32$ and $40$ H100 GPUs, respectively.
The complete adaptation cost after initialization from the released MMaDA checkpoint is 63,425 H100-hours: 5,200 for Stage~1, 34,560 for Stage~2, and 23,665 for Stage~3. This accounting covers all three adaptation stages and excludes the inherited MMaDA pretraining cost.

\paragraph{Vocabulary extension.}
Starting from the MMaDA backbone, \method{} extends the original text--vision vocabulary with a dedicated speech vocabulary.
The unified vocabulary is organized as
\[
\mathcal{V}
=
\mathcal{V}_{\text{text}}
\cup
\mathcal{V}_{\text{vision}}
\cup
\mathcal{V}_{\text{speech}},
\]
where $|\mathcal{V}_{\text{text}}|=126080$, $|\mathcal{V}_{\text{vision}}|=8192$, and $|\mathcal{V}_{\text{speech}}|=4096$.
Images and videos share the same visual vocabulary, with videos represented by tokenizing sampled frames using the image tokenizer.
Speech tokens are added as newly introduced vocabulary entries and processed by the same shared embedding table and masked-diffusion Transformer.

\paragraph{Stage~1: new-modality adaptation.}
Stage~1 adapts newly introduced modalities through text-centric alignment tasks, including video captioning, automatic speech recognition (ASR), and text-to-speech (TTS).
As shown in Tab.~\ref{tab:training_hyperparams}, we use $8$ video frames at $128\times128$ resolution and speech segments up to approximately $10$ seconds.
This low-resolution, higher-frame setting emphasizes temporal alignment during early adaptation.
During this stage, \texttt{[EOS]} tokens are excluded from supervision, allowing the model to prioritize semantic alignment before learning termination behavior.

\paragraph{Stage~2: omnimodal supervised fine-tuning.}
Stage~2 jointly trains text, image, video, and speech tasks under the unified objective.
Image understanding and image editing use $256\times256$ resolution, while image generation uses $336\times336$ resolution.
For video, we use $5$ frames at $224\times224$ resolution.
For speech, utterances are limited to approximately $15$ seconds.
From Stage~2 onward, \texttt{[EOS]} supervision is enabled so that the model learns flexible-length termination after semantic alignment has been established.

\paragraph{Stage~3: capability scaling.}
Stage~3 further improves reasoning, high-resolution visual generation, image editing, video understanding, and long-form speech modeling.
We increase image understanding resolution to $480\times480$, image generation resolution to $512\times512$, and image editing resolution to $336\times336$.
Video inputs remain at $5$ frames with $224\times224$ resolution.
For speech, Stage~3 extends the maximum duration to approximately $21$ seconds.

{For the joint omnimodal examples described in App.~\ref{sec:data_composition}, images are generated at $336\times336$ resolution and speech outputs are capped at $5$ seconds.}

\begin{table}[t]
\centering
\tiny
\setlength{\tabcolsep}{12pt}
\caption{Detailed hyperparameters and configurations of the training recipe across different stages.}
\label{tab:training_hyperparams}
\begin{tabular}{lccc}
\toprule
\textbf{Hyperparameters} &
\textbf{Stage~1} &
\textbf{Stage~2} &
\textbf{Stage~3} \\
&
(New Modality Adapt.) &
(Omni. SFT) &
(Cont. SFT) \\
\midrule
Learning Rate        & $1.0 \times 10^{-5}$ & $2.0 \times 10^{-5}$ & $2.0 \times 10^{-5}$ \\
LR Scheduler         & Cosine & Cosine & Cosine \\
Weight Decay         & 0.1 & 0.1 & 0.1 \\
Gradient Norm Clip   & 1.0 & 1.0 & 1.0 \\
Optimizer & \multicolumn{3}{c@{}}{\hspace{6pt}AdamW ($\beta_1=0.9$, $\beta_2=0.999$)} \\
Batch Size           & 128 & 128 & 128 \\
Training GPUs      & $32\times$H100 & $32\times$H100 & $40\times$H100 \\
GPU Hours            & 5,200 & 34,560 & 23,665 \\
\midrule
Image Und. Resolution     & -- & 256 & 480 \\
Image Gen. Resolution     & -- & 336 & 512 \\
Image Edit Resolution     & -- & 256 & 336 \\
Video Resolution / $\#$ Frame & 128 / 8 & 224 / 5 & 224 / 5 \\
Speech Duration~(sec)        & $\sim$10 & $\sim$15 & $\sim$21 \\
\bottomrule
\end{tabular}
\end{table}

\begin{figure}[!t]
    \centering
    \includegraphics[width=0.75\linewidth]{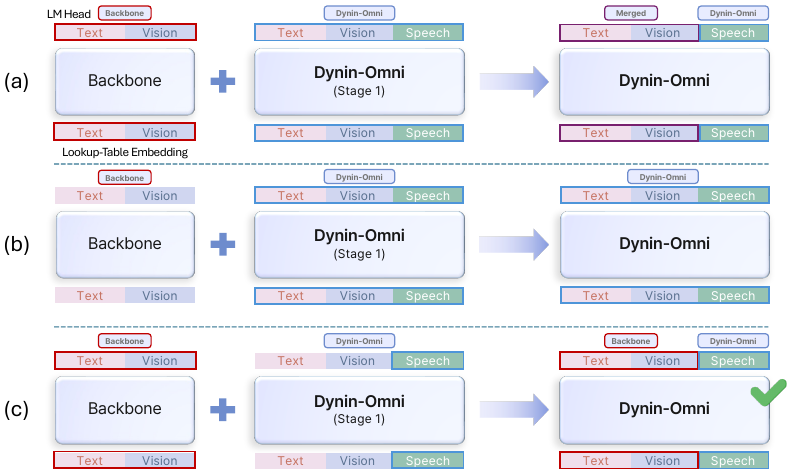}
    \caption{
Comparison of model merging strategies under vocabulary extension.
(a) Shared Merging: Original vocabulary parameters are interpolated between the backbone and Stage~1, while new speech embeddings come from Stage~1.  
(b) Stage~1-Only Merging: All vocabulary-related parameters are taken from Stage~1.  
(c) Modality-Disentangled Merging (Ours): Original vocabulary parameters remain from the backbone, while new speech-specific dimensions come from Stage~1. Remaining backbone weights are interpolated.
}
    \label{fig:merging_choices}
\end{figure}

\subsection{Formulation of Merging Strategies}
\label{app:merging_strategies}

Fig.~\ref{fig:merging_choices} illustrates the three vocabulary-merging strategies considered under vocabulary expansion.
We provide formal definitions of these strategies, which are compared in \S\ref{subsec:training_recipe_analysis}.
Let $\theta^{(0)}$ denote the original backbone and $\theta^{(1)}$ denote the Stage~1-adapted model.
Stage~1 extends the vocabulary from $\mathcal{V}^{(0)}$ to
$\mathcal{V}=\mathcal{V}^{(0)}\cup\mathcal{V}^{\mathrm{new}}$,
where $\mathcal{V}^{\mathrm{new}}$ contains newly introduced modality tokens.
We decompose the model parameters into
$\theta=\{\theta_{\mathrm{shared}},\theta_{\mathrm{vocab}}\}$,
where $\theta_{\mathrm{shared}}$ denotes vocabulary-independent parameters and
$\theta_{\mathrm{vocab}}$ denotes vocabulary-dependent parameters, such as token embeddings and output heads.

For parameters with matching dimensionality, we define the interpolation operator as
\begin{equation}
\mathcal{M}(\theta^{(0)},\theta^{(1)};\alpha)
=
(1-\alpha)\theta^{(0)}+\alpha\theta^{(1)}.
\end{equation}
Here, $\alpha$ denotes the interpolation weight assigned to the Stage~1-adapted model.
All strategies apply this interpolation to vocabulary-independent parameters:
\begin{equation}
\theta_{\mathrm{shared}}^{\mathrm{merge}}
=
\mathcal{M}(\theta_{\mathrm{shared}}^{(0)},\theta_{\mathrm{shared}}^{(1)};\alpha)
=
(1-\alpha)\theta_{\mathrm{shared}}^{(0)}
+\alpha\theta_{\mathrm{shared}}^{(1)}.
\end{equation}
The strategies differ only in how they handle vocabulary-dependent parameters.

\paragraph{Shared Merging.}
For parameters corresponding to the original vocabulary, Shared Merging interpolates the original and Stage~1 parameters.
Newly introduced token parameters are inherited from the Stage~1 model:
\begin{equation}
\theta_{\mathrm{vocab}}^{\mathrm{merge}}(i)
=
\begin{cases}
\mathcal{M}\!\left(
\theta_{\mathrm{vocab}}^{(0)}(i),
\theta_{\mathrm{vocab}}^{(1)}(i);
\alpha
\right), & i\in\mathcal{V}^{(0)},\\
\theta_{\mathrm{vocab}}^{(1)}(i), & i\in\mathcal{V}^{\mathrm{new}}.
\end{cases}
\end{equation}
This strategy follows standard parameter interpolation on the overlapping vocabulary, but may disrupt token-level representations learned by either model.

\paragraph{Stage~1-Only Merging.}
Stage~1-Only Merging inherits all vocabulary-dependent parameters from the Stage~1-adapted model:
\begin{equation}
\theta_{\mathrm{vocab}}^{\mathrm{merge}}(i)
=
\theta_{\mathrm{vocab}}^{(1)}(i),
\quad
i\in\mathcal{V}.
\end{equation}
This maximally preserves newly acquired modality-specific representations, but may retain catastrophic forgetting of original text--image token representations.

\paragraph{Modality-Disentangled Merging.}
Our modality-disentangled strategy preserves original vocabulary parameters from the backbone and inherits only newly introduced token parameters from the Stage~1 model:
\begin{equation}
\theta_{\mathrm{vocab}}^{\mathrm{merge}}(i)
=
\begin{cases}
\theta_{\mathrm{vocab}}^{(0)}(i), & i\in\mathcal{V}^{(0)},\\
\theta_{\mathrm{vocab}}^{(1)}(i), & i\in\mathcal{V}^{\mathrm{new}}.
\end{cases}
\end{equation}
This separates backbone preservation from new-modality incorporation, allowing the merged model to retain existing text--image representations while adding newly learned modality tokens.

\paragraph{Discussion.}
The key difficulty in omnimodal modality expansion is that vocabulary extension introduces new rows in the embedding table and output head.
As a result, naive parameter interpolation is ill-defined for the full vocabulary-dependent parameters.
The above strategies instantiate different choices for resolving this mismatch.
Shared merging interpolates overlapping rows, stage~1-only favors newly adapted representations, and modality-disentangled merging explicitly preserves original-token representations while importing new-token representations.
As shown in \S\ref{subsec:training_recipe_analysis}, this disentangled treatment provides the most stable initialization for Stage~2 omnimodal supervised fine-tuning.

\section{Inference Details}
\label{app:inference_configs}

\subsection{Inference Configurations}
\label{app:inference_config}

We use block-wise decoding with block size 16 and 1024 steps for text reasoning, and block size 16 with 128 steps for image/video understanding and ASR.
For TTS, we use block-wise decoding with block size 128 and 512 steps.
For image synthesis, we use fully parallel decoding with 16 steps and classifier-free guidance (CFG)~\citep{ho2022classifier} scale 3.5 for generation, and 128 steps with CFG scale 2.5 for editing.
For all tasks, we use confidence-based remasking~\citep{Chang_2022_CVPR,nie2025large,dream2025} during iterative decoding.

\subsection{Efficiency Experiment Details}
\label{app:inference_efficiency_details}

\paragraph{Accelerated inference setup.}
We evaluate text-to-text and text-to-speech inference speed on a single NVIDIA A100 80GB SXM GPU.
Following Fast-dLLM~\citep{wu2026fastdllm}, we apply KV-cache approximation and confidence-based parallel decoding.
The optimized inference path is implemented in a customized version of dInfer~\citep{ma2025dinferefficientinferenceframework}.
For both GSM8K and LibriSpeech-clean TTS, we use DualCache with block size 64 and threshold 0.7.
For the matched MMaDA comparison, we port the same dInfer/DualCache path to MMaDA and hold batch size, maximum generation length, confidence threshold, block size, \texttt{torch.compile}, and CUDA Graph fixed.

\paragraph{Baseline configurations.}
For text-to-text evaluation, Qwen3-8B is evaluated in a 4-shot setting with exact autoregressive decoding using vLLM.
MMaDA is evaluated in a 5-shot setting using its official repository, with generation length 256, fixed-step parallel decoding with 128 steps, and block length 64; KV caching is not supported in this setting.
The standard released-checkpoint evaluation, the paper-reported final UniGRPO result, and the accelerated dInfer throughput run use distinct checkpoints or inference configurations.
For \method{}, the accelerated run scores 84.23 on GSM8K, while the standard benchmark result is 83.7; the corresponding accelerated scores and throughput are reported in Tab.~\ref{tab:matched_inference} and Fig.~\ref{fig:speedup_report}.
Qwen2.5-Omni-7B is evaluated in a zero-shot setting with exact autoregressive decoding using vLLM.
For text-to-speech evaluation, Qwen3-TTS-1.7B is evaluated using vLLM-Omni, while Qwen2.5-Omni-7B is evaluated using vLLM.

\subsection{Statistical Details}
\label{app:statistical}

For the large-scale benchmark tables, we evaluate a single trained checkpoint using the official
evaluation splits and protocols.
{We  report item-bootstrap intervals for separate supplementary GSM8K and  MATH panels, and four-decoding-seed intervals for GenEval, ImgEdit, TTS WER, and TTS UTMOS; diagnostic plots show benchmark-level dispersion.}

\paragraph{Merging-strategy analysis.}
For Fig.~\ref{fig:merging_strategies}, each point corresponds to one merged checkpoint under a
specific merging strategy and interpolation ratio $\alpha$. We compute original-capability retention
over GSM8K, POPE, and MME-P, and new-modality retention over LibriSpeech\textsubscript{clean} ASR and
TTS WER. For higher-is-better metrics, retention is computed by normalizing the score of the merged
checkpoint by the corresponding reference checkpoint. For lower-is-better WER metrics, retention is
computed using the inverse normalization so that larger values consistently indicate better retention.
Each normalized retention score is clipped to $[0,1]$ before averaging.
The plotted coordinate is the mean retention over the corresponding benchmark group. Error bars
denote standard deviation across the normalized benchmark-level retention scores within
each group.

\paragraph{Omnimodal-generation analysis.}
For Fig.~\ref{fig:omnimodal_generation}, each method and NFE budget is evaluated on the same Omni-StoryBench~\citep{omnistorybench} examples. The main-table Overall (Tab.~\ref{tab:omnimodal_generation}) equally averages text, image, and speech scores, each the mean of its two displayed metrics after scaling 0--1 metrics to 0--10. For this NFE analysis, Overall averages five metrics: BERTScore, text judge, CLIP similarity, image judge, and speech judge, with BERTScore and CLIP similarity similarly scaled. We omit speech metadata accuracy because it varies little across the tested budgets; the Omni judge is excluded from both averages and reported separately in the appendix tables. To visualize how evenly each method performs across evaluation categories, we compute the standard deviation across category-level scores at each method and NFE budget. Shaded regions denote $\pm 0.2$ times this cross-category standard deviation around the NFE-specific Overall score.

\paragraph{{Benchmark human validation.}}
{Omni-StoryBench~\citep{omnistorybench}, Appendix~F.5, reports human validation with 24 annotators, 32 systems, and 687 scored evaluations. Its integrated system-ranking Spearman correlation is 0.884. This is validation reported by the benchmark; the speech-content consistency checks in our added evaluation compare intended utterances with independently transcribed generated audio and retain missing outputs.}

\section{Serving Infrastructure}
\label{app:deployment}
For serving, we use a vLLM-Omni~\citep{yin2026vllmomni}-compatible implementation for text-to-image generation, text-to-speech synthesis, and image editing. It follows the framework's stage-based execution interface and supports block-wise parallel decoding with iterative remasking for diffusion-based outputs.
To further optimize inference efficiency, we incorporate \method{} with a unified inference framework supporting both autoregressive baselines and masked-diffusion generation. The masked-diffusion inference path is implemented on top of a customized version of dInfer~\citep{ma2025dinferefficientinferenceframework}, extended for modality-aware iterative decoding with block-wise refinement for sequential modalities.
To reduce overhead in iterative decoding, we retain dInfer's compiled execution path using \texttt{torch.compile}, CUDA Graph replay, and loop unrolling across diffusion refinement steps.
For efficient long-context inference, we additionally integrate KV-cache approximation following Fast-dLLM~\citep{wu2026fastdllm}, reusing KV states for unchanged context and committed regions across refinement steps.

\section{Additional Results}
\label{app:additional_results}

\subsection{Training Recipe Controls}
\label{app:matched_controls}
We provide the full controls underlying the summarized results in Tab.~\ref{tab:training_recipe_effects}.
For merging, $K{=}0$ denotes the checkpoint immediately after initialization and before Stage~2; $K{=}10$K denotes exactly 10,000 Stage~2 updates with identical data and compute.
Original-capability retention averages clipped GSM8K, POPE, and MME-P ratios to the released backbone; new-modality retention averages clipped inverse-WER ratios to the Stage~1 checkpoint; and H-mean is their harmonic mean (Tab.~\ref{tab:merging_controls}).
For the \texttt{[EOS]} controls, disabling supervision removes \texttt{[EOS]} from the training targets while still permitting its generation at inference.
The termination metrics are computed over the pooled ASR and TTS samples. Non-premature termination means that the first \texttt{[EOS]} occurs at or after 80\% of the reference length and before the decoding cap; premature termination occurs earlier, and a cap hit means that no \texttt{[EOS]} appears before the maximum decoding length.
Repetition denotes a four-token sequence repeated consecutively and can overlap a cap hit (Tab.~\ref{tab:eos_controls}).
\begin{table}[H]
\centering
\tiny
\color{black}
\caption{Full merging controls at initialization and after 10K Stage~2 updates. Ratios denote backbone-to-Stage~1 interpolation weights.}
\label{tab:merging_controls}
\setlength{\tabcolsep}{2.5pt}
\renewcommand{\arraystretch}{1.0}
\begin{adjustbox}{max width=\linewidth}
\begin{tabular*}{\linewidth}{@{\extracolsep{\fill}}clrrrrr@{}}
\toprule
$K$ & Initialization & ASR/TTS WER$\downarrow$ & GSM8K$\uparrow$ & POPE$\uparrow$ & MME-P$\uparrow$ & H-mean$\uparrow$ \\
\midrule
0 & Stage~1 / no merge & 12.50 / 31.50 & 4.2 & 30.7 & 413.8 & 38.9 \\
0 & Shared, 6:4 & 120.00 / 150.00 & 35.0 & 70.0 & 1000.0 & 25.8 \\
0 & Stage~1-only, 6:4 & 90.00 / 120.00 & 28.0 & 68.0 & 950.0 & 30.7 \\
0 & Modality-disent., 6:4 & 69.70 / 81.40 & 23.4 & 72.1 & 946.3 & \textbf{39.3} \\
\midrule
10K & No merge & 10.50 / 20.00 & 18.0 & 50.0 & 820.0 & 66.1 \\
10K & Modality-disent., 4:6 & 9.00 / 17.00 & 27.0 & 62.0 & 950.0 & 76.9 \\
10K & Modality-disent., 6:4 & \textbf{6.46 / 10.22} & \textbf{38.0} & \textbf{76.0} & \textbf{1150.0} & \textbf{88.3} \\
\bottomrule
\end{tabular*}
\end{adjustbox}
\end{table}

\begin{table}[H]
\centering
\tiny
\color{black}
\caption{Detailed controls for scheduled \texttt{[EOS]} supervision.}
\label{tab:eos_controls}
\setlength{\tabcolsep}{1.5pt}
\renewcommand{\arraystretch}{1.0}
\begin{adjustbox}{max width=\linewidth}
\begin{tabular}{@{}llccrrrr@{}}
\toprule
Stage
& \texttt{[EOS]} supervision
& ASR WER (S/D/I)$\downarrow$
& TTS WER (S/D/I)$\downarrow$
& Non-prem.$\uparrow$
& Prem.$\downarrow$
& Repeat$\downarrow$
& Cap$\downarrow$ \\
\midrule
1 & Enabled
& 131.80 (9.20/88.60/34.00)
& 55.00 (5.00/42.00/8.00)
& \textbf{35\%} & 60\% & \textbf{8\%} & \textbf{5\%} \\
1 & Disabled
& \textbf{100.65 (20.00/3.00/77.65)}
& \textbf{35.07 (5.00/2.00/28.07)}
& 20\% & \textbf{3\%} & 60\% & 77\% \\
2 & Disabled
& 14.00 (3.00/1.50/9.50)
& 24.00 (4.00/2.00/18.00)
& 30\% & \textbf{2\%} & 55\% & 68\% \\
2 & Enabled
& \textbf{6.46 (2.50/1.50/2.46)}
& \textbf{10.22 (3.00/1.22/6.00)}
& \textbf{94\%} & 3\% & \textbf{4\%} & \textbf{3\%} \\
\bottomrule
\end{tabular}
\end{adjustbox}
\end{table}

\subsection{Text-Diffusion and Diffusion-MLLM Comparisons}
\label{app:baseline_reproduction}
We evaluate LLaDA-8B-Base and LLaDA-8B-Instruct from their released checkpoints; Trida-7B and Dream-7B use their reported results (Tab.~\ref{tab:llada_official}).
\begin{table}[H]
\centering
\tiny
\color{black}
\caption{Comparison with text-diffusion baselines.}
\label{tab:llada_official}
\setlength{\tabcolsep}{2.0pt}
\renewcommand{\arraystretch}{1.0}
\begin{adjustbox}{max width=\linewidth}
\begin{tabular}{@{}lcccc@{}}
\toprule
Model & MMLU$\uparrow$ & GSM8K$\uparrow$ & MATH-500$\uparrow$ & GPQA$\uparrow$ \\
\midrule
LLaDA-Base$\dagger$ & 65.9$\pm$0.8 (5-shot) & 71.3$\pm$2.8 (5-shot) & 30.0$\pm$5.1 (4-shot) & 31.5$\pm$4.9 (5-shot) \\
LLaDA-Instruct$\dagger$ & 65.4$\pm$1.5 (5-shot) & 68.7$\pm$2.4 (4-shot) & 29.8$\pm$4.2 (0-shot) & 31.8$\pm$6.0 (5-shot) \\
Trida-7B~\citep{trida} & 67.2 (5-shot) & 65.1 (0-shot CoT) & 33.6 (0-shot CoT) & -- \\
Dream-7B-Base~\citep{dream2025} & 69.5 (5-shot) & 77.2 (8-shot) & 39.6 (4-shot) & 36.6 (5-shot) \\
Dream-7B-Instruct~\citep{dream2025} & 67.0 (4-shot) & 81.0 (0-shot) & 39.2 (0-shot) & 33.0 (5-shot) \\
\method{} & \textbf{75.9} (0-shot) & \textbf{83.7} (0-shot) & \textbf{47.4} (0-shot) & \textbf{37.9} (0-shot) \\
\bottomrule
\end{tabular}
\end{adjustbox}
\end{table}

We also evaluate LaViDa-LLaDA~\citep{li2025lavida},
LLaDA-o~\citep{you2026lladao},
SDAR-VL-8B-Instruct~\citep{cheng2025sdarvl},
and Dream-VL~\citep{ye2025dreamvl}
on the same POPE, MME-P, and MMMU-val splits (Tab.~\ref{tab:recent_diffusion_mllm}).
\begin{table}[t]
\centering
\tiny
\color{black}
\caption{Comparison with recent diffusion MLLMs.
Baseline entries report the published result followed by our evaluation in parentheses.
N/R indicates that the corresponding result was not reported.
}
\label{tab:recent_diffusion_mllm}
\setlength{\tabcolsep}{1.5pt}
\renewcommand{\arraystretch}{1.0}
\begin{tabular*}{\linewidth}{@{\extracolsep{\fill}}lccccc@{}}
\toprule
Benchmark & LaViDa-LLaDA & LLaDA-o & SDAR-VL-8B-Inst. & Dream-VL & \method{} \\
\midrule
POPE (\%)     & N/R (83.4)    & N/R (86.2)    & N/R (84.3)    & N/R (84.0)    & \textbf{88.0} \\
MME-P         & 1365.6 (1349.4) & 1412.0 (1454.0) & 1632.0 (1465.0) & N/R (1546.9) & \textbf{1672.9} \\
MMMU-val (\%) & 43.3 (44.4)   & 44.9 (44.3)   & 53.0 (47.9)   & 52.2 (51.7)   & 52.8 \\
\bottomrule
\end{tabular*}
\end{table}

\subsection{Statistical Uncertainty and Perceptual TTS Quality}
\label{app:uncertainty}
We report decoding-seed intervals for generation and speech metrics and bootstrap intervals for reasoning benchmarks; UTMOS complements WER with perceptual speech quality (Tab.~\ref{tab:uncertainty}).

The main tables report the fixed-run point estimates used for model comparison.
To quantify evaluation uncertainty, we repeat image generation, image editing, and speech decoding with four random seeds and report the mean and 95\% confidence interval across seeds.
{For  the supplementary reasoning panels, we  use 20,000 item-bootstrap resamples of 256 GSM8K and 128 MATH examples. These panels are separate from the  full-benchmark results in the main tables.}

The generation and speech means can differ from the main-table point estimates because they aggregate repeated decoding runs.

\begin{table}[t]
\centering
\tiny
\color{black}
\caption{
Evaluation uncertainty.
Generation and speech metrics are aggregated over four decoding seeds. {Reasoning rows report the separate fixed-subset evaluations with 20,000 item-bootstrap resamples; they are not confidence intervals for the full-benchmark main-table scores.}
}
\label{tab:uncertainty}
\setlength{\tabcolsep}{3.0pt}
\renewcommand{\arraystretch}{1.0}
\begin{tabular*}{\linewidth}{@{\extracolsep{\fill}}lcc@{}}
\toprule
Metric & Mean & 95\% CI \\
\midrule
GenEval & 0.894 & [0.871, 0.916] \\
ImgEdit & 3.745 & [3.710, 3.780] \\
{GSM8K (207/256)} & { 80.86\%} & { [75.78, 85.55]\%} \\
{ MATH (45/128)} & { 35.16\%} & { [27.34, 43.75]\%} \\
TTS WER & 2.69\% & [1.94, 3.44]\% \\
TTS UTMOS & 4.049 & [3.984, 4.114] \\
\bottomrule
\end{tabular*}
\end{table}

\subsection{Omni-StoryBench Decontamination}
\label{app:omnistory_audit}

We flag a test example if it is identified by any of four overlap checks: an exact match to one of three sampled 50-character spans (GPT-4-style~\cite{gpt4}), at least 20\% token coverage at $L=40$ or $L=50$ (LLaMA-style~\cite{grattafiori2024llama3}), or visual similarity between its target image and a training image. For the visual checks, we extract SSCD (Self-Supervised Copy Detection) ResNet-50 descriptors and compute cosine similarity; an example is flagged if its maximum similarity to any training image is at least 0.95, or at least 0.75 under the DataComp-style setting~\cite{gadre2023datacomp}. The union flags 188 of the 900 examples as potential overlaps. After excluding them, we re-evaluate all nine systems on the same 712-item subset. The ranking is unchanged, and \method{} retains the highest Overall score, with 7.55 on the clean subset versus 7.63 on the full set (Tab.~\ref{tab:omnistory_audit}).
\begin{table*}[t]
\centering
\footnotesize
\caption{
Omni-StoryBench results on the full set and the 712-item clean subset.
All systems are evaluated on the same clean examples.
Overall averages the six displayed metrics after scaling BERTScore,
CLIP similarity, and metadata accuracy to $[0,10]$.
$\Delta$ is Clean minus Full, computed from unrounded scores.
}
\label{tab:omnistory_audit}
\setlength{\tabcolsep}{2pt}
\renewcommand{\arraystretch}{1.05}
\begin{adjustbox}{max width=\textwidth}
\begin{tabular}{lccrrrrrrrrr}
\toprule
&
\multicolumn{2}{c}{Generator} &
\multicolumn{2}{c}{Text$\uparrow$} &
\multicolumn{2}{c}{Image$\uparrow$} &
\multicolumn{2}{c}{Speech$\uparrow$} &
\multicolumn{3}{c}{Overall$\uparrow$} \\
\cmidrule(lr){2-3}
\cmidrule(lr){4-5}
\cmidrule(lr){6-7}
\cmidrule(lr){8-9}
\cmidrule(lr){10-12}
Model & Image & Speech
& BERTScore & LLM Judge
& CLIP Sim. & LLM Judge
& Metadata Acc. & LLM Judge
& Full & Clean & $\Delta$ \\
\midrule
InternVL3.5-4B & FLUX.1 Kontext & Parler
& 0.87 & 7.46 & 0.75 & 5.46 & 0.55 & 8.06
& 7.12 & 7.12 & +0.01 \\
Qwen3.5-4B & FLUX.1 Kontext & Parler
& 0.88 & 8.19 & 0.77 & 6.42 & 0.52 & 8.19
& 7.44 & 7.42 & -0.01 \\
\midrule
EMOVA-7B & FLUX.1 Kontext & Native
& 0.87 & 7.79 & 0.78 & 4.73 & 0.53 & 8.27
& 7.06 & 7.08 & +0.02 \\
Qwen2.5-Omni-7B & FLUX.1 Kontext & Native
& 0.88 & 7.96 & 0.76 & 5.55 & 0.54 & 8.14
& 7.24 & 7.24 & +0.00 \\
\midrule
Emu3 & Native & Parler
& 0.87 & 5.52 & 0.58 & 1.79 & 0.51 & 7.94
& 5.81 & 5.82 & +0.02 \\
MMaDA & Native & Parler
& 0.87 & 4.78 & 0.65 & 1.99 & 0.50 & 8.10
& 5.82 & 5.84 & +0.02 \\
\midrule
AnyGPT & Native & Native
& 0.80 & 3.28 & 0.64 & 1.68 & 0.43 & 5.52
& 4.81 & 4.85 & +0.04 \\
HyperCLOVAX & Native & Native
& 0.87 & 6.43 & 0.68 & 2.53 & 0.54 & 8.20
& 6.32 & 6.33 & +0.01 \\
\rowcolor{blue!8}
\textbf{\method{} (Ours)} & Native & Native
& 0.91 & 6.57 & 0.85 & 6.26 & 0.61 & 8.61
& \textbf{7.63} & \textbf{7.55} & -0.08 \\
\bottomrule
\end{tabular}
\end{adjustbox}
\end{table*}

\subsection{Inference Efficiency and Sampling Budget}
\label{app:inference_efficiency_results}
\begin{figure}[t]
    \centering
    \begin{subfigure}[t]{0.6\linewidth}
        \centering
        \includegraphics[width=0.95\linewidth]{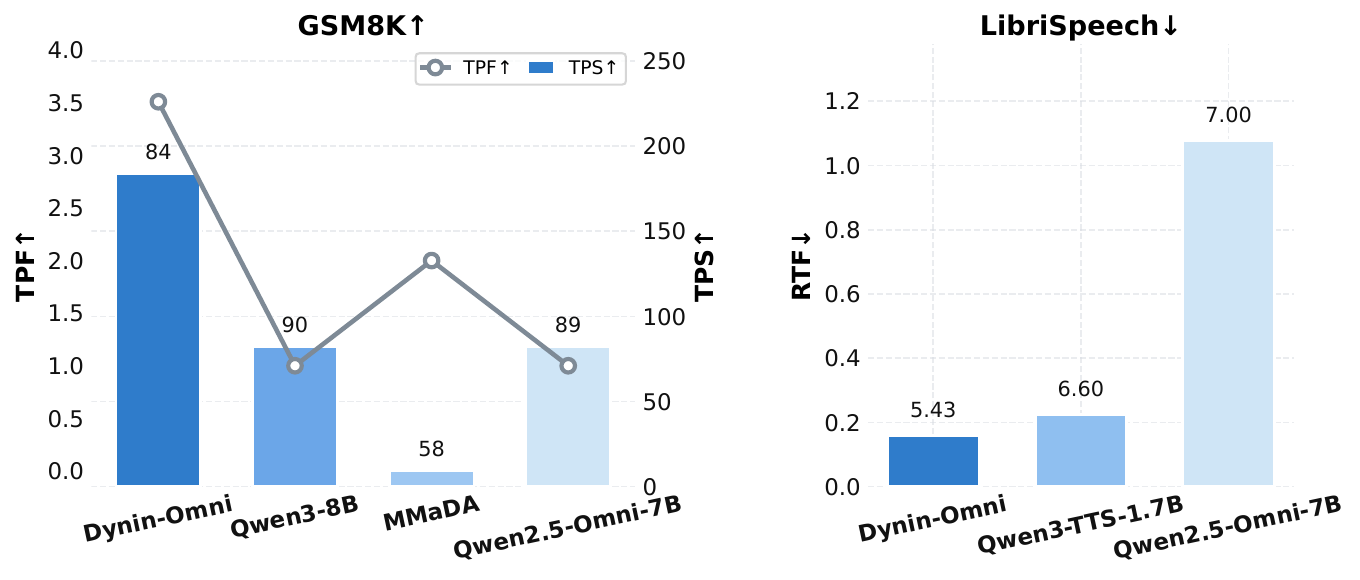}
        \caption{Accelerated inference on reasoning and TTS.}
        \label{fig:speedup_report}
    \end{subfigure}
    \hfill
    \begin{subfigure}[t]{0.38\linewidth}
        \centering
        \includegraphics[width=0.98\linewidth]{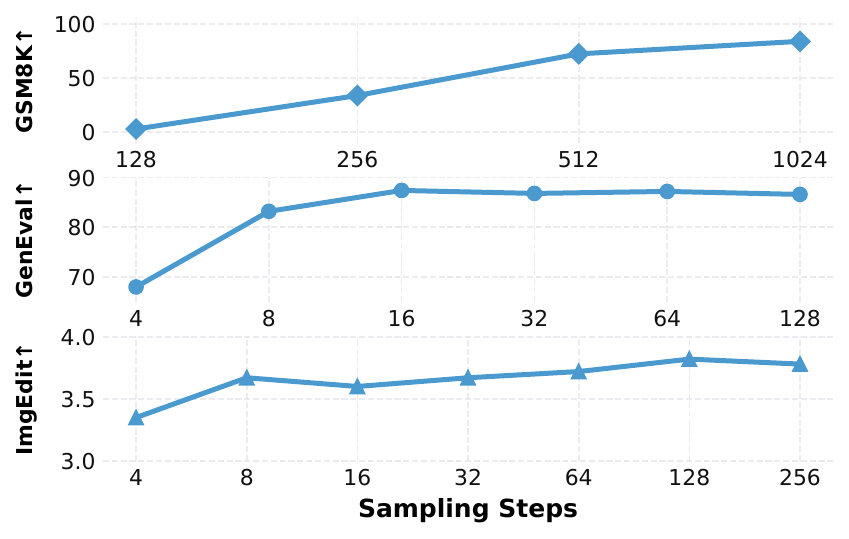}
        \caption{Effect of diffusion sampling steps.}
        \label{fig:sampling_steps}
    \end{subfigure}
    \caption{
    Inference efficiency analysis.
    (a) For reasoning on GSM8K, bars show tokens-per-second (TPS), the line shows tokens-per-forward (TPF), and labels show accuracy. For TTS on LibriSpeech\textsubscript{clean}, bars show real-time-factor (RTF) and labels show WER.
     (b) Reasoning scales with larger refinement budgets, while image generation and editing saturate with moderate sampling steps.
    }
    \vspace{-1em}
    \label{fig:inference_efficiency}
\end{figure}

With the accelerated dInfer/DualCache path, \method{} reaches 183.68 TPS on GSM8K, compared with 150.70 TPS for MMaDA under the same implementation settings (Tab.~\ref{tab:matched_inference}).
The larger comparisons with Qwen baselines in Fig.~\ref{fig:speedup_report} are implementation-level measurements across different decoding backends and should not be interpreted as an architecture-only speedup.
For TTS, the accelerated path in Fig.~\ref{fig:speedup_report} yields 5.43 WER, whereas the 2.1 WER in Tab.~\ref{tab:gen_speech} uses the standard benchmark decoding configuration.
Across sampling budgets, reasoning improves as the refinement budget increases, while image generation and editing reach their strongest or near-saturated performance at moderate step counts (Fig.~\ref{fig:sampling_steps}).
\begin{table}[t]
\centering
\tiny
\color{black}
\caption{Matched dInfer/DualCache comparison under shared implementation settings.}
\label{tab:matched_inference}
\setlength{\tabcolsep}{3.0pt}
\renewcommand{\arraystretch}{1.0}
\begin{tabular*}{\linewidth}{@{\extracolsep{\fill}}lrrrr@{}}
\toprule
Model & Shots & GSM8K$\uparrow$ & TPS$\uparrow$ & TPF$\uparrow$ \\
\midrule
MMaDA-MixCoT & 5 & 47.60 & 150.70 & 4.09 \\
\method{} & 0 & \textbf{84.23} & \textbf{183.68} & 3.51 \\
\bottomrule
\end{tabular*}
\end{table}

\subsection{Additional Quantitative and Qualitative Results}

\paragraph{Textual reasoning}
\label{app:text}

\begin{figure*}[tp] 
    \centering
    \vspace{-6em} 
  \makebox[\textwidth][c]{%
    \includegraphics[width=1.2\textwidth]{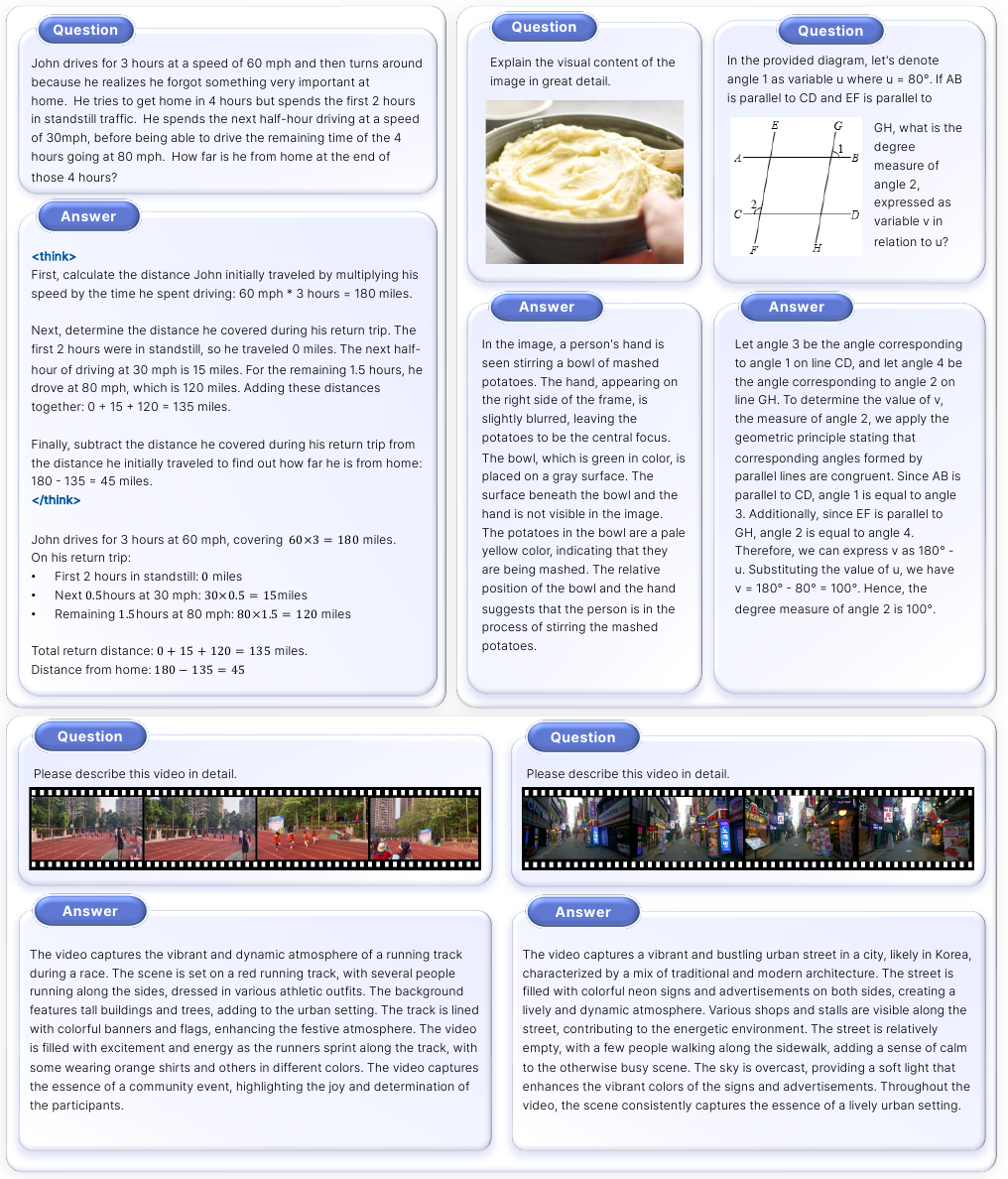}%
  }
    \caption{Qualitative examples of \method{} for textual reasoning (top-left), image understanding (top-right), and video understanding (bottom).}
    \label{fig:text_qual}

\end{figure*}

Tab.~\ref{tab:llm_benchmarks} reports textual reasoning performance across representative language benchmarks, and Fig.~\ref{fig:text_qual} provides qualitative examples for textual reasoning, image understanding, and video understanding.
\method{} remains competitive with prior unified and omnimodal models on the reported reasoning benchmarks, while text-specialized LLMs retain an advantage on several tasks.

\paragraph{Multimodal understanding}
\label{app:mmu}

Tab.~\ref{tab:mm_understanding} and Tab.~\ref{tab:video_understanding} summarize image and video understanding benchmarks, with qualitative examples shown in Fig.~\ref{fig:text_qual}.
\method{} remains competitive with perception-centric omnimodal systems despite additionally supporting native image generation, editing, and speech synthesis within the same backbone.
In particular, \method{} achieves strong MME-P and POPE performance while remaining competitive on MVBench, NExT-QA, and Video-MME.

\paragraph{Image generation}
\label{app:t2i}

\begin{figure*}[p]
  \centering
  \vspace{-5em}
  \makebox[\textwidth][c]{%
    \includegraphics[width=1.03\textwidth]{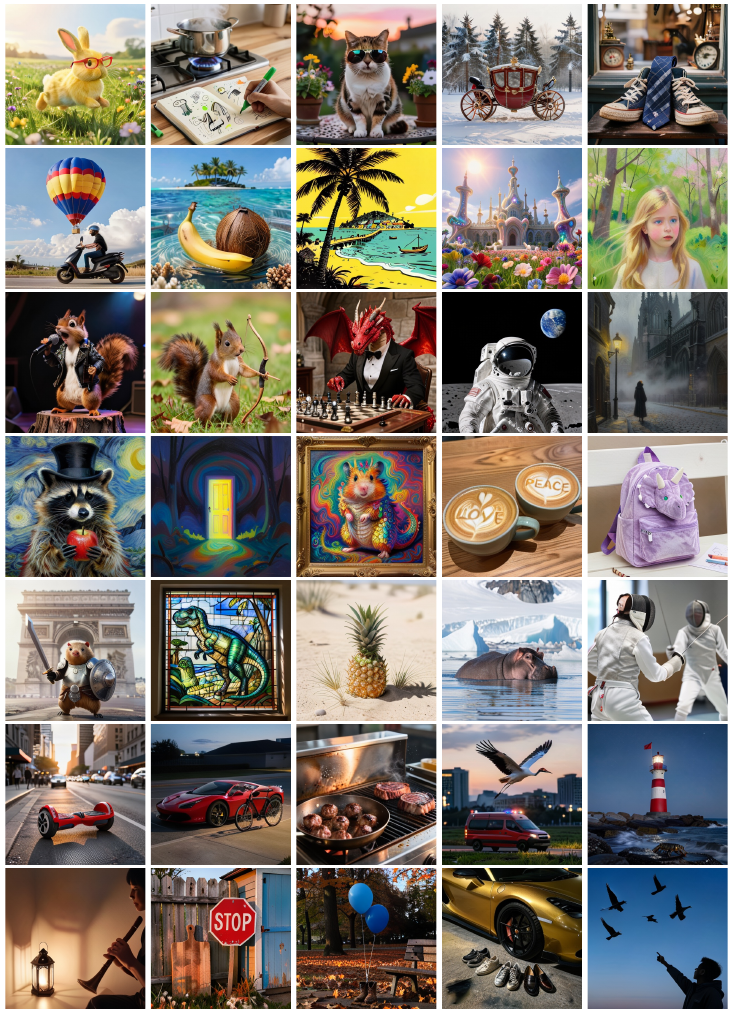}%
  }
  \caption{Qualitative examples of \method{} for image generation task.}
  \label{fig:t2i_qual}
\end{figure*}

Tab.~\ref{tab:geneval_full} reports image generation results, and Fig.~\ref{fig:t2i_qual} provides qualitative generation examples.
\method{} is competitive with prior unified models on GenEval and DPG-Bench and matches Qwen-Image on GenEval while also supporting video and speech.

\paragraph{Image editing}
\label{app:i2i}

\begin{figure*}[p]
  \centering
  \vspace{-7em}
  \makebox[\textwidth][c]{%
    \includegraphics[width=1.03\textwidth]{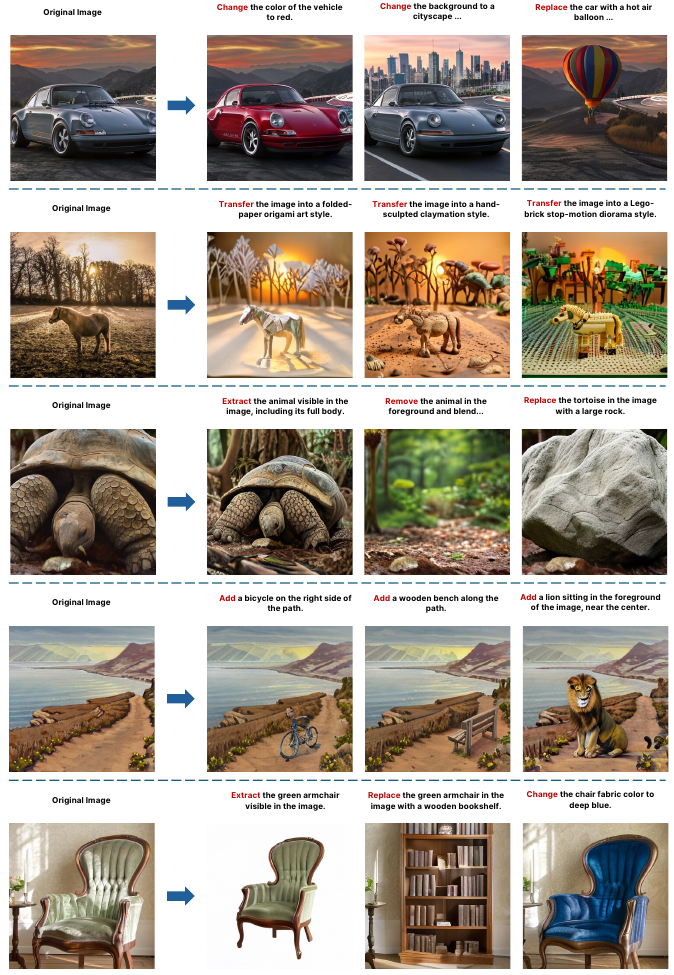}%
  }
    \caption{Qualitative examples of \method{} for image editing task.}
  \label{fig:i2i_qual}
\end{figure*}

Tab.~\ref{tab:image_edit} summarizes image editing performance across diverse editing categories, with qualitative examples shown in Fig.~\ref{fig:i2i_qual}.
\method{} demonstrates balanced editing capability across object addition, replacement, extraction, removal, and style transfer tasks.
Its Overall editing score is close to the strongest unified baselines while the same backbone also supports speech input and output.

\paragraph{Speech recognition and synthesis}
\label{app:speech}

Tab.~\ref{tab:speech_benchmarks_split} reports ASR and TTS performance.
\method{} achieves competitive speech recognition and synthesis quality while using the same masked-diffusion backbone employed for text, image, and video modeling.
Compared with the reported unified omnimodal baselines, \method{} obtains lower TTS WER on LibriSpeech test-clean while maintaining competitive ASR WER, without an external speech generator.

\paragraph{Omnimodal generation}
\label{app:omnimodal_generation_full}
Tab.~\ref{tab:omni_nonblock} and Tab.~\ref{tab:omni_block} provide modality-wise analysis of joint omnimodal generation and Modality Anchoring (MA) under non-block and block-wise decoding.
Non-block decoding refines all target tokens in parallel, whereas block-wise decoding generates sequential blocks and refines tokens in parallel within each block, using a fixed block size of 8.
We compare \textit{Joint}, which generates text, image, and speech in a single masked-diffusion process, with \textit{MA-Text}, \textit{MA-Image}, and \textit{MA-Speech}, which first generate an anchor modality and then condition the remaining modalities on it.
We also evaluate omnimodal consistency using Qwen3-Omni-30B-A3B-Instruct~\citep{xu2025qwen3omni} as a multimodal judge; the \emph{Omni} score measures whether generated text, image, and speech remain semantically aligned and narratively coherent.

At 128 NFEs, Joint scores 7.93 Overall/6.97 Omni and the strongest single anchor scores 7.91/6.93; with 256 NFEs, the best two-stage anchor reaches 7.95/7.05. Thus, joint decoding achieves comparable cross-modal quality with a lower sampling budget.
Performance improves with more NFEs and gradually saturates.
Among MA variants, text is the strongest anchor, while MA-Image substantially degrades text-related metrics; for example, under block-wise decoding at 64 NFEs, MA-Image obtains 3.38 on text LLM evaluation compared with 6.34 for Joint generation, which also lowers Omni consistency.
Speech scores remain relatively stable across strategies, but strong speech-specific performance alone does not imply higher global omnimodal coherence.
Finally, non-block decoding generally achieves stronger Omni and Overall scores at larger NFEs, e.g., Joint improves from 6.69/7.82 under block-wise decoding at 128 NFEs to 6.97/7.93 under non-block decoding, although block-wise decoding remains competitive for sequential modalities such as speech.

\begingroup
\subsection{Training Schedule and Vocabulary-Merging Controls}
\label{app:adaptation-controls}

\paragraph{Controlled adaptation setup.}
Full-scale adaptation required 63,425 H100 GPU-hours. To compare adaptation policies while retaining the full 8B backbone and new speech  vocabulary, we use a reduced-data control with 256 speech--transcript pairs from the EMOVA Alignment 7M LibriSpeech--AISHELL-2 training subset, Flickr8k examples for captioning and text-to-image, and language replay examples. The EOS  comparisons use the same MMaDA initialization, data,  optimizer, training batches, decoding, and  6,144 total updates. Each staged condition first receives 4,096 ASR/TTS alignment updates corresponding to Stage~1,  then continues without model merging for 2,048 multimodal integration  updates corresponding to Stage~2. We compare AlwaysOn, Off$\rightarrow$On, and Omni-Diffusion attenuation $\rightarrow$On ($\gamma=0.6$ ); all  staged conditions enable EOS supervision during  integration. Mixed AlwaysOn mixes all modalities from the start for 6,144 updates, with EOS supervision enabled throughout.

\begin{table}[htbp]

\centering
\caption{{EOS  schedules and  staged versus mixed training at the common 6,144-update endpoint. All WER values are percentages.}}
\label{tab:control-eos}
\tiny
\setlength{\tabcolsep}{3.5pt}
\renewcommand{\arraystretch}{1.05}
\textbf{(a) Whole-path EOS schedules.}\par\smallskip
\begin{tabular*}{\linewidth}{@{\extracolsep{\fill}}lrr@{}}\toprule
Condition & ASR WER $\downarrow$ & TTS WER $\downarrow$ \\
\midrule
Staged AlwaysOn & 71.29 & 16.41 \\
Staged attenuation$\rightarrow$On & 65.16 & 17.08 \\
Staged Off$\rightarrow$On & \textbf{64.27} & 15.57 \\
Mixed AlwaysOn & 80.19 & \textbf{14.78} \\
\bottomrule\end{tabular*}
\par\medskip
\textbf{(b) Staged versus mixed training across speech, text, and image tasks.}\par\smallskip
\begin{tabular*}{\linewidth}{@{\extracolsep{\fill}}lrrrrrr@{}}\toprule
Condition & ASR WER $\downarrow$ & TTS WER $\downarrow$ & Cap. BLEU $\uparrow$ & Text BLEU $\uparrow$ & CLIP-T $\uparrow$ & CLIP-I $\uparrow$ \\
\midrule
Mixed AlwaysOn & 80.19 & \textbf{14.78} & \textbf{11.91} & 13.77 & .3160 & .6981 \\
Staged AlwaysOn & 71.29 & 16.41 & 9.13 & \textbf{30.75} & .3382 & .7431 \\
Staged Off$\rightarrow$On & \textbf{64.27} & 15.57 & 10.77 & 27.59 & \textbf{.3398} & \textbf{.7435} \\
\bottomrule\end{tabular*}
\end{table}

 At the  common endpoint,  Off$\rightarrow$On improves ASR/TTS WER by 7.02/0.84 percentage points over staged AlwaysOn and 0.89/1.51 points over attenuation $\rightarrow$On.  It achieves the lowest ASR WER among all four conditions and the lowest ASR and TTS WER among the three staged schedules. For captioning, text-only generation, and image generation, all conditions use the same 32 inputs per task; text BLEU measures GSM8K solutions. Under the same AlwaysOn policy, staging improves ASR WER, text BLEU, and both image CLIP scores relative to mixed training. Mixed training achieves lower TTS WER and higher caption BLEU. The proposed Staged Off$\rightarrow$On setting further reduces ASR WER to 64.27\%, with TTS WER of 15.57\%.

\paragraph{Vocabulary rules after matched integration.}
The three vocabulary initializations share the same Stage~1  speech-aligned checkpoint and 5:5  interpolation of shared Transformer parameters. Table~\ref{tab:control-merge} compares  the  Shared, Stage~1-only, and modality-disentangled vocabulary rules before and after the same 8,192 Stage~2 updates on the multimodal training mixture. Caption BLEU, text BLEU, CLIP-T, and CLIP-I use  32 examples per task. Modality-disentangled merging leads both alternatives on all six reported metrics after the same update count. This reduced-data comparison complements the full-scale 10K-update merge-versus-no-merge comparison in Table~\ref{tab:training_recipe_effects}.
\begin{table}[htbp]

\centering
\tiny
\setlength{\tabcolsep}{3.5pt}
\renewcommand{\arraystretch}{1.05}
\caption{{Vocabulary merging rules before and after matched Stage~2 updates. ASR and TTS report WER; image results are CLIP scores.}}
\label{tab:control-merge}
\begin{tabular*}{\linewidth}{@{\extracolsep{\fill}}llrrrrrr@{}}\toprule
Updates & Rule & ASR $\downarrow$ & TTS $\downarrow$ & Cap. BLEU $\uparrow$ & Text BLEU $\uparrow$ & CLIP-T $\uparrow$ & CLIP-I $\uparrow$ \\
\midrule
0 & Shared & 289.84 & 98.37 & 4.41 & 10.74 & .3152 & .6527 \\
0 & Stage~1-only & 131.94 & 97.98 & 4.41 & 10.06 & .3064 & .6411 \\
0 & Ours & \textbf{108.12} & \textbf{88.24} & \textbf{6.43} & \textbf{11.15} & \textbf{.3297} & \textbf{.6693} \\
\midrule
8192 & Shared & 28.69 & 7.74 & 8.20 & 14.97 & .5164 & .8489 \\
8192 & Stage~1-only & 29.25 & 7.79 & 5.52 & 23.04 & .5127 & .8364 \\
8192 & Ours & \textbf{22.11} & \textbf{6.43} & \textbf{10.25} & \textbf{31.16} & \textbf{.5372} & \textbf{.8502} \\
\bottomrule\end{tabular*}
\end{table}

\subsection{Joint-Decoding Costs and Uncertainty}
\label{app:control-costs}
\paragraph{Actual joint-decoding costs.}
Table~\ref{tab:control-cost} reports forward calls,  Transformer FLOPs, latency, and peak memory on one B200, measured on  two inputs with fixed token lengths allocated to text, image, and speech. Each method pair uses the same image refinement setting. Budget denotes total denoising steps across passes. Paper Overall is from the separate benchmark evaluation in Table~\ref{tab:omni_nonblock}; the costs use the fixed token lengths and two inputs  described above.
\begin{table}[htbp]

\centering
\tiny
\setlength{\tabcolsep}{3.5pt}
\renewcommand{\arraystretch}{1.05}
\caption{{Measured inference costs and  separately evaluated generation quality.}}
\label{tab:control-cost}
\begin{tabular*}{\linewidth}{@{\extracolsep{\fill}}llrrrrr@{}}\toprule
Budget (NFE) & Method & Forwards & TFLOPs $\downarrow$ & Seconds $\downarrow$ & GiB $\downarrow$ & Paper Overall $\uparrow$ \\
\midrule
128 & Joint & 159 & 5642.32 & 10.543 & 17.48 & 7.93 \\
128 & MA-Text & 160 & 5677.81 & 10.612 & 17.48 & 7.91 \\
256 & MA-Text & 283 & 10042.62 & 18.398 & 17.48 & 7.95 \\
\bottomrule\end{tabular*}
\end{table}

Joint-128 reduces measured latency by 42.7\% and Transformer FLOPs by 43.8\% relative to MA-Text-256, while their reported Overall scores are 7.93 and 7.95, respectively. Equal-budget costs at 128 steps are nearly identical. Thus, joint decoding achieves similar reported quality at lower measured cost using half the denoising-step budget in this comparison.

\paragraph{Uncertainty and figure bands.}
Figure~\ref{fig:omnimodal_generation}'s shaded bands are $\pm0.2$ standard deviations across benchmark categories, not confidence intervals. On a separate 32-input, two-repeat panel,  the  Joint$-$MA-Text Overall  differences are $+0.1098$ [ $0.0550$,  $0.2198$] at 16 NFE and $+0.0598$ [$-0.1050$, $0.2198$] at 128 NFE (95\% paired bootstrap confidence intervals).  This panel is separate from the evaluation in Table~\ref{tab:omni_nonblock}. For supplementary reasoning panels, 20,000 item-bootstrap draws give GSM8K 207/256, 80.86\% [75.78, 85.55], and MATH 45/128, 35.16\% [27.34, 43.75]. These are separate panels from the full benchmark results.

\subsection{ Cross-Modal Attention Intervention}
\label{app:control-attention}
\paragraph{ Cross-modal attention intervention.}
 Using the same checkpoint,  64 inputs, seeds, and refinement budgets, we block attention in both directions between one generated modality and the other two  while retaining input context. All conditions use a 128-step budget (Table~\ref{tab:control-direction}).  Full joint generation obtains the best mean on every displayed metric. Speech WER compares the utterance extracted from each condition's generated text with the independent ASR transcript of its generated audio. These interventions support the role of cross-modal communication during denoising.
\begin{table}[htbp]

\centering
\tiny
\setlength{\tabcolsep}{3.5pt}
\renewcommand{\arraystretch}{1.05}
\caption{{ Cross-modal attention intervention with the same checkpoint, 64 inputs, and 128-step budget. Speech WER is reported as a  percentage.}}
\label{tab:control-direction}
 \begin{tabular*}{\linewidth}{@{\extracolsep{\fill}}lrrrrr@{}}\toprule
 Condition &  Text BERT $\uparrow$ &  Text judge $\uparrow$ &  Image CLIP $\uparrow$ & Image judge $\uparrow$ & Speech WER $\downarrow$ \\
\midrule
 Full joint &  \textbf{.9342} &  \textbf{7.629} &  \textbf{.8288} &  \textbf{6.496} &  \textbf{31.32} \\
 Block text connections &  .9108 &  7.160 &  .8057 & 6.016 & 67.09 \\
 Block image connections &  .9222 &  7.223 &  .7994 & 5.889 & 42.04 \\
 Block speech connections &  .9185 &  7.084 &  .8185 &  6.152 &  62.64 \\
\bottomrule\end{tabular*}
\end{table}

\paragraph{Comparison with a frozen anchor.}
Modality Anchoring generates an anchor modality first, keeps its tokens fixed, and generates the remaining modalities in a second pass. With the same checkpoint and total NFE budget across passes, Joint exceeds MA-Text in Overall at 16, 32, 64, and 128 NFE: 7.71/7.57, 7.76/7.62, 7.86/7.80, and 7.93/7.91, respectively (Table~\ref{tab:omni_nonblock}). Together with the attention intervention, this comparison supports generating and refining modalities jointly.

\subsection{Joint Generation Beyond Stories}
\label{app:control-nonstory}
\paragraph{Beyond-story joint generation.}
We evaluate the unchanged model on  RecipeQA cooking~\citep{yagcioglu2018recipeqa}, VisualHow~\citep{yang2022visualhow} DIY installation/repair, and  sports/body movement, with 32 inputs  per domain. Each model generates the next step's text, image, and speech given the current step. For VisualHow, the next action title, target  sentence, and image are withheld.  We compare \method{} with MMaDA+Parler,  Qwen2.5-Omni+FLUX.1 Kontext,  and  Omni-Diffusion in Table~\ref{tab:control-nonstory}.  \method{} generates all  three modalities within one masked-diffusion process without external modality-specific generator orchestration.
\begin{table}[htbp]

\centering
\tiny
 \setlength{\tabcolsep}{2.5pt}
\renewcommand{\arraystretch}{1.05}
\caption{{Nonstory domain comparison  with 32 inputs per domain. Speech consistency  and auxiliary Overall are on a 0--10 scale.}}
\label{tab:control-nonstory}
\begin{tabular*}{\linewidth}{@{\extracolsep{\fill}}llrrrrrr@{}}\toprule
Domain & Model & BERT $\uparrow$ & CLIP $\uparrow$ & Text judge $\uparrow$ & Image judge $\uparrow$ & Speech $\uparrow$ &  Aux. Overall $\uparrow$ \\
\midrule
Recipe & Ours & .8378 & .6273 & 4.086 & 3.992 & 5.750 &  5.696 \\
Recipe & MMaDA+Parler & .1375 & .0784 & .828 & .156 & 1.500 &  .929 \\
Recipe & Qwen+Kontext & .8199 & .6712 & 3.758 & 6.188 & 9.125 &  6.796 \\
Recipe & Omni-Diffusion & .7366 & .5118 & 1.891 & 1.508 & 8.156 &  4.808 \\
\midrule
DIY & Ours & .8622 & .6838 & 1.383 & 5.156 & 5.844 &  5.569 \\
DIY & MMaDA+Parler & .6777 & .3386 & 1.461 & .898 & 7.469 &  3.998 \\
DIY & Qwen+Kontext & .8232 & .6930 & 3.953 & 6.078 & 9.250 &  6.889 \\
DIY & Omni-Diffusion & .8518 & .4997 & 1.367 & 1.070 & 7.969 &  4.784 \\
\midrule
Movement & Ours & .8599 & .6927 & 2.133 & 4.648 & 5.438 &  5.549 \\
Movement & MMaDA+Parler & .7856 & .3896 & 1.172 & .945 & 9.219 &  4.618 \\
Movement & Qwen+Kontext & .7061 & .6879 & 3.680 & 5.188 & 9.750 &  6.511 \\
Movement & Omni-Diffusion & .8481 & .4910 & 1.625 & 1.305 & 8.688 &  5.002 \\
\bottomrule\end{tabular*}
\end{table}

Auxiliary Overall is the mean of $10\times$BERTScore, $10\times$image CLIP, text judge, image judge, and speech semantic consistency. It is not the official Omni-StoryBench Overall because these domains lack reference labels for speech emotion, speaking rate, pitch, and gender. Speech consistency compares the utterance extracted from generated text with an independent ASR transcript of the generated audio; missing audio or unextractable utterance text scores zero.

\method{} has the highest text BERTScore among the four systems in all three domains. Its auxiliary Overall scores, 5.696/5.569/5.549 for Recipe/DIY/Movement, exceed Omni-Diffusion's 4.808/4.784/5.002 and MMaDA+Parler's .929/3.998/4.618. Qwen+Kontext has higher auxiliary Overall in all three domains, at 6.796/6.889/6.511.

\subsection{Comparison with Omni-Diffusion}
\label{app:control-omnidiff}
\paragraph{Comparison with Omni-Diffusion.}
Table~\ref{tab:control-omnidiff} combines our reported benchmark results with Omni-Diffusion's reported results and a new common-input reasoning evaluation. For the latter we use 256 GSM8K and 128 MATH items with common grading and each model's native prompts and 1,024-step decoding. Reported-benchmark rows and new common-input rows have different protocols and are identified separately.
\begin{table}[htbp]

\centering
\tiny
\setlength{\tabcolsep}{3.5pt}
\renewcommand{\arraystretch}{1.05}
\caption{{Comparison with Omni-Diffusion. Reported rows use our benchmark results and Omni-Diffusion's reported results; common-input rows use shared inputs and grading.}}
\label{tab:control-omnidiff}
\begin{tabular*}{\linewidth}{@{\extracolsep{\fill}}llrr@{}}\toprule
Benchmark & Protocol & Dynin-Omni & Omni-Diffusion \\
\midrule
MME-Perception $\uparrow$ & Reported & 1672.9 & 1176.1 \\
MMMU-val (\%) $\uparrow$ & Reported & 52.8 & 31.1 \\
LibriSpeech test-clean ASR WER (\%) $\downarrow$ & Reported & 2.1 & 6.69 \\
GSM8K, 256 (\%) $\uparrow$ & Common input & 80.86 & 3.91 \\
MATH, 128 (\%) $\uparrow$ & Common input & 35.16 & 2.34 \\
\bottomrule\end{tabular*}
\end{table}

\endgroup

\FloatBarrier
\section{License and Safeguards}
\label{app:license}
\method{} model weights will be released under the CC BY 4.0 license, together with code and evaluation scripts for research use. We cite the datasets, tokenizers, pretrained models, benchmarks, and synthetic-data generation models used in training and evaluation throughout the paper, and will provide release documentation summarizing their sources, scales, licenses, and filtering procedures. The data-construction prompts, configurations, filters, and manifests described in App.~\ref{sec:data_composition} will also be released; restricted source media will be represented by source identifiers, hashes, and reconstruction scripts.
For responsible release, we will provide model documentation describing intended uses, limitations, and misuse risks, along with usage guidelines that prohibit harmful applications such as impersonation, deception, harassment, and generation of unsafe or illegal content. We also recommend downstream deployments to use additional safety classifiers, provenance or watermarking mechanisms, and human review for sensitive use cases.

\clearpage
\begin{table}[t]
\centering
\footnotesize
\caption{Evaluation on textual reasoning benchmarks (higher is better for all metrics).
Notation follows Tab.~\ref{tab:understanding}.}
\label{tab:llm_benchmarks}
\setlength{\tabcolsep}{13pt}
\renewcommand{\arraystretch}{0.95}
\begin{adjustbox}{max width=\linewidth}
\begin{tabular}{lccccc}
\toprule
\textbf{Model} &
\textbf{MMLU~\citep{mmlu}} &
\textbf{ARC-C~\citep{arc}} &
\textbf{GSM8K~\citep{cobbe2021gsm8k}} &
\textbf{MATH~\citep{hendrycksmath2021}} &
\textbf{GPQA~\citep{rein2024gpqa}} \\
\midrule
\multicolumn{6}{l}{\textbf{Text-Experts}} \\
\midrule
Llama-3-8B~\citep{grattafiori2024llama3}        & 66.6 & - & 55.3 & 20.5 & 25.8 \\
Ministral-3-8B~\citep{liu2026ministral3}    & 76.1 & 88.0 & - & 62.6 & - \\
Qwen3-8B~\citep{yang2025qwen3}          & 76.9 & - & 89.8 & 60.8 & 44.4 \\
LLaDA-8B-Base$^{\ddagger}$~\citep{nie2025large}
    & 65.9 & -- & 71.3 & 30.0 & 31.5 \\
LLaDA-8B-Instruct$^{\ddagger}$~\citep{nie2025large}
    & 65.4 & -- & 68.7 & 29.8 & 31.8 \\
Trida-7B~\citep{trida}          & 67.2 & - & 65.1 & 33.6 & - \\
\midrule
\multicolumn{6}{l}{\textbf{Perception-centric}} \\
\midrule
Qwen2.5-Omni-7B~\citep{Qwen2.5-Omni}           & 71.8 & - & 88.7 & 71.5 & 30.8 \\
Baichuan-Omni-1.5~\citep{li2025baichuan}         & 72.2 & - & - & - & - \\
\midrule
\multicolumn{6}{l}{\textbf{Unified}} \\
\midrule
Chameleon-7B~\citep{team2024chameleon}  & 52.1 & 46.5 & 41.6 & 11.5 & - \\
MMaDA~\citep{MMaDA}  &  68.4 & 57.4 & 73.4 & 36.0 & 28.4 \\
Show-o2$^{*}$~\citep{xie2025showo2}       & 70.7   & --   & 75.3   & --   & 31.5   \\
HyperCLOVAX-8B-Omni$^{\dagger}$~\citep{team2026hyperclova}  &  75.7  & 85.8$^{\ddagger}$ & 87.3 & 43.4$^{\ddagger}$ & 28.6$^{\ddagger}$ \\ 
\rowcolor{lightblue}\method{}$^{\dagger}$ (Ours)  & \nipsver{75.9}  & \nipsver{68.5}   & \nipsver{83.7} &  \nipsver{47.4} & \nipsver{37.9} \\
\bottomrule
\end{tabular}
\end{adjustbox}
\end{table}

\begin{table*}[!t]
\centering
\footnotesize
\caption{Evaluation on multimodal understanding (MMU) benchmarks (\emph{image}). Higher is better for all metrics. Notation follows Tab.~\ref{tab:understanding}.}
\label{tab:mm_understanding}
\setlength{\tabcolsep}{13pt}
\renewcommand{\arraystretch}{0.95}
\begin{adjustbox}{max width=\linewidth}
\begin{tabular}{lccccc}
\toprule
\textbf{Model} &
\textbf{POPE~\citep{pope}} &
\textbf{MME-P~\citep{fu2023mme}} &
\textbf{GQA~\citep{hudson2019gqa}} &
\textbf{MMMU~\citep{yue2023mmmu}} &
\textbf{MMBench~\citep{MMBench}} \\
\midrule
\multicolumn{6}{l}{\textbf{MMU-Experts}} \\
\midrule
InternVL3-8B~\citep{zhu2025internvl3} & 90.4 & 1748.4 & -- & 62.2 & 83.6 \\
Ovis2-8B~\citep{lu2024ovis} & 88.6 & 1628.1 & -- & 57.4 & 84.8 \\
Qwen2.5-VL-7B~\citep{Qwen2.5-VL} & 85.9 & 1698.1 & -- & 58.0 & 83.2 \\
LLaDA-V~\citep{you2025lladav} & -- & 1507.0 & -- & 48.6 & 82.9 \\
\midrule
\multicolumn{6}{l}{\textbf{Perception-centric}} \\
\midrule
OpenOmni~\citep{luo2025openomni} & 86.4$^{\ddagger}$ & 1536.9$^{\ddagger}$ & 62.2$^{\ddagger}$ & 46.7 & 76.2 \\
Baichuan-Omni-1.5~\citep{li2025baichuan} & 87.3$^{\ddagger}$ & 1632.1$^{\ddagger}$ & 58.6$^{\ddagger}$ & 53.9 & 85.6 \\
Qwen2.5-Omni-7B~\citep{Qwen2.5-Omni} & 86.8$^{\ddagger}$ & 1481.3$^{\ddagger}$ & 49.1$^{\ddagger}$ & 59.2 & 81.8 \\
OmniVinci~\citep{ye2025omnivinci} & 89.5$^{\ddagger}$ & 1651.0$^{\ddagger}$ & 65.7$^{\ddagger}$ & 49.7 & - \\
\midrule
\multicolumn{6}{l}{\textbf{Unified}} \\
\midrule
SEED-X~\citep{ge2024seed} & 84.2 & 1435.7 & 47.9 & 35.6 & 70.1 \\
Janus-Pro~\citep{chen2025januspro} & 87.4 & 1567.1 & 62.0 & 41.0 & 79.2 \\
BAGEL~\citep{bagle} & -- & 1687.0 & -- & 55.3 & 85.0 \\
Emu3$^{*}$~\citep{wang2024emu3} & 85.2 & -- & 60.3 & 31.6 & 58.5 \\
MMaDA~\citep{MMaDA} & 86.1 & 1410.7 & 61.3 & 30.2 & 68.5 \\
Fudoki~\citep{wang2025fudoki} & 86.1 & 1485.4 & 57.6 & 34.3 & 73.9 \\
Lumina-DiMOO~\citep{xin2025luminadimooomnidiffusionlarge} & 87.4 & 1534.2 & -- & 58.6 & 83.1 \\
Show-o2$^{*}$~\citep{xie2025showo2} & -- & 1620.5 & 63.1 & 48.9 & 79.3 \\
NExT-OMNI$^{*,\dagger}$~\citep{next-omni} & 87.4 & 1537.8 & 62.7 & 43.7 & 78.9 \\
HyperCLOVAX-8B-Omni$^{\dagger}$~\citep{team2026hyperclova} & 79.8$^{\ddagger}$ & 1307.0$^{\ddagger}$ & 45.5$^{\ddagger}$ & 31.0$^{\ddagger}$ & 65.9$^{\ddagger}$ \\
\rowcolor{lightblue}\method{}$^{\dagger}$ (Ours) & \nipsver{88.0} & \nipsver{1672.9} & \nipsver{57.0} & \nipsver{52.8} & \nipsver{74.7} \\
\bottomrule
\end{tabular}
\end{adjustbox}
\end{table*}

\begin{table*}[!t]
\centering
\footnotesize
\caption{Evaluation on multimodal understanding (MMU) benchmarks (\emph{Video}). Higher is better for all metrics. Notation follows Tab.~\ref{tab:understanding}.}
\label{tab:video_understanding}
\setlength{\tabcolsep}{8pt}
\renewcommand{\arraystretch}{0.95}
\begin{adjustbox}{max width=\linewidth}
\begin{tabular}{lccccc}
\toprule
\textbf{Model} &
\textbf{ActivityNet~\citep{yu2019activitynet}} &
\textbf{MVBench~\citep{li2024mvbench}} &
\textbf{NExT-QA~\citep{xiao2021nextqa}} &
\textbf{TempCompass ~\citep{liu2024tempcompass}} &
\textbf{Video-MME~\citep{fu2024videomme}} \\
\midrule
\multicolumn{6}{l}{\textbf{MMU-Experts}} \\
\midrule
InternVL3-8B~\citep{zhu2025internvl3} & 63.9$^{\ddagger}$ & 73.2 & 82.0$^{\ddagger}$ & 70.4$^{\ddagger}$ & 66.0 \\
InternVL2.5-8B~\citep{chen2024internvl} & 62.8$^{\ddagger}$ & 70.5 & 84.1$^{\ddagger}$ & 67.7$^{\ddagger}$ & 63.7 \\
IXC-2.5~\citep{icx2_5} & 52.8 & 69.1 & 55.8 & -- & 62.8 \\
LLaVA-OV~\citep{li2024llavaonevision} & 56.6 & 56.7 & 79.4 & 64.8 & 58.2 \\
LLaDA-V~\citep{you2025lladav} & -- & 53.1 & -- & -- & 56.4 \\
\midrule
\multicolumn{6}{l}{\textbf{Perception-centric}} \\
\midrule
MiniCPM-o-2.6~\citep{yao2024minicpm} & 63.0 & 58.6 & - & 57.0 & 63.4 \\
VITA-1.5~\citep{fu2025vita} & 59.6 & 55.5 & - & -- & 57.6 \\
Baichuan-Omni-1.5~\citep{li2025baichuan} & 62.0 & 63.7 & -- & -- & 60.1 \\
Qwen2.5-Omni-7B~\citep{Qwen2.5-Omni} & 62.7$^{\ddagger}$ & 70.3 & 76.4$^{\ddagger}$ & 69.3$^{\ddagger}$ & 64.3 \\
\midrule
\multicolumn{6}{l}{\textbf{Unified}} \\
\midrule
Show-o2$^{*}$~\citep{xie2025showo2} & 56.4 & 55.8 & 79.0 & -- & 57.4 \\
HyperCLOVAX-8B-Omni$^{\dagger}$~\citep{team2026hyperclova} & - & 49.5$^{\ddagger}$ & - & 58.3$^{\ddagger}$ & 58.2 \\
\rowcolor{lightblue}\method{}$^{\dagger}$ (Ours) & \nipsver{56.8} & \nipsver{66.4} & \nipsver{81.3} & \nipsver{66.7} & \nipsver{63.4} \\
\bottomrule
\end{tabular}
\end{adjustbox}
\end{table*}

\clearpage
\begin{table}[t]
\centering
\footnotesize
\caption{Evaluation on image generation benchmarks (higher is better for all metrics).
Notation follows Tab.~\ref{tab:understanding}.}
\label{tab:geneval_full}
\setlength{\tabcolsep}{6.75pt}
\renewcommand{\arraystretch}{0.95}
\begin{adjustbox}{max width=\linewidth}
\begin{tabular}{l ccccccc c}
\toprule
 &
\multicolumn{7}{c}{\textbf{GenEval~\citep{ghosh2023geneval}}} &
 \\
\cmidrule(lr){2-8}
\textbf{Model} &
\textbf{Single Obj.} &
\textbf{Two Obj.} &
\textbf{Count.} &
\textbf{Colors} &
\textbf{Pos.} &
\textbf{Color Attr.} &
\textbf{Overall} & \textbf{DPG-Bench~\citep{hu2024ella_dpgbench}} \\
\midrule
\multicolumn{8}{l}{\textbf{Generation-Experts}} \\
\midrule
HiDream-I1-Full~\citep{hidreami1technicalreport}   & 1.00 & 0.98 & 0.79 & 0.91 & 0.60&  0.72 & 0.83 & 85.9 \\
FLUX.1 [Dev]~\citep{flux2024}      & 0.98& 0.81 &0.74& 0.79 &0.22 &0.45& 0.66 & 83.8 \\
Qwen-Image~\citep{wu2025qwenimagetechnicalreport} & 0.99 & 0.92 & 0.89 & 0.88 & 0.76 & 0.77 & 0.87 & 88.3 \\
GPT-Image 1 [High]~\citep{gptimage} & 0.99 & 0.92 & 0.85 & 0.92 & 0.75 & 0.61  & 0.84 & 85.2 \\
Seedream 3.0~\citep{gao2025seedream3} & 0.99 & 0.96 & 0.91 & 0.93 & 0.47 & 0.80  & 0.84 & 88.3 \\
\midrule
\multicolumn{8}{l}{\textbf{Unified}} \\
\midrule
Emu3$^{*}$~\citep{wang2024emu3}     & 0.99 & 0.81&  0.42 & 0.80 & 0.49 & 0.45 & 0.66 & 81.6 \\
Janus-Pro~\citep{chen2025januspro} & 0.99 & 0.89 & 0.59 & 0.90 & 0.79 & 0.66 & 0.80 & 84.2 \\
BAGEL~\citep{bagle}     & 0.99 & 0.94 & 0.81 & 0.88 & 0.64 & 0.63 & 0.82 & 85.1 \\
MMaDA~\citep{MMaDA}     & 0.99 & 0.76 & 0.61 & 0.84 & 0.20 & 0.37 & 0.63 & 70.0 \\
Fudoki~\citep{wang2025fudoki}     & 0.96 & 0.85 & 0.56 & 0.88 & 0.68 & 0.67 & 0.77 & 83.6 \\
Lumina-DiMOO~\citep{xin2025luminadimooomnidiffusionlarge} & 1.00 & 0.94 & 0.85 & 0.89&  0.85 & 0.76 & 0.88 & 86.0 \\
Show-o2$^{*}$~\citep{xie2025showo2}    & 1.00 &0.87& 0.58 &0.92& 0.52 &0.62  & 0.76 & 86.1 \\
NExT-OMNI$^{*,\dagger}$~\citep{next-omni}    & 0.99 & 0.92 & 0.79&  0.85 & 0.78&  0.74 & 0.85 & 84.2 \\
HyperCLOVAX-8B-Omni$^{\dagger}$~\citep{team2026hyperclova}    & -- & -- & -- & -- & -- & -- & 0.64 & 60.8$^{\ddagger}$ \\
\rowcolor{lightblue}\method{}$^{\dagger}$ (Ours)& \nipsver{0.98} & \nipsver{0.93} & \nipsver{0.82} & \nipsver{0.90} & \nipsver{0.86} & \nipsver{0.74} & \nipsver{0.87} & \nipsver{85.5} \\
\bottomrule
\end{tabular}
\end{adjustbox}
\end{table}

\begin{table*}[t]
\centering
\footnotesize
\caption{Evaluation on the ImgEdit benchmark~\citep{ye2025imgedit}. 
Higher is better. All metrics are evaluated using GPT-4.1. 
\emph{Overall} denotes the average score across all tasks. 
Notation follows Tab.~\ref{tab:understanding}.}
\label{tab:image_edit}
\setlength{\tabcolsep}{7pt}
\renewcommand{\arraystretch}{0.95}
\begin{adjustbox}{max width=\linewidth}
\begin{tabular}{lccccccccc}
\toprule
\textbf{Model} &
\textbf{Add} &
\textbf{Adjust} &
\textbf{Extract} &
\textbf{Replace} &
\textbf{Remove} &
\textbf{Background} &
\textbf{Style} &
\textbf{Action} &
\textbf{Overall} \\
\midrule
\multicolumn{10}{l}{\textbf{Editing-Experts}} \\
\midrule
Step1X-Edit~\citep{liu2025step1x}         & 3.88 & 3.14 & 1.76 & 3.40 & 2.41 & 3.16 & 4.63 & 2.52 & 3.11 \\
OmniGen2~\citep{wu2025omnigen2}            & 3.57 & 3.06 & 1.77 & 3.74 & 3.20 & 3.57 & 4.81 & 4.68 & 3.55 \\
Flux.1 Kontext [Pro]~\citep{labs2025flux1kontextflowmatching}       & 3.76 & 3.45 & 2.15 & 3.98 & 2.94 & 3.78 & 4.38 & 4.26 & 3.59 \\
GPT-Image 1 [High]~\citep{gptimage} & 4.61 &  4.33&  2.90&  4.35& 3.66 & 4.57 & 4.93 & 4.89 & 4.28\\
Qwen-Image~\citep{wu2025qwenimagetechnicalreport} & 4.38 & 4.16 & 3.43 &4.66 &4.14 &4.38 &4.81 &4.69 & 4.33\\ 
\midrule
\multicolumn{10}{l}{\textbf{Unified}} \\

\midrule
UniWorld-V1~\citep{lin2025uniworld}         & 3.82 & 3.64 & 2.27 & 3.47 & 3.24 & 2.99 & 4.21 & 2.74 & 3.30 \\
BAGEL~\citep{bagle}               & 3.56 & 3.31 & 1.70 & 3.30 & 2.62 & 3.24 & 4.49 & 4.17 & 3.30 \\
LaViDa-O~\citep{li2025lavidao}            & 4.04 & 3.62 & 2.01 & 4.39 & 3.98 & 4.06 & 4.82 & 3.54 & 3.81 \\
Lumina-DiMOO~\citep{xin2025luminadimooomnidiffusionlarge}  & 3.82    & --   & --    & 3.83    & 2.76    & --    & 4.18    & --    & -- \\
HyperCLOVAX-8B-Omni$^{\dagger}$~\citep{team2026hyperclova}        & --    & --    & --    & --    & --    & --    & --    & -    & 3.83 \\
\rowcolor{lightblue}\method{}$^{\dagger}$ (Ours) &

\nipsver{4.15} & \nipsver{3.45} & \nipsver{3.30} & \nipsver{4.29} &

\nipsver{3.98} & \nipsver{3.74} & \nipsver{4.10} & \nipsver{3.54} &

\nipsver{3.82} \\
\bottomrule
\end{tabular}
\end{adjustbox}
\end{table*}

\begin{table*}[t]
\centering
\footnotesize
\caption{Evaluation on automatic speech recognition (ASR) and text-to-speech (TTS) benchmarks. 
Lower is better. All metrics report word error rate (WER) and are evaluated using Whisper-large-v3~\citep{radford2022whisper}. Notation follows Tab.~\ref{tab:understanding}.}
\label{tab:speech_benchmarks_split}
\setlength{\tabcolsep}{10.0pt}
\renewcommand{\arraystretch}{0.95}

\begin{subtable}[t]{0.445\linewidth}
\centering
\caption{ASR Benchmarks (WER$\downarrow$)}
\begin{adjustbox}{max width=\linewidth}
\begin{tabular}{lcc}
\toprule
\textbf{Model} & \multicolumn{2}{c}{\textbf{LibriSpeech~\citep{panayotov2015librispeech}}} \\
\cmidrule(lr){2-3}
 & \textbf{Clean} & \textbf{Other} \\
\midrule
\multicolumn{2}{l}{\textbf{ASR-Experts}} \\
\midrule
SpeechVerse~\citep{das2024speechverse}            & 2.1 & 4.4 \\
MinMo~\citep{chen2025minmo}                   & 1.7 & 3.9 \\
Qwen2-Audio~\citep{Qwen2-Audio}              & 1.6 & 3.6 \\
Qwen3-ASR-1.7B~\citep{Qwen3-ASR}       & 1.6  & 3.4  \\
\midrule
\multicolumn{2}{l}{\textbf{Perception-centric}} \\
\midrule

EMOVA-7B~\citep{chen2024emova}             & 4.0 & --  \\ 
OpenOmni~\citep{luo2025openomni}             & 3.1 & 7.0  \\ 
VITA-1.5~\citep{fu2025vita}             & 3.0 & 7.2  \\ 
Qwen2.5-Omni-7B~\citep{Qwen2.5-Omni}      & 1.8  & 3.4  \\ 
MiniCPM-o-2.6~\citep{yao2024minicpm}            & 1.7 & --  \\ 
\midrule
\multicolumn{2}{l}{\textbf{Unified}} \\
\midrule
HyperCLOVAX-8B-Omni~\citep{team2026hyperclova}         & 1.9  & 4.5$^{\ddagger}$  \\ 
NExT-OMNI$^{*}$~\citep{next-omni}            & 3.1 & 7.0 \\ 
\rowcolor{lightblue}\method{} (Ours)           & \nipsver{2.1} & \nipsver{4.8}  \\ 
\bottomrule
\end{tabular}
\end{adjustbox}
\end{subtable}
\begin{subtable}[t]{0.515\linewidth}
\centering
\caption{TTS Benchmarks (WER$\downarrow$)}
\begin{adjustbox}{max width=\linewidth}
\begin{tabular}{lcc}
\toprule
\textbf{Model} & \textbf{LibriSpeech} & \textbf{SEED-TTS~\citep{anastassiou2024seedtts}} \\
\midrule
\multicolumn{3}{l}{\textbf{TTS-Experts}} \\
\midrule
MaskGCT~\citep{wang2024maskgct}                 & 2.6  & 2.6 \\
Llasa-8B~\citep{ye2025llasa}                & 1.5  & 3.0 \\
MiniMax-Speech~\citep{zhang2025minimax}          & --   & 1.7 \\
Qwen3-TTS-1.7B~\citep{hu2026qwen3tts}          & 6.6$^{\ddagger}$   & 1.2 \\
\midrule
\multicolumn{3}{l}{\textbf{Perception-centric}} \\
\midrule
EMOVA-7B~\citep{chen2024emova}                & 3.6  & --  \\
OpenOmni~\citep{luo2025openomni}                & 3.4  & --  \\
Qwen2.5-Omni-7B~\citep{Qwen2.5-Omni}         & 7.0$^{\ddagger}$    & 2.3 \\
\midrule
\multicolumn{3}{l}{\textbf{Unified}} \\
\midrule
HyperCLOVAX-8B-Omni~\citep{team2026hyperclova}            & 7.9$^{\ddagger}$   & 7.7$^{\ddagger}$  \\
NExT-OMNI$^{*}$~\citep{next-omni}               & 3.1  & --  \\
\rowcolor{lightblue}\method{} (Ours)             & \nipsver{2.1}  & \nipsver{4.3} \\
\bottomrule
\end{tabular}
\end{adjustbox}
\end{subtable}

\end{table*}

\clearpage
\begin{table}[t]
\centering
\footnotesize
\caption{Omni-StoryBench results under non-block decoding.}
\label{tab:omni_nonblock}
\setlength{\tabcolsep}{11pt}
\renewcommand{\arraystretch}{0.95}
\begin{adjustbox}{max width=\linewidth}

\begin{tabular}{c l cccc c cc}
\toprule

&
&
\multicolumn{2}{c}{\textbf{Text$\uparrow$}}
& \multicolumn{2}{c}{\textbf{Image$\uparrow$}}
& \multicolumn{1}{c}{\textbf{Speech$\uparrow$}}
&
&
\\

\cmidrule(lr){3-4}
\cmidrule(lr){5-6}
\cmidrule(lr){7-7}

\textbf{NFE}
& \textbf{Method}
& \textbf{BERT}
& \textbf{LLM Judge}
& \textbf{CLIP}
& \textbf{LLM Judge}
& \textbf{LLM Judge}
& \textbf{Omni}
& \textbf{Overall}
\\

\midrule

\multirow{4}{*}{16}
& Joint     & 0.91 & 6.17 & 0.85 & 6.29 & 8.45 & 6.32 & 7.71 \\
& MA-Text   & 0.91 & 5.88 & 0.84 & 6.03 & 8.45 & 6.13 & 7.57 \\
& MA-Image  & 0.85 & 2.27 & 0.84 & 5.44 & 8.44 & 3.31 & 6.61 \\
& MA-Speech & 0.91 & 5.60 & 0.84 & 6.07 & 8.50 & 5.92 & 7.53 \\
\midrule

\multirow{4}{*}{32}
& Joint     & 0.91 & 6.03 & 0.86 & 6.39 & 8.68 & 6.51 & 7.76 \\
& MA-Text   & 0.91 & 5.50 & 0.85 & 6.29 & 8.75 & 6.18 & 7.62 \\
& MA-Image  & 0.85 & 2.54 & 0.85 & 6.30 & 8.55 & 4.20 & 6.89 \\
& MA-Speech & 0.91 & 5.37 & 0.85 & 6.30 & 8.59 & 6.02 & 7.57 \\
\midrule

\multirow{4}{*}{64}
& Joint     & 0.92 & 6.51 & 0.86 & 6.37 & 8.69 & 6.83 & 7.86 \\
& MA-Text   & 0.91 & 6.22 & 0.85 & 6.33 & 8.73 & 6.68 & 7.80 \\
& MA-Image  & 0.87 & 3.20 & 0.85 & 6.35 & 8.58 & 4.82 & 7.07 \\
& MA-Speech & 0.91 & 5.86 & 0.85 & 6.38 & 8.58 & 6.43 & 7.70 \\
\midrule

\multirow{4}{*}{128}
& Joint     & 0.92 & 6.78 & 0.86 & 6.40 & 8.70 & 6.97 & 7.93 \\
& MA-Text   & 0.92 & 6.66 & 0.86 & 6.40 & 8.74 & 6.93 & 7.91 \\
& MA-Image  & 0.88 & 3.81 & 0.86 & 6.38 & 8.65 & 5.24 & 7.24 \\
& MA-Speech & 0.92 & 6.29 & 0.86 & 6.39 & 8.55 & 6.57 & 7.79 \\
\midrule

\multirow{3}{*}{256}
& MA-Text   & 0.92 & 6.89 & 0.86 & 6.36 & 8.74 & 7.05 & 7.95 \\
& MA-Image  & 0.89 & 4.18 & 0.86 & 6.41 & 8.69 & 5.51 & 7.34 \\
& MA-Speech & 0.92 & 6.52 & 0.86 & 6.37 & 8.61 & 6.72 & 7.85 \\

\bottomrule
\end{tabular}
\end{adjustbox}
\end{table}

\begin{table}[t]
\centering
\footnotesize
\caption{Omni-StoryBench results under block-wise decoding.}
\label{tab:omni_block}
\setlength{\tabcolsep}{11pt}
\renewcommand{\arraystretch}{0.95}
\begin{adjustbox}{max width=\linewidth}
\begin{tabular}{c l cccc c cc}
\toprule

&
&
\multicolumn{2}{c}{\textbf{Text$\uparrow$}}
& \multicolumn{2}{c}{\textbf{Image$\uparrow$}}
& \multicolumn{1}{c}{\textbf{Speech$\uparrow$}}
&
&
\\

\cmidrule(lr){3-4}
\cmidrule(lr){5-6}
\cmidrule(lr){7-7}

\textbf{NFE}
& \textbf{Method}
& \textbf{BERT}
& \textbf{LLM Judge}
& \textbf{CLIP}
& \textbf{LLM Judge}
& \textbf{LLM Judge}
& \textbf{Omni}
& \textbf{Overall}
\\

\midrule

\multirow{4}{*}{16}
& Joint     & 0.90 & 5.05 & 0.85 & 6.25 & 8.32 & 5.60 & 7.43 \\
& MA-Text   & 0.90 & 4.62 & 0.84 & 6.05 & 8.33 & 5.24 & 7.27 \\
& MA-Image  & 0.83 & 1.82 & 0.84 & 6.06 & 8.26 & 3.02 & 6.56 \\
& MA-Speech & 0.90 & 4.53 & 0.84 & 6.08 & 8.44 & 5.06 & 7.28 \\
\midrule

\multirow{4}{*}{32}
& Joint     & 0.91 & 5.92 & 0.85 & 6.37 & 8.35 & 6.28 & 7.66 \\
& MA-Text   & 0.90 & 5.33 & 0.85 & 6.27 & 8.35 & 5.90 & 7.50 \\
& MA-Image  & 0.85 & 2.54 & 0.85 & 6.28 & 8.29 & 3.83 & 6.83 \\
& MA-Speech & 0.90 & 5.05 & 0.85 & 6.28 & 8.48 & 5.67 & 7.46 \\
\midrule

\multirow{4}{*}{64}
& Joint     & 0.91 & 6.34 & 0.86 & 6.43 & 8.43 & 6.60 & 7.78 \\
& MA-Text   & 0.91 & 6.23 & 0.86 & 6.39 & 8.42 & 6.54 & 7.74 \\
& MA-Image  & 0.87 & 3.38 & 0.85 & 6.35 & 8.39 & 4.76 & 7.08 \\
& MA-Speech & 0.91 & 5.81 & 0.86 & 6.37 & 8.48 & 6.26 & 7.66 \\
\midrule

\multirow{4}{*}{128}
& Joint     & 0.91 & 6.53 & 0.86 & 6.37 & 8.50 & 6.69 & 7.82 \\
& MA-Text   & 0.91 & 6.51 & 0.86 & 6.33 & 8.45 & 6.66 & 7.80 \\
& MA-Image  & 0.88 & 3.97 & 0.86 & 6.39 & 8.43 & 5.26 & 7.23 \\
& MA-Speech & 0.91 & 6.14 & 0.86 & 6.40 & 8.57 & 6.49 & 7.76 \\
\midrule

\multirow{3}{*}{256}
& MA-Text   & 0.91 & 6.53 & 0.86 & 6.35 & 8.42 & 6.75 & 7.81 \\
& MA-Image  & 0.88 & 4.01 & 0.86 & 6.40 & 8.40 & 5.27 & 7.24 \\
& MA-Speech & 0.89 & 6.10 & 0.84 & 6.22 & 8.44 & 6.43 & 7.62 \\

\bottomrule
\end{tabular}
\end{adjustbox}
\end{table}

\clearpage

\clearpage

\clearpage

\clearpage

\FloatBarrier
\clearpage

\end{document}